\documentclass[11pt, a4paper]{memoir}
\usepackage[english, science, dropcaps, hyperref, submissionstatement]{ku-frontpage}
\usepackage[utf8]{inputenc}

\usepackage{pdfpages}
\usepackage{booktabs}
\usepackage{amsmath}
\usepackage{amssymb}
\usepackage{url}
\usepackage{subcaption}
\usepackage[backend=biber, style=authoryear, natbib=true,
            maxcitenames=2, mincitenames=1, maxbibnames=99,
            uniquelist=false]{biblatex}
\newcommand{\mypartref}[1]{Part~\ref{#1}}
\newcommand{\mychapref}[1]{Chapter~\ref{#1}}
\newcommand{\mysecref}[1]{Section~\ref{#1}}
\newcommand{\myappref}[1]{Appendix~\ref{#1}}
\newcommand{\myfigref}[1]{Figure~\ref{#1}}
\newcommand{\myeqnref}[1]{Equation~(\ref{#1})}
\newcommand{\mytabref}[1]{Table~\ref{#1}}

\makeatletter
\newcommand{\MonthName}[1]{%
  \ifcase#1
    \or January\or February\or March\or April\or May\or June%
    \or July\or August\or September\or October\or November\or December%
  \fi}

\newcommand{\addthesuffix}[1]{%
  \ifnum#1=11 \textsuperscript{th}\else%
    \ifnum#1=12 \textsuperscript{th}\else%
      \ifnum#1=13 \textsuperscript{th}\else%
        \ifcase\numexpr#1\relax\mod10 %
          \textsuperscript{th}
        \or \textsuperscript{st}
        \or \textsuperscript{nd}
        \or \textsuperscript{rd}
        \else \textsuperscript{th}
        \fi%
      \fi%
    \fi%
  \fi%
}

\newcommand{\fancytoday}{%
  \MonthName{\month}~\number\day\addthesuffix{\day},~\number\year%
}
\makeatother

\newcommand{\mb}[1]{\mathbf{#1}}
\def\bTR{\mb{T}}
\def\X{\mb{X}}
\def\XT{\X^\bTR}
\def\Y{\mb{Y}}
\def\YT{\Y^{\bTR}}
\def\B{\mb{B}}
\def\w{\mb{w}}

\def\y{\mb{y}}

\settocdepth{chapter}
\setsecnumdepth{subsection}

\assignment{PhD thesis}
\author{Ole-Christian Galbo Engstr{\o}m}

\frontpageauthor{Ole-Christian Galbo Engstr{\o}m}
\frontpagetitle{Near-Infrared Hyperspectral Imaging\\Applications in Food Analysis}
\frontpagedate{Submitted: May 31\textsuperscript{st}, 2025}

\title{Near-Infrared Hyperspectral Imaging Applications in Food Analysis}
\subtitle{Improving Algorithms and Methodologies}

\date{Submitted: \fancytoday}
\advisor{Principal supervisor: Kim Steenstrup Pedersen}
\frontpageimage{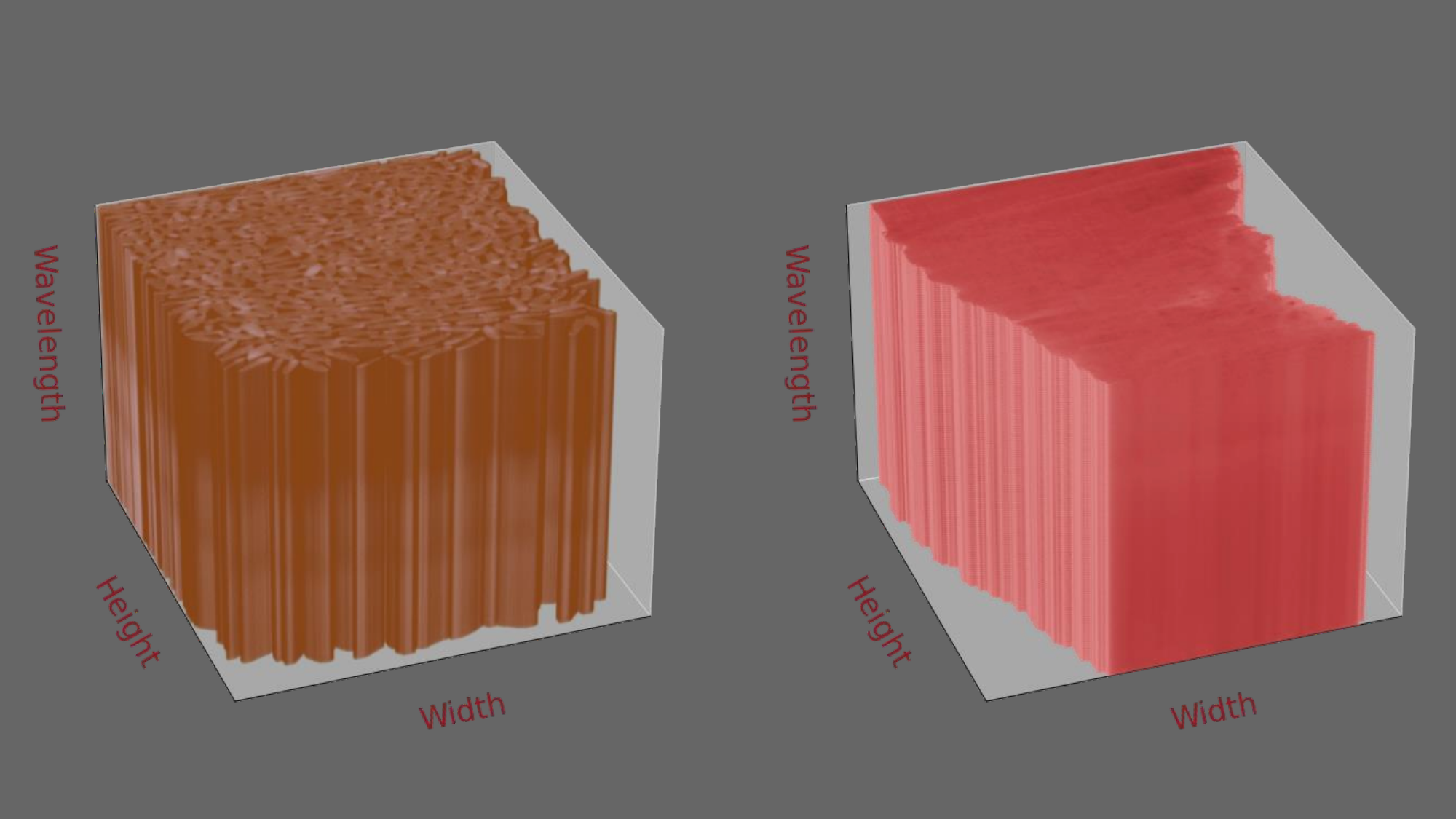}

\kupdfsetup{Near-Infrared Hyperspectral Imaging Applications in Food Analysis - Improving Algorithms and Methodologies}{}{Ole-Christian Galbo Engstr{\o}m}

\begin{document}
\begingroup
  \fontencoding{T1}\fontfamily{LinuxLibertineT-OsF}\selectfont
  \maketitle
\endgroup

\noindent\textbf{Author}\\
Ole-Christian Galbo Engstrøm, PhD Student, MSc\\
FOSS Analytical A/S, Denmark\\
Department of Computer Science, University of Copenhagen, Denmark\\
Department of Food Science, University of Copenhagen, Denmark\\

\noindent\textbf{Principal supervisor}\\
Kim Steenstrup Pedersen, Professor, PhD\\
Department of Computer Science, University of Copenhagen, Denmark\\
Natural History Museum of Denmark, University of Copenhagen, Denmark\\

\noindent\textbf{Industrial supervisor}\\
Erik Schou Dreier, Senior Scientist, PhD\\
FOSS Analytical A/S, Denmark\\

\noindent\textbf{Academic co-supervisor}\\
Birthe Møller Jespersen, Docent, PhD\\
UCL University College, Denmark\footnote{During the first year of this PhD project, Birthe Møller Jespersen was an associate professor at The Department of Food Science, University of Copenhagen, Denmark.}\\

\noindent\textbf{Industrial co-supervisor}\\
Toke Lund-Hansen, Director, PhD\\
FOSS Analytical A/S, Denmark\\
\newpage

\section*{Acknowledgments}
Kim, we first met when I took your course, Robot Lab, as an undergraduate student interested in image analysis and computer vision. I enjoyed your teaching style and personality, so I contacted you again while searching for a supervisor for my bachelor's project. I wanted to do something with deep learning and image analysis. We decided to do a project about implementing the YOLO object detector and using it on the Cityscapes dataset. I had no experience with deep learning, but thanks to your mentorship, I learned how to mathematically model the object detection task. When the time came for my master's project, I knew I wanted you as my supervisor again. Once again, I learned a great deal through your mentorship. This time, we built something new, and based on that work, we wrote an article together. It became my first publication and laid the foundation for my PhD project, where you again agreed to supervise me. Kim, I am extremely grateful for everything you have taught me. It has been incredibly fun and educational. Thank you.\\

\noindent Erik, we met when I was looking for a master's project with Kim as a supervisor. Kim was involved in about ten potential projects, and about half of them were yours concerning your postdoctoral research. You had one project about so-called hyperspectral images, which I had never heard about. It sounded pretty cool, and you seemed both friendly and intelligent. So, I decided to work on your project. It was the single best choice I have ever made. Because of our collaboration, so many doors have opened for me, and this PhD project is a direct consequence. Working with you has been an incredible educational, professional, and personal experience. I can not put a price on your mentorship. Thank you.\\

\noindent Birthe, you have taught me a great deal about food science. You have fantastic knowledge about grain, and I have benefited particularly from your insights into the barley germination process. I have always found our conversations uplifting, encouraging, and insightful. This project would not have been possible without you or your knowledge. I am very grateful that you agreed to supervise me. Thank you.\\

\noindent Toke, you have been my manager for the duration of this project. I have always felt thankful that you let me work my odd hours and allowed me to be wherever I felt best, whether at FOSS, DIKU, or home. At the same time, you made me feel like a valuable employee of FOSS, not only by actively showing interest in my research but by inviting me to give talks about it in front of both specialists and leadership. I am pleased about the way you have acted as my manager. Thank you.\\

\noindent Martin, when I had written the first proof related to fast cross-validation, you offered to proofread it. When you were done, you had come up with an alternative proof. It looked like something that had come straight out of a textbook on algorithms. Immediately, I knew that your expertise could take the work to another level. I am grateful you agreed to work on the project with me, and I am delighted about how it turned out. Thank you.\\

\noindent Puneet and Yamine, you invited me to stay with you at WUR. I am grateful for the hospitality you showed me and the work we did together. It was a great experience. Thank you.\\

\noindent Maria and Michela, thank you for your help with the work on chemical maps. Your insights significantly improved the quality of the work. I am happy that I got to work with you.\\

\noindent I want to thank Thomas Nikolajsen for taking a chance on me by employing me at FOSS Analytical A/S to write a grant proposal for The Innovation Fund Denmark to get funding for this project. At the same time, I want to thank Nicolai Bork for helping me write the grant proposal and Tommy Bysted for signing off on it on behalf of FOSS.\\

\noindent I also want to thank my employer, FOSS Analytical A/S, and The Innovation Fund Denmark for funding this project.\\

\noindent Thanks to all my friends and colleagues at FOSS and DIKU for scientific discussion and social gatherings and for welcoming me into your communities. You made this journey a joyful experience.\\

\noindent Thanks to Uncle Bo for insightful and cozy mathematical sparring sessions.\\

\noindent Thanks to Dad, Mom, Micky, Mads, and Niels for your unwavering support.\\\label{sec:acknowledgments}
\newpage

\section*{Abstract - Dansk}\label{sec:abstract_danish}
\lettrine{D}{enne} afhandling undersøger anvendelsen af nær-infrarød hyperspektral billeddannelse (NIR-HSI) til analyse af fødevarekvalitet. Undersøgelsen gennemføres gennem fire studier, der arbejder med fem forskningshypoteser. Studierne sammenligner modeller baseret på såkaldte convolutional neural networks (CNN’er) og såkaldt partial least squares (PLS) til en række analyser.

Generelt overgår fælles spatio-spektral analyse med CNN’er både rumlig analyse med CNN’er og spektral analyse med PLS, når der modelleres parametre, hvor kemisk og fysisk visuel information er relevant. Ved modellering af kemiske parametre med et todimensionelt (2D) CNN forbedres den prædiktive ydeevne væsentligt, hvis CNN'et udvides med et indledende lag dedikeret til spektral foldning. Dette lag lærer en spektral forbehandling svarende til den, som kemometrikere anvender. PLS-baseret spektral modellering klarer sig dog lige så godt ved analyse af gennemsnitsindholdet af kemiske parametre i fødevarer og anbefales derfor som tilgang.

Grundet de høje omkostninger forbundet med indsamling af referenceværdier for enkelte fødevarer undersøger afhandlingen, om det er muligt at træne modeller med gennemsnitlige referenceværdier fra mængder af fødevarer sammen med et stort antal NIR-HSI-billedudsnit fra disse mængder. Modeleringsresultaterne var præcise for både klassifikation af homogene prøver og regression af heterogene prøver. I sidstnævnte tilfælde skal gennemsnitsprædiktionerne dog justeres additivt og multiplikativt --- dette gælder både CNN'er og PLS-baserede modeller.

Modelering af den rumlige fordeling af kemiske parametre med NIR-HSI er typisk begrænset af mulighederne for at anskaffe rumligt fordelte referenceværdier. Derfor anvendte et studie gennemsnitlige mængdereferenceværdier til at generere kemiske kort over fedtindhold i svinemaver. En PLS-baseret tilgang resulterede i betydelige problemer med ujævne kemiske kort og pixelvise prædiktioner uden for det fysisk mulige område mellem 0–100\% fedtindhold. Omvendt lykkedes det et 2D-CNN, udvidet med det førnævnte lag til spektral foldning, at generere kemiske kort direkte fra det originale NIR-HSI-billede og afhjælpe de problemer, der opstod med PLS.

Det sidste studie forsøgte at modellere bygs spiringsevne ved at analysere NIR-spektre, RGB-billeder og NIR-HSI-billeder. Resultaterne var dog inkonklusive grundet datasættets lave spiringsgrad.

Derudover har denne afhandling ført til udviklingen af to open source Python-pakker. Den første muliggør hurtig PLS-baseret modellering, mens den anden muliggør ekstremt hurtig krydsvalidering af PLS og andre klassiske maskinlæringsmodeller baseret på et nyt banebrydende algoritmisk resultat.
\newpage

\section*{Abstract - English}\label{sec:abstract_english}
\lettrine{T}{his} thesis investigates the application of near-infrared hyperspectral imaging (NIR-HSI) for food quality analysis. The investigation is conducted through four studies operating with five research hypotheses. For several analyses, the studies compare models based on convolutional neural networks (CNNs) and partial least squares (PLS).

Generally, joint spatio-spectral analysis with CNNs outperforms spatial analysis with CNNs and spectral analysis with PLS when modeling parameters where chemical and physical visual information are relevant. When modeling chemical parameters with a 2-dimensional (2D) CNN, augmenting the CNN with an initial layer dedicated to performing spectral convolution significantly enhances its predictive performance by learning a spectral preprocessing similar to that applied by chemometric domain experts. Still, PLS-based spectral modeling performs equally well for analysis of the mean content of chemical parameters in food samples and is the recommended approach.

Due to the expensive nature of obtaining reference values for individual food samples, this thesis investigates the feasibility of training models with bulk mean reference values paired with large quantities of NIR-HSI image crops from said bulk sample. The results were accurate for both the classification of homogeneous samples and the regression of heterogeneous samples. In the latter case, however, a bias and scale correction must be applied to the bulk-wise mean predictions of image crops for both PLS- and CNN-based models.

Modeling the spatial distribution of chemical parameters with NIR-HSI is typically limited by the ability to obtain spatially resolved reference values. Therefore, a study used bulk mean references for chemical map generation of fat content in pork bellies. A PLS-based approach yielded significant issues with non-smooth chemical maps and pixel-wise predictions outside the physically possible range of 0-100\%. Conversely, a 2D CNN augmented with the previously mentioned spectral convolution layer successfully generated chemical maps directly from the input NIR-HSI image and mitigated the issues arising with PLS.

The final study attempted to model barley's germinative capacity by analyzing NIR spectra, RGB images, and NIR-HSI images. However, the results were inconclusive due to the dataset's low degree of germination.

Additionally, this thesis has led to the development of two open-sourced Python packages. The first facilitates fast PLS-based modeling, while the second facilitates \textit{extremely} fast cross-validation of PLS and other classical machine learning models based on a new groundbreaking algorithmic result.
\newpage

\section*{Preface}
\lettrine{T}{his} thesis has been submitted to the PhD School of The Faculty of Science, University of Copenhagen. It is structured as a synopsis of published works, including three peer-reviewed articles, one manuscript currently under review, two non-peer-reviewed technical reports, two software packages, and one dataset. It also contains a chapter describing a previously unpublished study. \mytabref{tab:works} summarizes the works. Additionally, in continuation of Ole-Christian Galbo Engstr{\o}m's master's project, he, Erik Schou Dreier, and Kim Steenstrup Pedersen conducted some preliminary work on adapting convolutional neural networks (CNNs) for protein content regression analysis on hyperspectral images of wheat grain kernels (\citealt{engstrom2021predicting}). That work contains relevant context to this PhD thesis but is not part thereof.

\begin{table}[h]
\centering
\begin{tabular}{@{}llll@{}}
\toprule
Work                                                  & Type             & Peer reviewed & Year \\
\midrule
\cite{engstrom2023improving}                          & Conference workshop & Yes        & 2023 \\
\cite{engstrom2024ikpls}                              & Journal          & Yes           & 2024 \\
\cite{engstrom2025fast}                               & Journal          & Yes           & 2025 \\
\cite{engstrom2025chem}                               & Journal          & Under review  & 2025 \\
\cite{engstrom2023analyzing}                          & Technical report & No            & 2023 \\
\cite{engstrom2025dataset}                            & Technical report & No            & 2025 \\
\nameref{chap:barley_germination}                     & Chapter in this thesis & No            & 2025 \\ 
\href{https://github.com/sm00thix/ikpls}{ikpls}       & Software package & Yes           & 2024 \\
\href{https://github.com/sm00thix/cvmatrix}{cvmatrix} & Software package & No            & 2024 \\
\href{https://doi.org/10.17894/ucph.71c22737-8005-4588-bd74-80bf7b5ac6b4}{Barley Germination Dataset} & Dataset          & No            & 2025 \\
\bottomrule
\end{tabular}
\caption{Works included in this PhD thesis.}
\label{tab:works}
\end{table}

\noindent Ole-Christian Galbo Engstr{\o}m is the first author and primary contributor to all the works included in this thesis. While the research hypotheses were developed with supervisors and co-authors, Ole-Christian Galbo Engstrøm carried out most of the software development, experimental work, data analysis, and writing.
\newpage

\cleardoublepage
\phantomsection 
\addcontentsline{toc}{part}{Contents}
\tableofcontents*

\chapter{Introduction}\label{chap:intro}
\lettrine{N}{ear-infrared} hyperspectral imaging (NIR-HSI) (also known as spectral imaging \citealt{polder2020hype}) platforms can capture NIR-HSI images which, like regular color images, are three-dimensional cuboids with height, width, and depth. Unlike regular color images, a NIR-HSI image's depth dimension contains an entire NIR wavelength spectrum at each spatial pixel, typically with more than 100 uniformly distributed wavelength channels. The NIR spectral dimension is beneficial for food analysis as contents of chemical quality parameters such as protein and fat correlate with light absorption\footnote{The NIR spectra from a NIR-HSI image are reflectance spectra but can be transformed to pseudo-absorbance by negating their logarithm.} in the near-infrared part of the electromagnetic spectrum (\citealt{osborne2006nir, Manley2021}). While the spatial dimensions enable analysis of a food sample's physical and morphological features, their combination with the spectral dimension enables the modeling of the spatial distribution of chemical parameters. Thus, NIR-HSI is especially well-suited for food analysis (\citealt{AMIGO2013343, Ishikawa2021, MEDINAGARCIA2025113994}).


A NIR-HSI platform enables rapid image acquisition. When coupled with, for example, a segmentation or object detection model, individual elements or smaller subsamples can be extracted from the original images to generate large datasets for subsequent analysis quickly. This is particularly useful for powerful deep learning models that require much more data to train than, for example, linear regression models that are otherwise widely used in chemometrics, the multidisciplinary field concerned with the study of extracting chemically relevant information from data (\citealt{wold1995chemometrics}). These large datasets, however, must still be annotated to enable supervised learning. In classical chemometrics, bulk samples are analyzed with spectrometers to get a mean spectrum, which is then matched with a bulk mean reference value, which is easy and cheap to obtain. Conversely, there is generally no cheap or easy way to obtain subsample reference values. Consider, for example, estimating the protein content in grain or the fat content in pork bellies. Here, it is easy to obtain reference values for the mean protein content in a grain batch or the fat content in an entire pork belly. Obtaining reference values for individual grains or a tiny subsection of a pork belly is a much more costly and time-consuming affair. Therefore, it is convenient if the cheap bulk mean references can be paired with the subsamples and still enable accurate modeling.

Another benefit of a NIR-HSI platform is the combination and potential synergies of spectral and spatial information. The expressive power of a spatio-spectral model should be at least the same as that of a strictly spatial or spectral model. In particular, for problems where both spatial and spectral features are relevant, a spatio-spectral model should outperform the alternatives. It may also be the case that a property exists that requires both spatial and spectral information to be modeled accurately. An NIR-HSI image contains vast amounts of data, not all of which may be relevant to a given task. Classical chemometrics relies on dimensionality reduction techniques such as principal component analysis (\citealt{Pearson01111901, hotelling1933analysis}) or partial least squares (PLS, \citealt{wold1966estimation}) in attempts to suppress noise and build models based on the essential information.

Additionally, as NIR spectra are noisy and convoluted signals, manual preprocessing is often applied to extract the underlying chemical information (\citealt{rinnan2009review}). Instead of using these classical methods that require manual tuning of hyperparameters and preprocessing, deep learning models should be able to simultaneously extract the relevant information directly from the NIR-HSI image and learn to reveal the chemical information hidden in the NIR spectra without any manual preprocessing. All these observations and ideas lay the foundation for the following research hypotheses.

\begin{enumerate}
    \item NIR-HSI images can improve the modeling of properties that can already be modeled with spectral or spatial information alone.
    \item Deep learning facilitates the learning of spectral preprocessing that is otherwise manually selected in chemometrics.
    \item Models trained using bulk mean references for subsamples can estimate the bulk references when averaging subsample predictions.
    \item Using bulk mean references, NIR-HSI images can be used to model the spatial distribution of chemical parameters throughout the images.
    \item A food quality parameter exists that can be accurately modeled with neither color (RGB) imaging nor NIR spectroscopy but can be accurately modeled with NIR-HSI.
\end{enumerate}

Four different projects investigate the hypotheses. The first project develops and analyzes models for grain variety classification and protein content regression in wheat. It compares end-to-end deep learning with classical chemometrics and focuses primarily on learnable preprocessing for deep learning models. The second project builds on the first study's results and investigates the implications of using bulk mean references for subsamples. The third project also uses bulk mean references and is concerned with modeling the spatial distribution of fat in pork bellies to generate chemical maps; those are pixel-wise fat predictions across the entire NIR-HSI image. Finally, the fourth project attempts to analyze barley's germinative capacity ahead of the germination process. It compares models relying on RGB images, NIR spectra, and NIR-HSI images for this task.

In addition to those four projects, this thesis presents two projects on fast PLS-based modeling. The first project concerns the open-source implementation of a fast and numerically stable PLS model in Python. The second project is concerned with developing a new algorithm that can be used in unison with, among other models, PLS to significantly speed up cross-validation (\citealt{stone1974cross, hastie2009elements}).

\section{Modeling}

As mentioned at the beginning of this introduction, the structure of NIR-HSI images is similar to that of regular color images. Therefore, the works in this thesis extensively use convolutional neural networks (CNNs) for their analyses. CNNs have, since the CNN AlexNet (\citealt{krizhevsky2012imagenet}) won the 2012 ImageNet Large Scale Visual Recognition Challenge (ILSVRC, \citealt{russakovsky2015imagenet}), been the model of choice for various image analysis tasks including, but not limited to, classification and regression (\citealt{lecun2015deep}). More recently, Vision Transformers (ViTs, \citealt{dosovitskiy2020image}) have gained traction for image analysis and can even surpass the performance of CNNs. However, ViTs require more data to train than CNNs. Therefore, given the relatively small datasets used in this thesis, the works will use CNNs rather than ViTs. At the same time, PLS is a primary tool for the analysis of NIR spectra of food in chemometrics (\citealt{brereton2018chemometrics, sorensen2021nir}). Therefore, all studies in this thesis will compare CNN- and PLS-based models.

\subsection{Partial Least Squares}\label{sec:pls}
A dataset of NIR spectra can be organized into a matrix $\X \in \mathbb{R}^{N \times K}$ where $N$ is the number of spectra used in the analysis, and $K$ is the number of wavelength channels in each spectrum. These can be matched with another matrix $\Y \in \mathbb{R}^{N \times M}$ containing the parameters of interest, where $M$ is the number of parameters associated with each spectrum. If the values in $\Y$ correspond to some chemical parameters, such as, for example, protein or fat, linear regression can be used to model the relationship between absorbance NIR spectra in $\X$ and the chemical parameters in $\Y$. Generally, ordinary least squares (OLS) can be used to model such a relationship. OLS computes the regression matrix, $\B=\left(\XT\X\right)^{-1}\XT\Y$, that minimizes the sum of squared residuals between predicted values, $\X\B$, and actual values, $\Y$.

However, NIR spectra suffer from high multicollinearity in the wavelength channels (\citealt{osborne2006nir}). Additionally, spectrometers can measure thousands of wavelength channels, meaning datasets of NIR spectra often have more wavelengths, $K$, than samples, $N$. In both cases, $\XT\X$ becomes rank deficient, implying it has no inverse, so OLS fails to compute $\B$. Several dimensionality reduction methods can mitigate this. Perhaps the most common dimensionality reduction method is principal component analysis (PCA), where OLS can then subsequently model the relationship from dimensionality-reduced $\X$ to $\Y$, a technique known as principal component regression (PCR, \citealt{hotelling1957relations, kendall1957course}).

In the supervised setting, however, there is generally no guarantee that the variance retained by PCA is relevant when used to predict $\Y$. Here, PLS is a more well-suited alternative. PLS is a dimensionality reduction and regression method, initially developed by \citet{wold1966estimation} with the Nonlinear Iterative Partial Least Squares (NIPALS) algorithm. Despite the name, PLS is a linear method that models a linear relationship between $\X$ and $\Y$. PLS performs a dimensionality reduction of $\X$ and $\Y$ to so-called score spaces. The score spaces are determined to maximize covariance between $\X$ scores and $\Y$ scores while keeping $\X$ scores orthogonal to each other. From this reduced space, a linear regression can be computed from the $\X$-scores to the original $\Y$-values. The corresponding regression coefficients are then projected back to the original $\X$-space to express the regression directly from $\X$ to $\Y$.

PLS was initially designed for regression (PLS-R, \citealt{wold1966estimation, wold2001pls}) where $\Y$ contains continuous values. However, it can also be used for classification  (commonly known as PLS discriminant analysis, PLS-DA, \citealt{sjostrom1986pls, staahle1987partial, barker2003partial, brereton2014partial}) where $\Y$ contains discrete values. When $\Y$ is a vector ($M=1$), PLS is commonly called PLS1. Otherwise, if $\Y$ is a matrix with $M>1$, PLS is commonly called PLS2 (even if $M>2$). When clear from context, this thesis will use the general term PLS instead of PLS1 and PLS2 and will not discriminate between PLS-R and PLS-DA.

\subsection{Convolutional Neural Networks}\label{sec:cnn}
\begin{figure*}[ht]
    \centering
    \includegraphics[width=\textwidth]{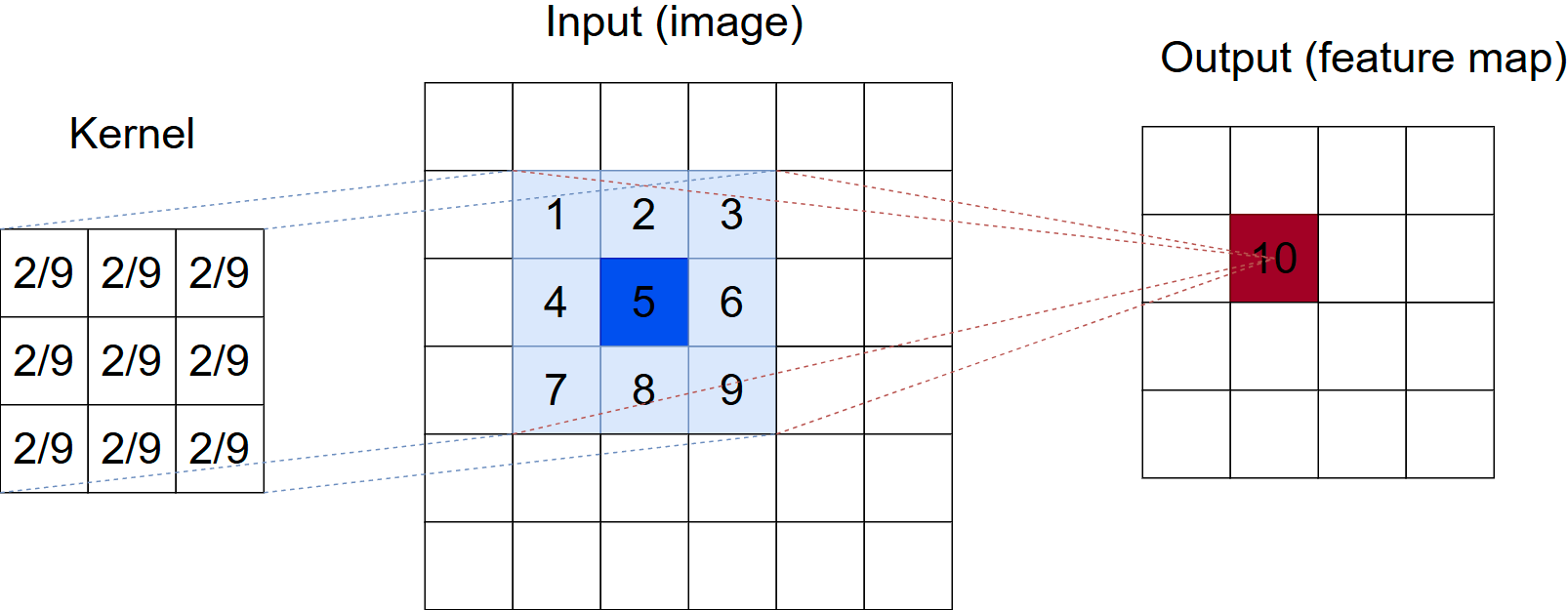}
    \caption{An illustration of the 2D convolution (actually cross-correlation) operation. The kernel is placed with its center on the blue $5$. Element-wise multiplication follows, and the result is summed up and placed at the associated location (illustrated in red) in the output feature map. This kernel computes a feature equal to twice the average of the overlapped part of the input image.}
    \label{fig:convolution}
\end{figure*}

CNNs (\citealt{lecun1989handwritten, lecun2002gradient, lecun2015deep}) are neural networks that perform discrete convolutions between the input data and kernels with learnable weights.\footnote{In practice, cross-correlation is used in place of convolution. This is because convolution is cross-correlation with a flipped kernel. Since the kernel's weights are learned, the flipped kernel might as well be learned directly, circumventing the need for and computational cost associated with flipping the kernel.} The convolution operation strides a typically small kernel across the input image and, at each location, computes the sum of the element-wise multiplication between the kernel and the section of the image with which it overlaps. At each location, the result of the sum, a so-called feature, is stored at the corresponding position in the output array, a so-called feature map. \myfigref{fig:convolution} shows a basic illustration of a two-dimensional (2D) convolution. Convolutions are generalized to any number of dimensions, but CNNs for image analysis typically use the 2D variant. Convolutions are linear operations, and since linearity is closed under composition, they are often combined with non-linear activation functions to facilitate learning non-linear feature maps. While this is the basic idea behind a convolution, there are many additional details when used in the context of CNNs, the explanations of which are out of scope for this thesis.

Perhaps the most popular CNN variant of all time is ResNet (\citealt{he2016deep})\footnote{At the time of writing, the ResNet article has accumulated over 275,000 citations on Google Scholar.}, an ensemble of which was also the first to surpass human performance on ILSVRC. Like many other CNNs, a ResNet is structured as a sequence of convolution layers with interleaving non-linearities forming a feature extractor. This is followed by an average pooling that transforms the extracted feature map to a vector, which is input to a final, fully connected layer that performs linear regression. Thus, a ResNet is, in essence, a convolution-based feature extractor followed by a linear regression. When used for analysis in NIR-HSI, ResNet can extract and combine features across spatial and spectral locations and subsequently perform a regression on these. Combined with an appropriate loss (objective) function, gradient descent can be used to optimize the parameters of the entire ResNet for a given task directly from the input data. This thesis extensively uses ResNet-18, the smallest ResNet, for regression and classification tasks.

\section{Related Work}\label{sec:related_work}
NIR-HSI has seen widespread application in food analysis research in the previous decades (\citealt{GOWEN2007590, AMIGO2013343, WIEME2022156, MEDINAGARCIA2025113994}). To limit the scope of this section, it focuses on work related to the five research hypotheses enumerated earlier in this chapter.

As mentioned, a NIR-HSI image can be analyzed by looking at its spectral or spatial dimensions alone or both in unison. Models trained jointly on both spatial and spectral modalities should be able to perform at least as well as models trained on either modality alone. Indeed, it seems that spatio-spectral modeling may outperform or perform as well as either spatial and spectral modeling in cases where both modalities contain signals relevant to the task (\citealt{dreier2022hyperspectral, Pourdarbani2023, ebrahimi2023harvest}). Other times, however, spectral modeling outperforms joint spatio-spectral modeling (\citealt{engstrom2021predicting}). In particular, chemical parameters may be more challenging to model with joint analysis of the entire NIR-HSI images than spectra. This may seem peculiar, as all the chemical information in the spectrum is also in the NIR-HSI image. However, when using PLS to analyze chemical parameters, it is common to apply some combination of preprocessing that includes smoothing and taking derivatives of the spectral dimension. This reduces the physical effects of light scattering in the spectrum and reveals the underlying chemical information (\citealt{rinnan2009review}). Likewise, when training 1D CNNs for spectral analysis, such preprocessing is not necessarily applied as these can learn to mimic the spectral preprocessing in their first layer and may perform as well or better than preprocessed PLS models (\citealt{acquarelli2017convolutional, cui2018modern, helin2022possible, MISHRA2022116804}).

On the other hand, 2D and 3D CNNs are commonly used to analyze the entire NIR-HSI images (\citealt{zhu2020deep, su2021application, engstrom2021predicting, dreier2022hyperspectral, Pourdarbani2023, ebrahimi2023harvest}). While they can also learn such preprocessing, it requires tuning many more parameters than for 1D CNNs and is thus more challenging to achieve in practice. This phenomenon may contribute to joint spatio-spectral modeling sometimes being outperformed by spectral modeling for chemical analysis. Indeed, our preliminary work shows that adding a specialized spectral convolution layer prior to a 2D ResNet significantly enhances its performance on wheat protein content regression from NIR-HSI images (\citealt{engstrom2021predicting}). We explore and improve this idea in \mychapref{chap:cnn_design}.

Another caveat for CNNs is that they require more training data than simpler models such as PLS. Unfortunately, large NIR-HSI ImageNet-like datasets for food analysis are unavailable (\citealt{MISHRA2022116804}). Therefore, practitioners and researchers must often generate datasets themselves. NIR-HSI images can be divided by, for example, cropping into subsamples to inflate the amount of images in the dataset. However, these subsamples must be accompanied by reference values for supervised learning. These will often be both expensive and time-consuming to obtain. Therefore, a strategy is to take the reference value of the entire NIR-HSI image and assign that to every subsample. This approach is entirely valid for classifying homogeneous images, as the true value for the subsamples will be identical to that of the original image (\citealt{dreier2022hyperspectral}).

On the other hand, if the reference value is a mean of a heterogeneous sample, as it will be for regression of a chemical parameter, it may not be accurate for every subsample. Still, by definition, the errors will average out when considering all subsamples. Thus, regression modeling with such a dataset may yield an accurate model (\citealt{engstrom2021predicting}). However, as we shall see in \mychapref{chap:bulk_references}, this approach introduces subtle yet significant biases in the trained model, regardless of whether it is CNN- or PLS-based.

Another use of NIR-HSI imaging is chemical map generation. As previously mentioned, chemical map generation is the modeling of a chemical parameter's spatial distribution throughout the NIR-HSI image. Here, a model assigns a prediction to every pixel of the NIR-HSI image. For over a decade, researchers handled this task with PLS-models trained on mean spectra with associated reference values (\citealt{B605386F, ELMASRY2013235, KAMRUZZAMAN2012218, BARBIN20131162, KAMRUZZAMAN20168, MA2021103230, FLORIANHUAMAN2022104407, CRUZTIRADO2023109266, HE2023105069, ADESOKAN2024106692, albano2025visible}). As deep learning has become more commonly used in chemometrics (\citealt{MISHRA2022116804}), 1D CNNs have also been applied to this task (\citealt{CAI2024139847, SHI2024140651, XI2025143997}). PLS models and 1D CNNs may seem well-suited for this task as they naturally operate on the pixel-wise level. Indeed, a 2D CNN that consumes the entire NIR-HSI image to make a single prediction, such as, for example, ResNet, can not be used to generate pixel-wise predictions. However, as we shall see in \mychapref{chap:chem_maps}, not considering the context of the entire NIR-HSI may lead to significant faults in the generated chemical maps. Indeed, \citet{BARBIN20131162} note that their PLS model gives reasonable predictions on mean spectra but that the chemical maps generated by the same model are unsatisfactory.

Researching this topic revealed no works that generate chemical maps directly from the input NIR-HSI image and associated mean reference value without relying on 1D models for pixel-wise predictions. However,\citet{SHI2024140651} train a linear regression model on mean spectra for quantification of nutrients in wheat. Then, they train a generative adversarial network (GAN) to generate chemical maps. The GAN requires reference chemical maps for the training process. To the best of Ole-Christian Galbo Engstr{\o}m's understanding, the reference chemical maps are generated by the linear regression model, and, as such, the approach still directly relies on the pixel-wise predictions of a 1D model. Indeed, if this interpretation is correct, then the GAN is directly optimized to predict chemical maps similar to the ones generated by their pixel-wise linear regression model and, as such, is simply a more computationally expensive path to the same result. In any case, their GAN approach requires a reference chemical map, which we circumvent in \mychapref{chap:chem_maps}.

As mentioned in this section, some parameters are best modeled with joint spatio-spectral analysis of entire NIR-HSI images. Generally, however, these parameters can also be modeled fairly accurately using either spatial or spectral modalities alone. It seems challenging to find a parameter for which utilization of the entire NIR-HSI image not only improves performance but is required to get performance surpassing random guessing. \mychapref{chap:barley_germination} argues why barley's germinative capacity may be one such parameter. That is the task of predicting if and when a given barley kernel will germinate after exposure to moisture. This task has not received much attention from the research community. Indeed, \citet{orth2025current} offers a review on the closely related task of pre-harvest germination detection in barley, where the latest study referenced is around two years old and uses mean spectra from NIR-HSI images to classify germinated and ungerminated barley kernels (\citealt{HELMUTORTH2023108742}). The review mentions that both physical and chemical parameters are necessary for accurate detection. Their review shows that analysis through NIR-HSI may be a promising direction, although it is not yet established as a viable alternative to the slow and destructive standard methods. Indeed, \citet{arngren2011analysis} have attempted to model barley germination time. Their accuracy, however, is relatively low for individual kernels, and they must rely on bulk averages and reduced granularity in germination time intervals between classes to achieve better performance. Their study was conducted before CNNs became mainstream for image analysis, so they relied on manual feature engineering to extract both spatial and spectral features for modeling. In \mychapref{chap:barley_germination}, we use CNNs to model barley's germinative capacity.

\section{Structure of Thesis}
The remainder of this thesis is structured in parts. \mypartref{part:fast_pls} contains work related to fast PLS modeling. \mychapref{chap:ikpls} elaborates on the need for a fast and versatile PLS model implementation. \mychapref{chap:fast_cv} shows how it can be improved further when used in a cross-validation setting. \mypartref{part:hsi} contains work related to the five research hypotheses presented earlier. \mychapref{chap:cnn_design} improves 2D ResNet's capability to perform regression analysis on chemical parameters; \mychapref{chap:bulk_references} analyzes the results further and identifies caveats when using bulk references for subsamples in the context of regression analysis. \mychapref{chap:chem_maps} improves chemical map modeling by identifying and addressing the issues arising when generating chemical maps directly from PLS trained on mean spectra. \mychapref{chap:barley_germination} presents a previously unpublished study regarding modeling barley's germinative capacity using both RGB and NIR-HSI image modalities. \mypartref{part:final_remarks} contains a discussion and conclusion upon the previous two parts in the context of the research hypotheses. Finally, \mypartref{part:appendices} consists of appendices with technical reports containing additional information regarding the works in Chapters \ref{chap:bulk_references} and \ref{chap:barley_germination}.

\part{Fast Partial Least Squares}\label{part:fast_pls}
\chapter{Improved Kernel Partial Least Squares}
\lettrine{P}{artial} least squares (PLS) models a linear relationship between $\X \in \mathbb{R}^{N \times K}$ and $\Y \in \mathbb{R}^{N \times M}$ and can be implemented by a plethora of algorithms. For example, $\X$ could be $N$ spectra with $K$ wavelengths, and $\Y$ could be the associated $M$ reference values for each spectrum. In HSI, we are particularly interested in cases where $N > K$ as samples can rapidly be imaged. In contrast, the number of wavelength channels is constant and typically in the hundreds, thus quickly dominated by the number of samples. In chemometrics, when using spectrometers with many wavelength channels to measure smaller datasets, it may be the case that $N < K$. Although not applicable in this thesis, \mychapref{chap:discussion} comments on what to do in this case.

The execution time and numerical stability of the different PLS algorithms vary, as shown by \citet{alin2009comparison} and \citet{andersson2009comparison}. While the original PLS algorithm, NIPALS (\citealt{wold1966estimation}), is the most numerically stable option, it is, unfortunately, also the slowest. When $N > K$, it seems that Improved Kernel PLS (IKPLS, \citealt{dayal1997improved}) is the best choice of PLS algorithm as it is extremely fast and also highly numerically stable. In some cases, however, it can be shown to be numerically unstable for ill-conditioned $\X$ (\citealt{bjorck2017fast}). IKPLS comes in two variants, Algorithm 1 and Algorithm 2. If $N \gg K$, Algorithm 2 is faster than Algorithm 1. \citet{andersson2009comparison} analyzed Algorithm 1 for numerical analysis in his study. However, empirical studies have shown that both algorithms almost always give the same result, as evidenced by the test suite by \citet{engstrom2024ikpls}.

NIPALS, however, is the only algorithm implemented by scikit-learn (\citealt{scikit-learn}), the ubiquitous machine learning library for Python. While forcing users to use the slow NIPALS algorithm, the scikit-learn implementation has several additional inconveniences. First, there are 16 combinations of centering and scaling for $\X$ and $\Y$. scikit-learn's implementation always enforces centering of both $\X$ and $\Y$ while allowing the user to determine whether to scale both $\X$ and $\Y$ or neither but not just one of them. So, scikit-learn allows only two of the 16 combinations.

Additionally, scikit-learn requires a user to input the exact number of components, $A$, and will only output predictions for this exact number of components. However, to get a PLS model with $A$ components, one must first compute the initial $1, \ldots, A-1$ components. This means that when a PLS model with $A$ components has been trained (calibrated), all predictions for that PLS model using one up to and including $A$ components are readily available. Getting all these predictions is required when determining the optimal number of components to use with, e.g., cross-validation. Getting all $A$ predictions from scikit-learn's implementation is not impossible but requires extensive knowledge and hacking of their implementation, which will still be slower than either IKPLS algorithm. Thus, to make PLS modeling convenient and straightforward for our studies and other practitioners, we developed and open-sourced IKPLS Algorithm 1 and Algorithm 2 in Python (\citealt{engstrom2024ikpls}).

\citet{dayal1997improved} have made an error, or perhaps a typo, in step 2 of their description of the IKPLS algorithm. They write that if $M < K$, $\mathbf{q}$ can be computed as the eigenvector corresponding to the largest eigenvalue of $\YT\XT\X\Y$. In reality, the matrix product should be $\YT\X\XT\Y$ and is correctly implemented by \citet{engstrom2024ikpls}. While the latter is the correct matrix product, it is likely also what \citet{dayal1997improved} intended to write as the former matrix product is generally undefined due to a dimensionality mismatch unless $\X$ is a square matrix, that is, $N=K$.

Later in this PhD project, after having implemented the original IKPLS formulation by \citet{dayal1997improved}, PLS had to be used for the imbalanced barley germination dataset that was curated for this PhD thesis (\citealt{engstrom2025dataset}). To address the class imbalance, the original IKPLS implementation was augmented to accommodate sample weighting as described by \citet{becker2016accounting}. When using weighted PLS and scaling is desired, it should be carried out with the weighted sample standard deviation, which is the square root of the weighted sample variance. In their Equation 4, \citet{becker2016accounting} define the weighted sample variance row vector of $\X$ as

\begin{equation}\label{eq:becker_ismail_wvar}
    \mathbf{\sigma}_{\w}{^2} = \frac{\sum_{i=1}^{N}\w_i\left(\X_i - \mathbf{\mu}_\w\right)^{\circ2}}{\left(\sum_{i=1}^{N}\w_i\right) - 1} .
\end{equation}

Here, $\w$ is a length $N$ vector of sample weights where $\w_i\geq 0$ for all $i \in \{1,\ldots, N\}$, $\mathbf{\mu}_\w$ is the weighted mean row vector of $\X$, and $\circ$ denotes Hadamard (element-wise) exponentiation. However, this formulation is not well defined in general. Consider, for example, the case where the weights are normalized to sum to one, $\sum_{i=1}^{N}\w_i=1$. Then, the denominator in the above expression becomes zero, leading to an undefined weighted sample variance.

Therefore, the IKPLS implementation (\citealt{engstrom2024ikpls}) instead uses the NIST formulation for weighted sample variance (\citealt{nist-weighvar}) and its square root, weighted sample standard deviation (\citealt{nist-weightsd}). The NIST weighted sample variance is given by

\begin{equation}\label{eq:nist_wvar}
    \mathbf{\sigma}^{2}_\w=\frac{N'\sum_{i=1}^{N}\w_i\left(\X_i - \mathbf{\mu}_\w\right)^{\circ2}}{\left(N'-1\right)\sum_{i=1}^{N}\w_i},
\end{equation}

where $N'$ is the number of non-zero weights. Here, scaling the weights by a constant factor does not change the weighted sample variance; therefore, this expression works seamlessly with weights normalized to one. Note, that if $\sum_{i=1}^{N}\w_i=N'$, then \myeqnref{eq:becker_ismail_wvar} and \myeqnref{eq:nist_wvar} are equal. This will, for example, be the case when using balanced class and sample weights from scikit-learn (\citealt{scikit-learn}). These weights are inspired by \citet{king2001logistic} and are the only source of class and sample weights in this thesis.

What follows is the article by \citet{engstrom2024ikpls} about the Python software package \href{https://github.com/sm00thix/ikpls}{ikpls}\footnote{https://github.com/sm00thix/ikpls}. While the article is peer-reviewed, most of the review process focused on reviewing the source code. The code review is linked in the article and \href{https://github.com/openjournals/joss-reviews/issues/6533}{here}\footnote{https://github.com/openjournals/joss-reviews/issues/6533} for convenience.

\includepdf[pages=-]{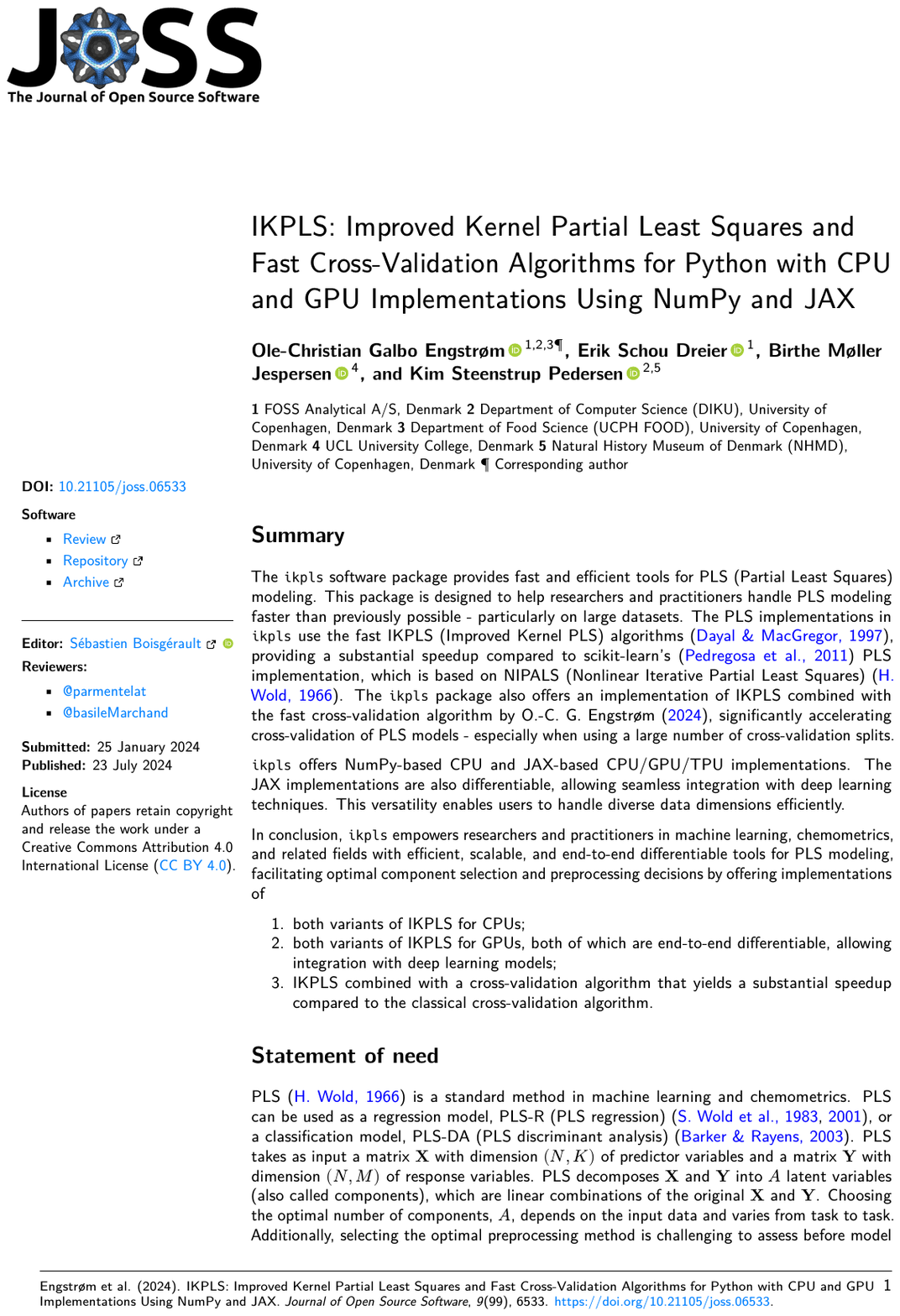}\label{chap:ikpls}
\chapter{Fast Cross-Validation}
\lettrine{I}{n} NIR-HSI, it is often the case that $N \gg K$ and $N \gg M$, that is, $\X$ and $\Y$ have significantly more rows than columns, that is, there are more spectra than both wavelength channels and reference values per spectrum. Here, IKPLS Algorithm 2 is the fastest choice for the PLS model, and most of the computational cost is due to the computation of $\XT\X$ and $\XT\Y$. These matrix products are computed initially, and the remainder of the PLS fit uses these two matrix products without looking at $\X$ and $\Y$ ever again. This implies that the only part of the PLS fit with a runtime complexity dependent on $N$ is the computation of $\XT\X$ and $\XT\Y$. Thus, if the computation of $\XT\X$ and $\XT\Y$ can be sped up, the total computation time can be drastically improved.

During cross-validation, $\XT\X$ and $\XT\Y$ must be computed independently for each training partition so as not to introduce data leakage from the corresponding validation partitions. However, as training partitions overlap significantly between the rows of $\X$ and $\Y$, they also overlap significantly in the computations required to compute training partition-wise $\XT\X$ and $\XT\Y$. For $P$ partitions (that is, $P$-fold cross-validation), the computational complexity of computing all training partition-wise $\XT\X$ and $\XT\Y$ is $\Theta\left(PNK(K+M)\right)$. \citet{engstrom2025fast} devise algorithms that address and eliminate the redundant calculations and decrease this computational complexity by a factor of $P$ to $\Theta\left(NK(K+M)\right)$ for all 16 combinations of centering and scaling of $\X$ and $\Y$. This means that all $P$ training partition-wise $\XT\X$ and $\XT\Y$ can be computed in the same time as computing the total $\XT\X$ and $\XT\Y$ once. We achieve this speed-up without increasing the space complexity. Combining these algorithms with IKPLS Algorithm 2 makes cross-validation of PLS models on datasets with $N \gg K$ \textit{extremely} fast. This is because the entire training partition-wise $\XT\X$ and $\XT\Y$ need not be recomputed for each training partition, and these computations would otherwise dominate the total runtime for the PLS cross-validation as shown by \citet{engstrom2024ikpls} and devised in the following article by \cite{engstrom2025fast}.

\includepdf[pages=-]{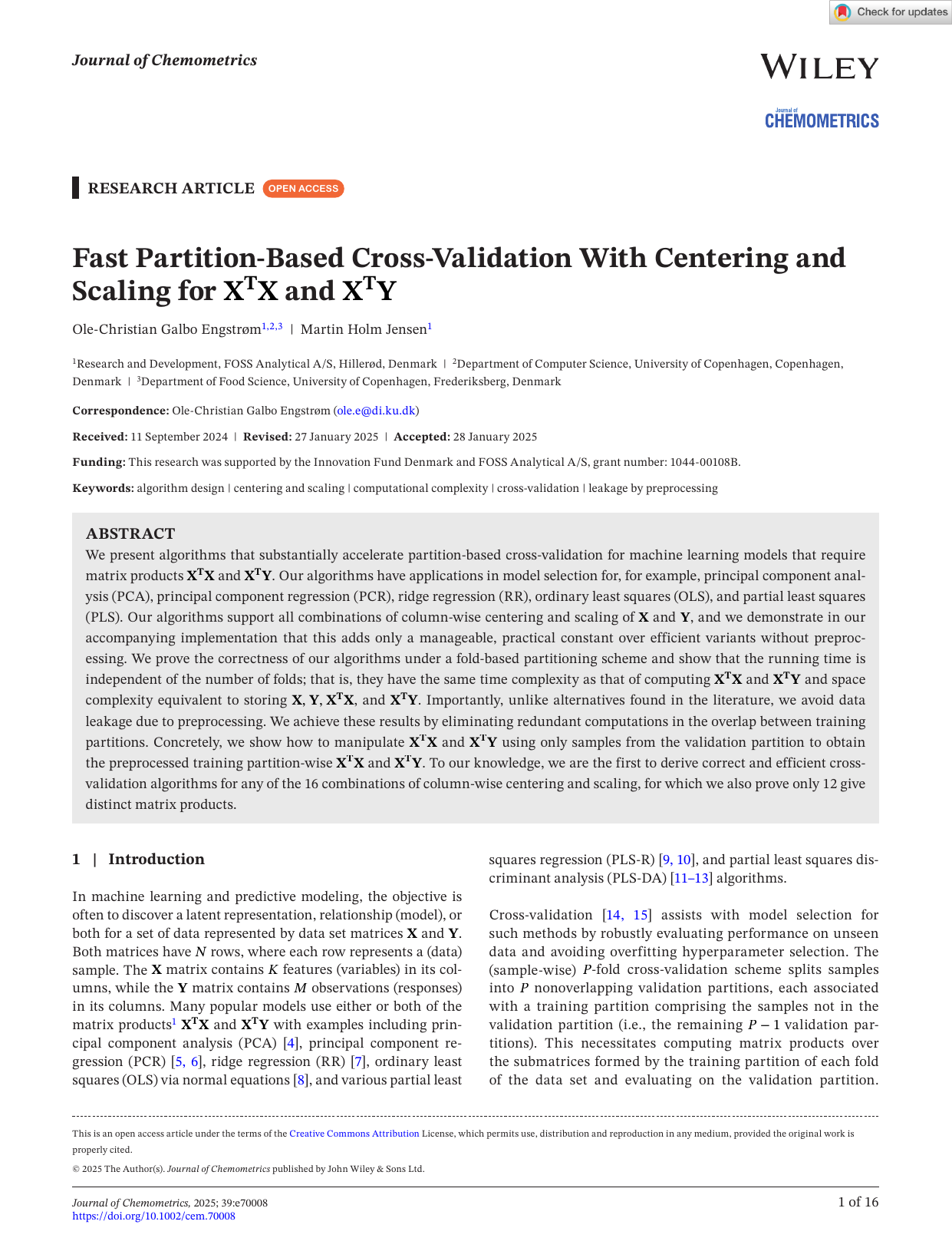}\label{chap:fast_cv}

\part{Near-Infrared Hyperspectral Image Analysis}\label{part:hsi}
\chapter{Convolutional Neural Networks for Near-Infrared Hyperspectral Image Analysis}\label{chap:cnn_design}
\lettrine{M}{odeling} a grain quality parameter with a convolutional neural network (CNN) applied to near-infrared hyperspectral images (HSI) is seemingly straightforward. A CNN, such as the common ResNet (\citealt{he2016deep}), can readily be applied to grain variety classification and protein content regression. While this works well for grain variety classification, it yields subpar results for protein content regression. In chemometrics, NIR spectra are often preprocessed to remove the physical effects of Rayleigh scattering and thus reveal the underlying chemical information of interest. One common way to do so (\citealt{rinnan2009review}) is to convolve the spectra with a Savitzky-Golay (SG) filter (\citealt{savitzky1964smoothing, steinier1972smoothing}). Parameterized by their window size, polynomial, and derivative order, SG filters can apply smoothing and take derivatives. With PLS, conducting a grid search cross-validation over many settings is feasible to determine the best one for a given task. Grid search cross-validation, however, quickly becomes computationally infeasible for CNN-based models.

Instead, by choosing only the filter size (analogous to SG window length), the CNN can learn the parameters for an initial spectral convolution layer during its training process. Learning a handful of parameters for a spectral convolution layer facilitates smoothing and the computation of derivatives. Learning the same with only 2D convolutions would require many filters with a very particular set of weights, making it much more difficult in practice. \citet{engstrom2023improving} show that adding an initial spectral convolution layer greatly increases CNN model performance for regression of protein content on NIR hyperspectral images. Ablation studies show that the spectral convolution layer is much more important than the choice of the CNN base model. In particular, a simple, shallow CNN with an initial spectral convolution layer will significantly outperform a ResNet-18 without an initial spectral convolution layer. Conversely, adding a spectral convolution filter does not increase grain variety classification performance for CNNs. Interestingly, this pattern repeats for PLS modeling where preprocessing with SG filtering decreases subsequent classification accuracy. This indicates that the physical phenomena in the NIR spectra are generally useful discriminators for grain variety classification but noise in the context of protein content regression. These studies show that insights from chemometrics typically exploited to increase PLS performance can also be used to improve the performance of deep learning models for hyperspectral image analysis of food.

Another insight due to \citet{engstrom2023improving} is that grain variety classification can be accurately modeled with the spatial and spectral dimensions alone. However, using the entire hyperspectral image enables even more accurate classification. This indicates that it is beneficial to analyze the entire hyperspectral image for problems where both spatial and spectral information are relevant instead of just its spatial or spectral dimensions alone.

There is a mistake in Figure 2 in the following article by \citet{engstrom2023improving}. The 3D convolution kernels do not have size $\{1,7\} \times 7 \times 7$ but $s \times s \times 7$ with $s \in \{1,7\}$.

\includepdf[pages=-]{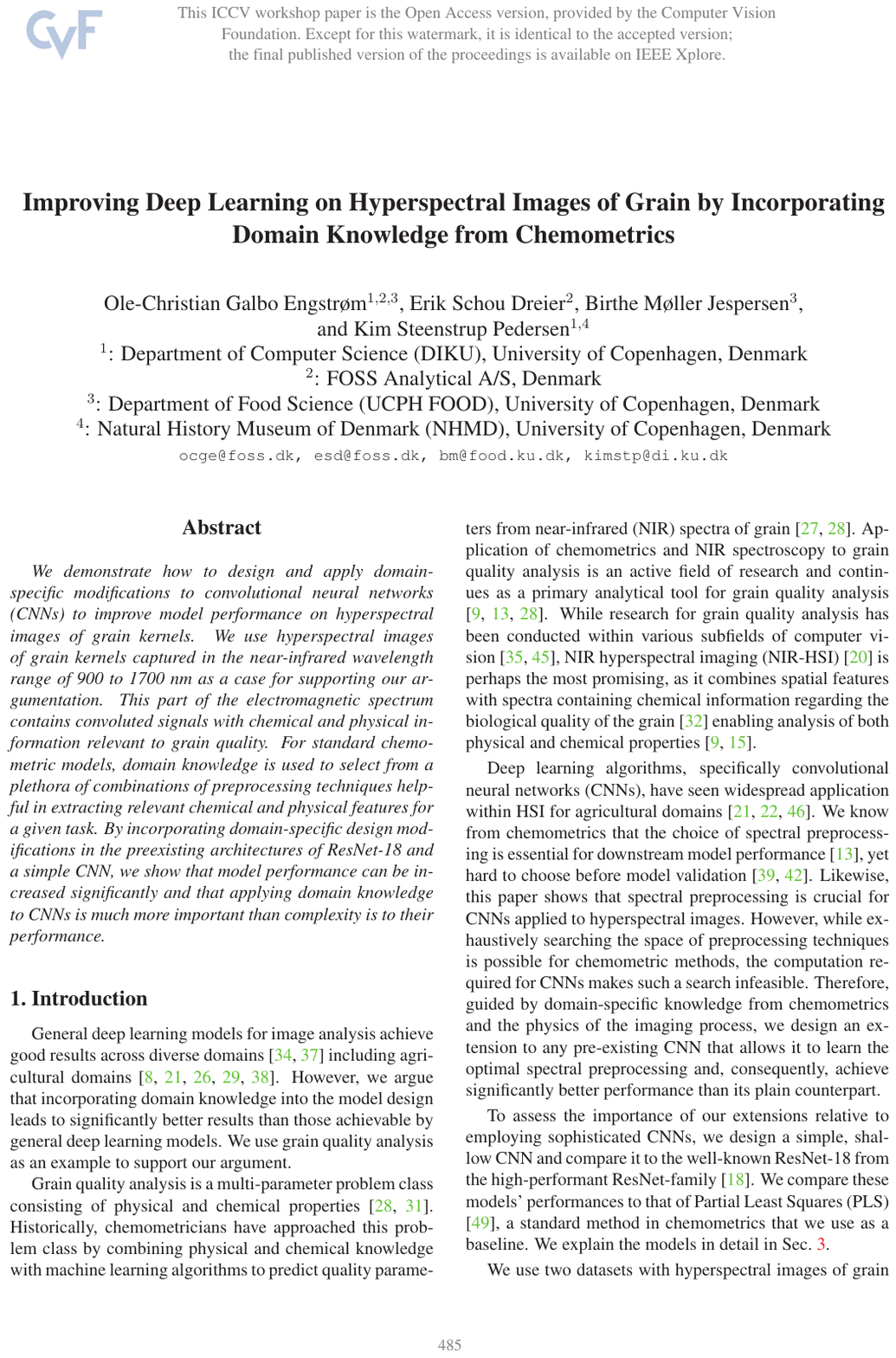}
\chapter{Bulk References for Subsample Predictions}\label{chap:bulk_references}
\lettrine{I}{n} the previous chapter, \citet{engstrom2023improving} trained models for grain variety classification and protein content regression by using bulk reference values as ground truths for image crops containing only a small subsample of the bulk. In addition to evaluating the performance of subsample (image crop) protein content predictions, we are interested in evaluating the mean prediction per bulk to compare with the bulk reference values. Interestingly, a bias and scale difference between the predictions and the references arises when evaluating performance on the mean bulk predictions. This means that bulks with a low reference value have their protein content overestimated, and those with a high reference value have their protein content underestimated. This phenomenon arises for both PLS and CNN predictions. A simple bias and scale correction corrects these errors. We also looked into the predictive performance as a function of the grain density (grain-to-background ratio) in image crops for both protein content regression and grain variety classification. We found that a higher density correlates with better performance for both tasks.

This chapter is written as a condensed version of \citet{engstrom2023analyzing} with some additional insights that arose after publishing the report, which is reprinted in its entirety in \myappref{app:bulk_ref}.

\section{Introduction}
\citealt{engstrom2023analyzing} analyzed two datasets of NIR hyperspectral images of grain. Dataset \#1 is from \citet{engstrom2021predicting} and contains bulk samples of wheat grain kernels with an associated mean protein reference value for each bulk. Dataset \#2 is from \citet{dreier2022hyperspectral} and contains bulk samples of eight different grain varieties, seven wheat, and one rye. A bulk sample yielded approximately ten images for both datasets, each containing many individual grain kernels. In an attempt to generate enough image data to train a CNN, \citet{engstrom2023improving} inflated the datasets by cropping the images (see, e.g., Figure 1 by \citealt{engstrom2023analyzing}), generating on the order of a thousand image crops per bulk sample. Subsequently, each image crop was assigned the same reference value as the bulk from which it originated. In Dataset \#2, the bulks and images are homogeneous, meaning that each contains grain kernels from exactly one of the eight varieties. Therefore, the reference value of the bulk is equally valid as a reference for the subsample. In contrast, the bulk mean protein references in Dataset \#1 are unlikely to be accurate for each image crop as some variation within the bulk is to be expected. However, by definition, the bulk reference will be correct for the image crops on average.

Despite this, averaging the image crop predictions for a bulk does not accurately estimate the bulk's protein content. \citet{engstrom2023analyzing} analyze why this happens and propose a simple bias and scale correction for obtaining accurate bulk protein content estimates.\mysecref{sec:bulk_and_subsample} elaborates on this analysis. It will also show that we were wrong in our hypothesis about why there is a discrepancy between the least squares solution for image crop predictions and bulk mean predictions. However, the bias and scale corrections are still valid and necessary. \mysecref{sec:grain_density} shows the relationship between grain density and the predictive performance of grain variety classifiers and protein content regressors due to \citet{engstrom2023analyzing}.

\section{Bulk Mean and Subsample Prediction Discrepancy}\label{sec:bulk_and_subsample}
\begin{figure*}[ht]
    \centering
    \begin{subfigure}[t]{0.48\textwidth}
        \centering
        \includegraphics[width=\textwidth]{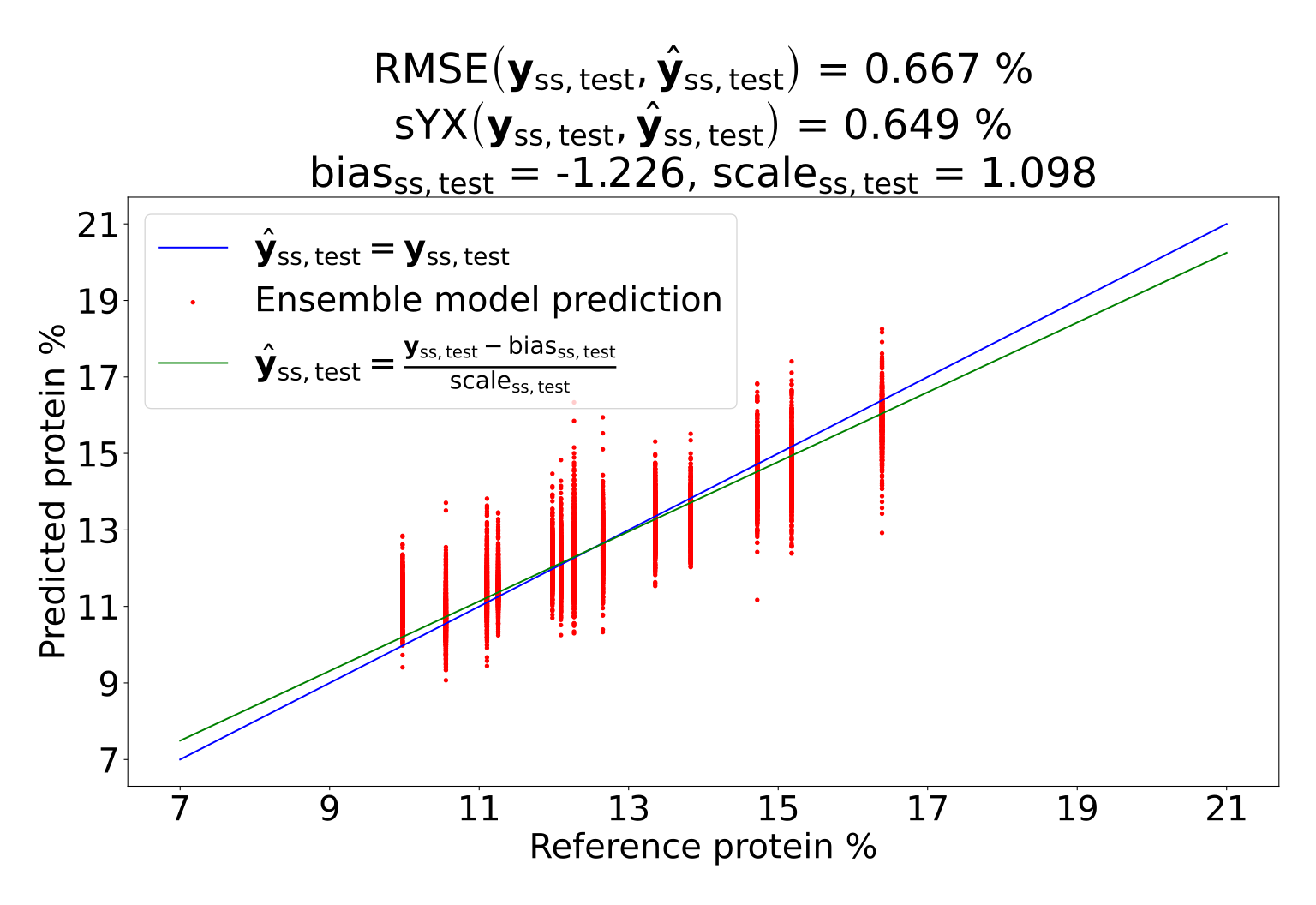}
        \caption{}
        \label{fig:pls_crop_dist}
    \end{subfigure}
    \hfill
    \begin{subfigure}[t]{0.48\textwidth}
        \centering
        \includegraphics[width=\textwidth]{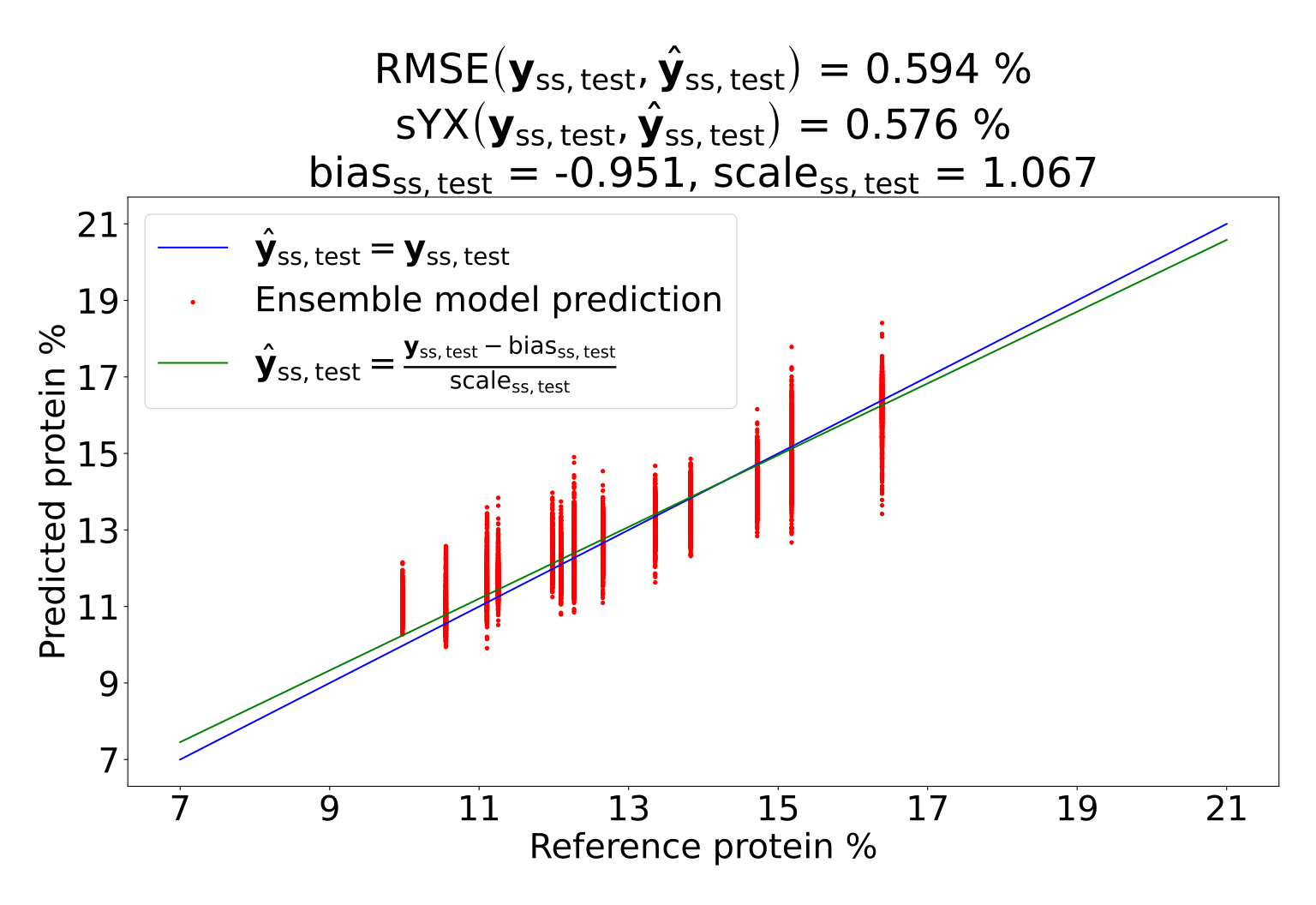}
        \caption{}
        \label{fig:resnet_crop_dist}
    \end{subfigure}
    \caption{Scatter plots of the predicted protein content distribution for each of the 13 bulk samples in the test set. (a) PLS. (b) ResNet-18. This figure is copied from \citet{engstrom2023analyzing}.}
    \label{fig:crop_dist}
\end{figure*}

In our work (\citealt{engstrom2023improving}), we determined the optimal preprocessing for PLS and the optimal modification of ResNet-18 for protein content regression. Then, we constructed an ensemble of each model type for final evaluation on the test set. \myfigref{fig:crop_dist} shows predictions for each image crop (subsample) made by the ensembles. Here, $\y$ is a vector of reference values, and $\hat{\y}$ is a vector of predictions. The subscripts \textsubscript{ss} and \textsubscript{test} mean subsample and test set, respectively. For example, $\hat{\y}_{\text{ss,test}}$ is the vector of subsample predictions for the test set, and $\y_{\text{ss,test}}$ are the associated reference values. sYX is the RMSE obtained after using ordinary least squares (OLS) to adjust the predictions to match the references best. Due to the estimation of bias and scale, sYX has two fewer degrees of freedom than ordinary RMSE. The estimated bias and scale that best adjust $\hat{\y}_{\text{ss,test}}$ to $\y_{\text{ss,test}}$ are given by $\text{bias}_{\text{ss,test}}$ and $\text{scale}_{\text{ss,test}}$. Note that because the predictions are on the vertical axis and the references are on the horizontal axis, the least squares solution minimizes the squared horizontal residuals. This corresponds to finding the linear transformation that must be applied to the predicted values to minimize RMSE between them and the reference values. In \myfigref{fig:crop_dist}, the RMSE quite closely matches the sYX. For the training sets, the lines of best fit almost perfectly match the identity (see Figure 5 in \myappref{app:bulk_ref}), indicating that the models are well-calibrated, meaning they have found a least squares solution.

\begin{figure*}[ht]
    \centering
    \begin{subfigure}[t]{0.48\textwidth}
        \centering
        \includegraphics[width=\textwidth]{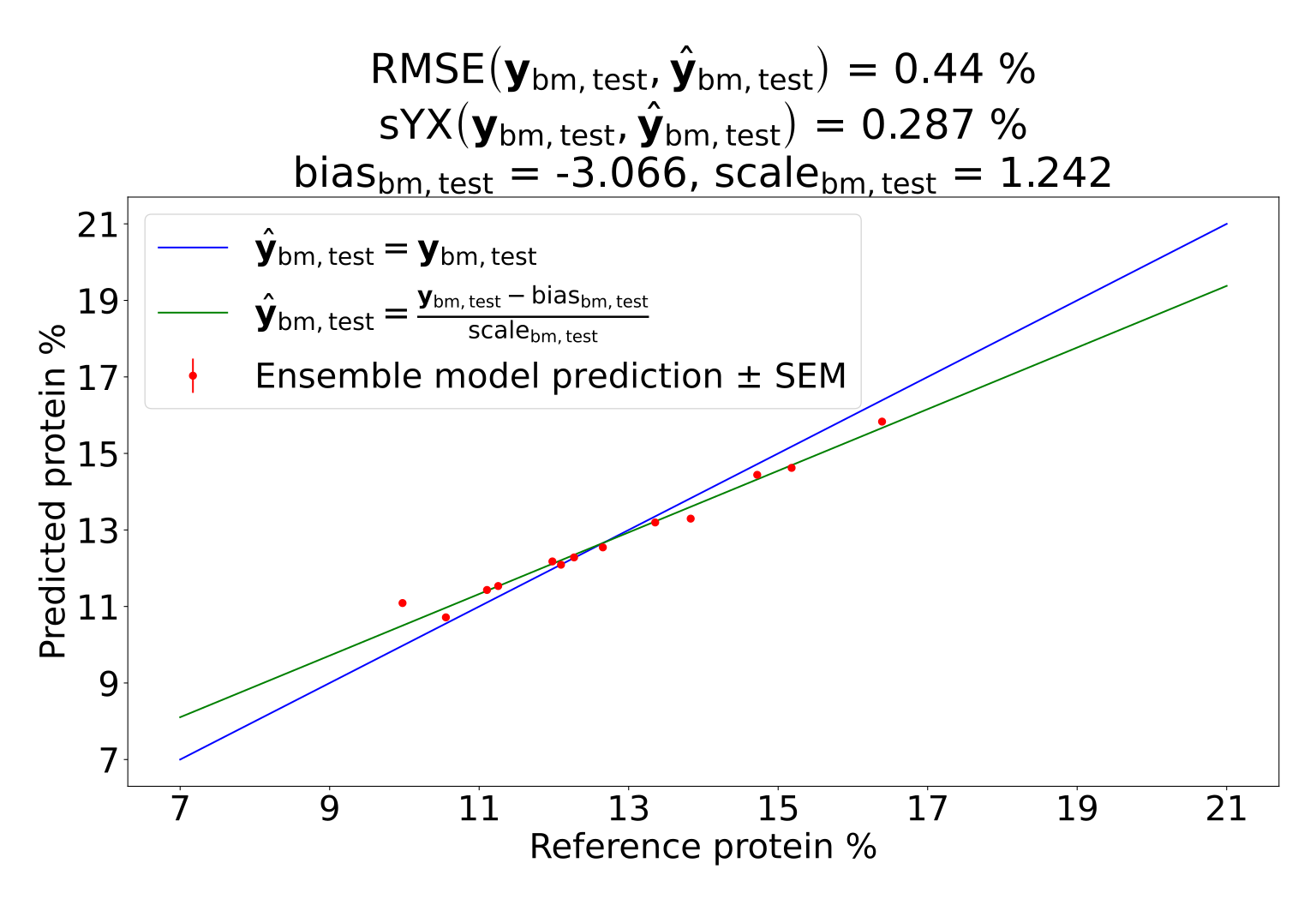}
        \caption{}
        \label{fig:pls_bulk_pred}
    \end{subfigure}
    \hfill
    \begin{subfigure}[t]{0.48\textwidth}
        \centering
        \includegraphics[width=\textwidth]{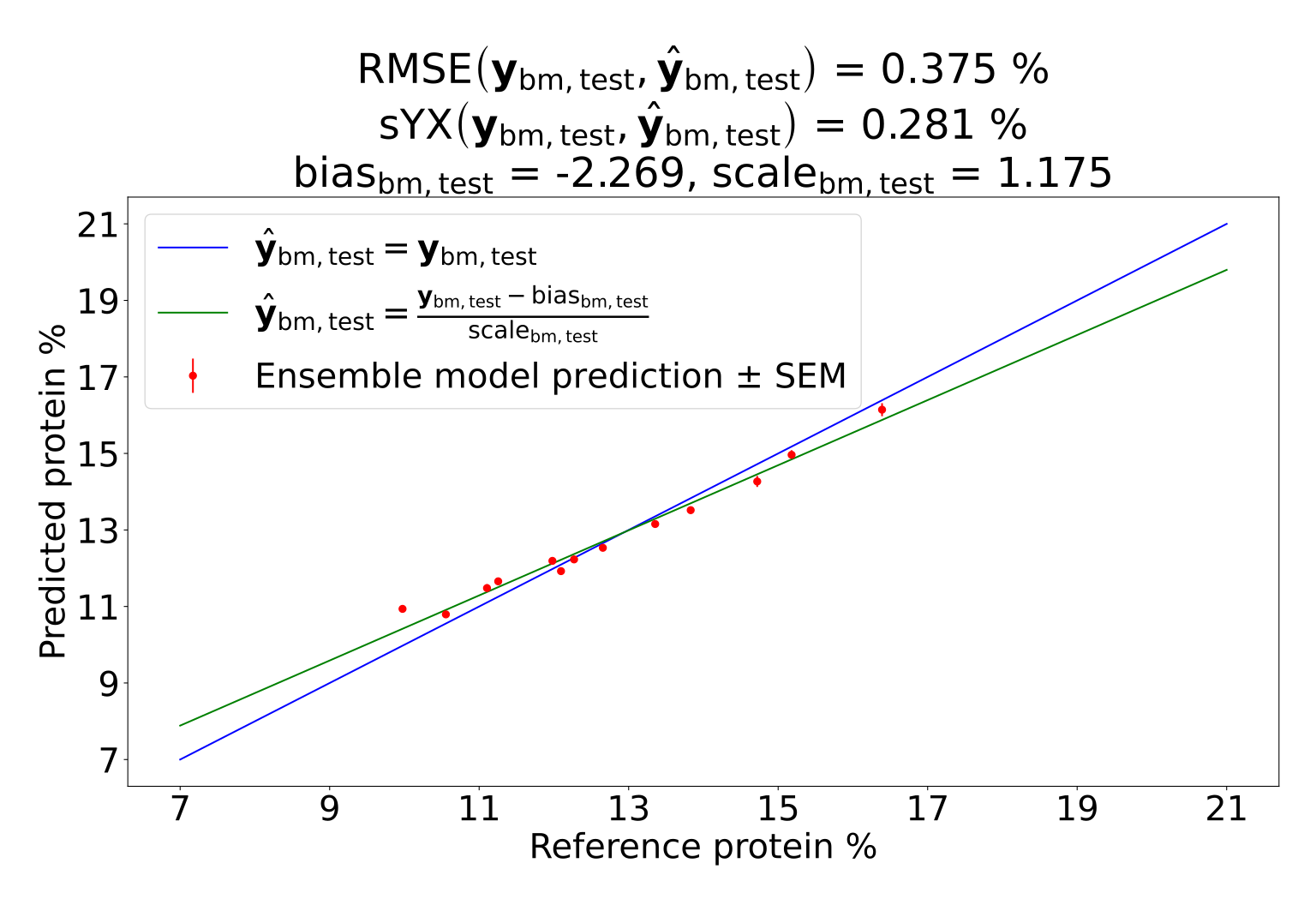}
        \caption{}
        \label{fig:resnet_bulk_pred}
    \end{subfigure}
    \caption{Scatter plots of the means of the predicted protein content distribution for each of the 13 bulk samples in the test set. (a) PLS. (b) ResNet-18. This figure is copied from \citet{engstrom2023analyzing}.}
    \label{fig:bulk_preds}
\end{figure*}

To assess the correctness of the predictions, we can compare the mean prediction for each bulk with its reference. The bulk mean predictions corresponding to the subsample predictions from \myfigref{fig:crop_dist} are shown in \myfigref{fig:bulk_preds}. The subscript $_\text{bm}$ stands for bulk mean. Here, the RMSE and sYX differ significantly, and a clear bias and scale correction are needed to get well-calibrated predictions. The same phenomenon arises for training and validation set predictions (see Figure 4 in \myappref{app:bulk_ref}). It seems always to be the case that the models overestimate the mean protein content for bulks with low reference values and underestimate the mean protein content for bulks with high reference values.

\begin{figure*}[ht]
    \centering
    \begin{subfigure}[t]{0.48\textwidth}
        \centering
        \includegraphics[width=\textwidth]{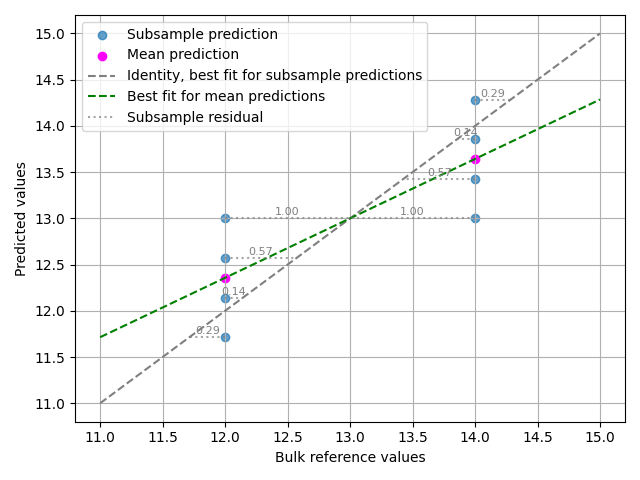}
        \caption{}
        \label{fig:bulk_pred_simulation_subsample_calirated}
    \end{subfigure}
    \hfill
    \begin{subfigure}[t]{0.48\textwidth}
        \centering
        \includegraphics[width=\textwidth]{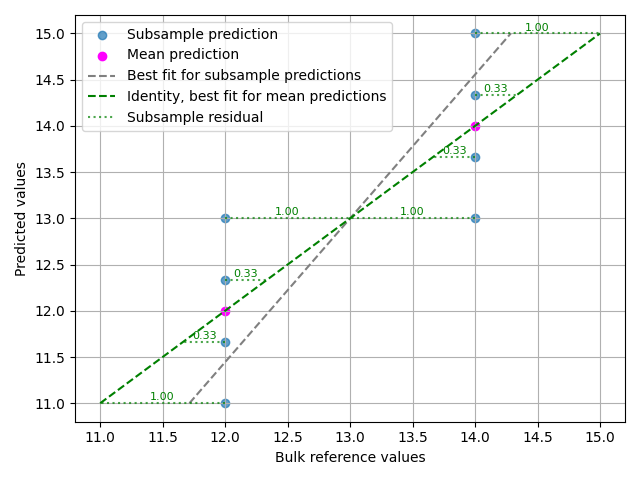}
        \caption{}
        \label{fig:bulk_pred_simulation_mean_calibrated}
    \end{subfigure}
    \caption{Simulation of symmetric predicted subsample distributions for two bulks. The plots show the horizontal residuals from the subsample predictions to the identity. (a) The identity is the line of best fit for the subsample predictions, indicating a well-calibrated model minimizing subsample RMSE. (b) Scale ($1.56$) and bias ($-7.22$) corrected predicted values from (a) based on the best fit for the mean predictions. Thus, the identity becomes the line of best fit for the mean predictions, indicating a well-calibrated model minimizing bulk mean RMSE.}
    \label{fig:bulk_pred_simulation}
\end{figure*}

In the technical report by \citealt{engstrom2023analyzing}, we claim that the discrepancy between the well-calibrated predictions for subsamples and the inaccurate bulk mean predictions is due to non-symmetric predicted subsample distributions for each bulk. In particular, we claim that, given the predicted subsample distributions for each bulk are symmetric, the least squares solution for adjusting $\hat{\y}_{\text{ss}}$ to $\y_{\text{ss}}$ is the same as that for adjusting $\hat{\y}_{\text{bm}}$ to $\y_{\text{bm}}$. Indeed, the predicted subsample distributions are asymmetric (\citealt{engstrom2023analyzing}). However, even if the predicted subsample distributions for each bulk are symmetric and well-calibrated, the mean prediction for bulks with low reference values will be overestimated, and the mean prediction for bulks with high reference values will be underestimated. Consider, for example, the simulation shown in \myfigref{fig:bulk_pred_simulation}. It shows predictions for two bulk samples, each with four subsamples. The two subsample prediction distributions are symmetric. In \myfigref{fig:bulk_pred_simulation_subsample_calirated}, the line of best fit for the subsample predictions is the identity. However, the line of best fit for the mean predictions is not the identity. To transform the predictions, such that the identity becomes the line of best fit for the mean predictions, the predicted values must be multiplied by a scale of $1.56$ and added with a bias of $-7.22$ as shown in \myfigref{fig:bulk_pred_simulation_mean_calibrated}.

When we trained our models to minimize RMSE, we trained to minimize the RMSE for subsamples, corresponding to \myfigref{fig:bulk_pred_simulation_subsample_calirated}. In practice, however, we want to minimize the RMSE for mean predictions, corresponding to \myfigref{fig:bulk_pred_simulation_mean_calibrated}. It is evident that these two problems, although deceptively similar, do not share a common solution. The underlying reason for this phenomenon is still unknown when writing this thesis. Still, it is straightforward to compensate for this discrepancy with a simple bias and scale correction as shown by \citet{engstrom2023analyzing}. When performing this correction, the bias and scale parameters must be chosen independently from the test set so as not to introduce data leakage. Thus, we can estimate the parameters from the training or validation set. After having conducted this correction in our work (\citealt{engstrom2023analyzing}), we became aware that we could also use the union of the training and validation set, analogous to what \citet{igel2023remember} suggests for bias and scale correction of deep learning regressors.

\begin{figure*}[ht]
    \centering
    \begin{subfigure}[t]{0.48\textwidth}
        \centering
        \includegraphics[width=\textwidth]{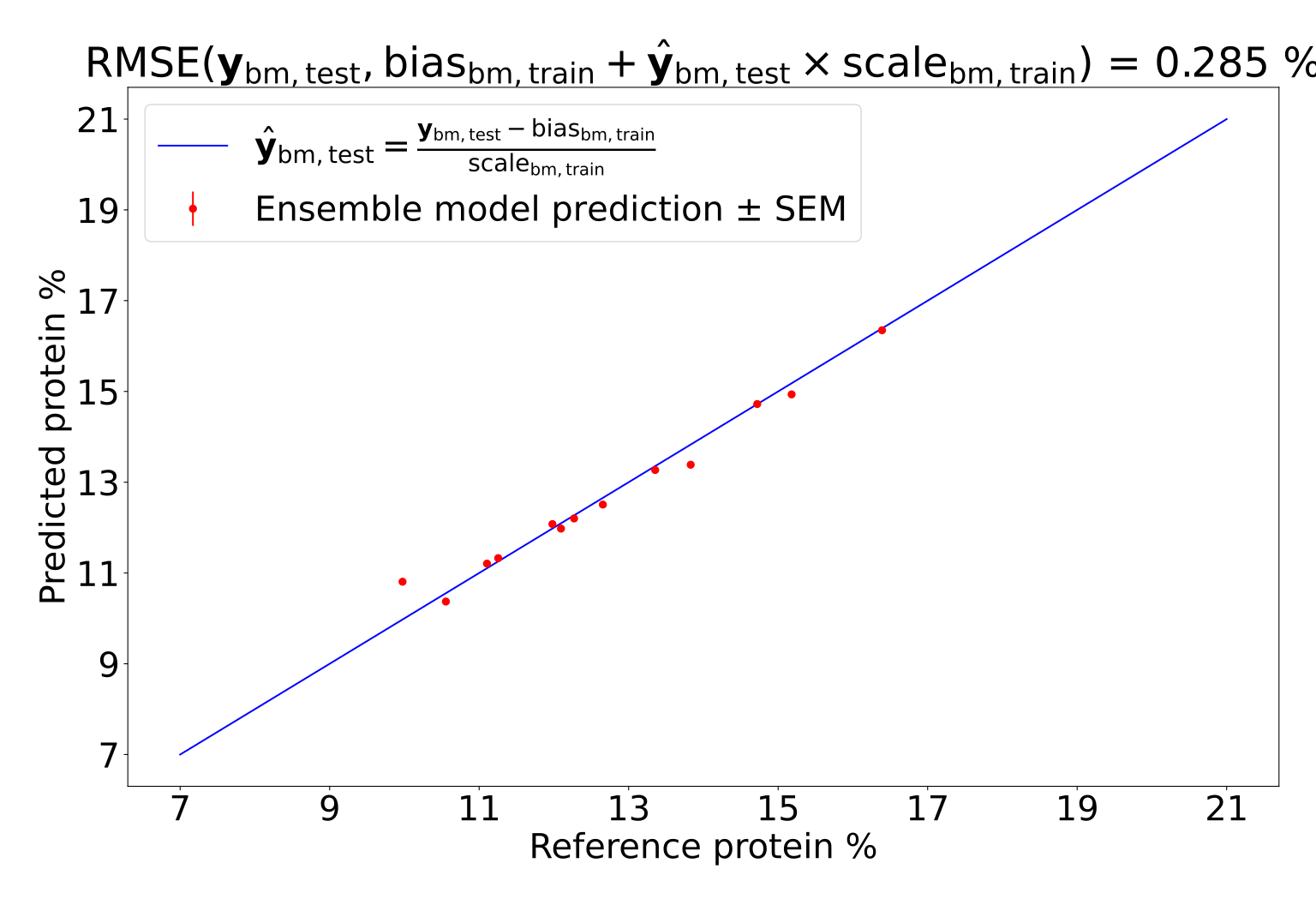}
        \caption{}
        \label{fig:pls_corrected_bulk_pred}
    \end{subfigure}
    \hfill
    \begin{subfigure}[t]{0.48\textwidth}
        \centering
        \includegraphics[width=\textwidth]{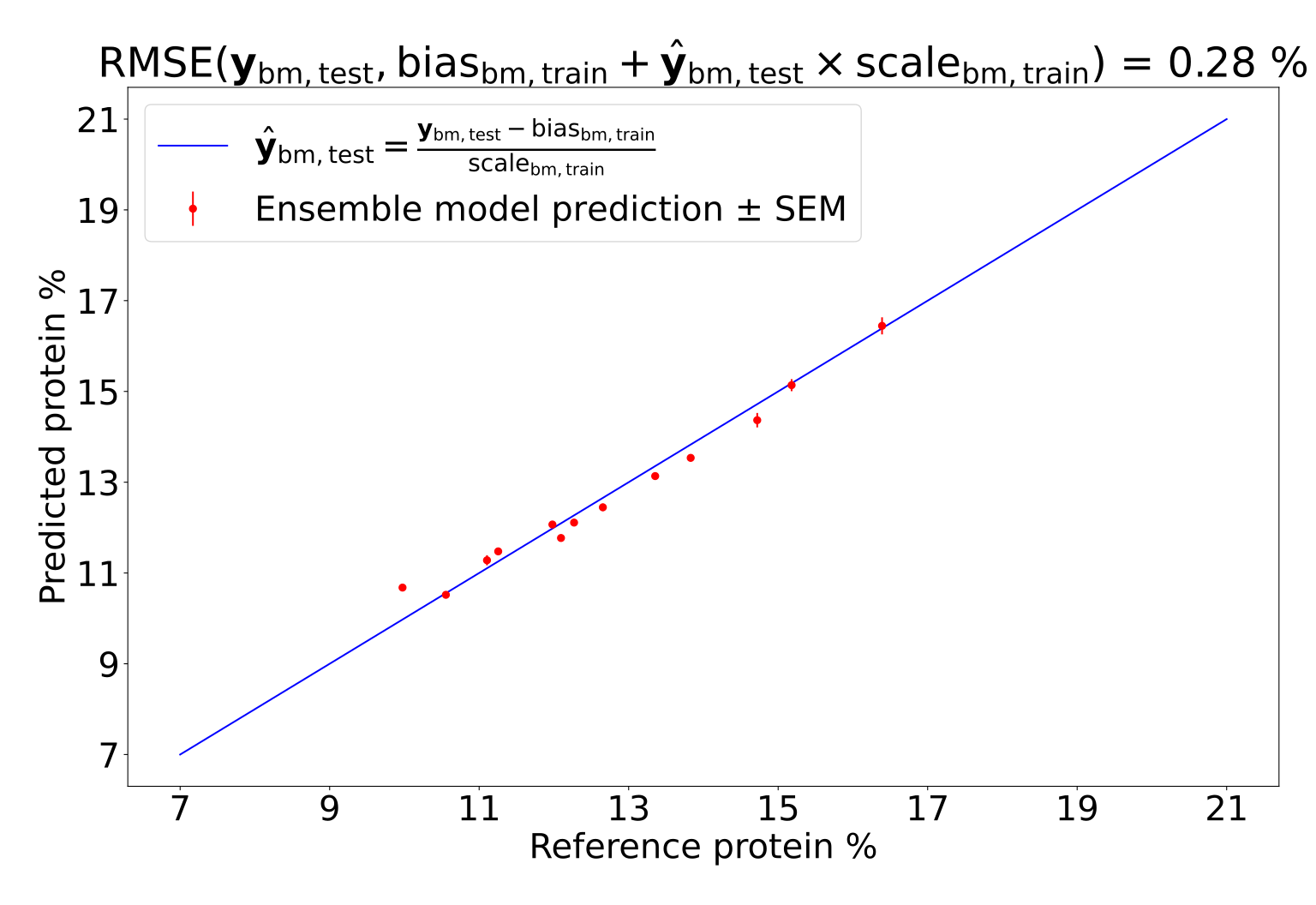}
        \caption{}
        \label{fig:resnet_corrected_bulk_pred}
    \end{subfigure}
    \caption{Scatter plots of the bias and scale corrected means of the predicted protein content distribution for each of the 13 bulk samples in the test set. (a) PLS. (b) ResNet-18.}
    \label{fig:corrected_bulk_preds}
\end{figure*}

\myfigref{fig:corrected_bulk_preds} shows the bulk mean predictions, corrected with bias and scale parameters from the training set. Here, we achieve an RMSE of approximately $0.28\%$ for both the PLS and ResNet-18 ensembles. This RMSE matches the standard deviation of the Infratec\texttrademark{} NOVA's (\citealt{InfratecNova}) measurement, which was used to obtain the reference values. Thus, this RMSE is the lowest that can be expected from a model that perfectly estimates protein content without considering the Infratec\texttrademark{} NOVA's measurement errors. Based on these results, inflating an NIR-HSI dataset by cropping the images and applying the mean reference to all image crops is a valid strategy for generating enough data to train a CNN. However, care must be taken to compensate for the discrepancy between the least squares solution for subsample predictions and the bulk mean predictions. As a final note in this section, inflating the dataset might be necessary to generate enough data for training CNNs, but it is not necessary for training PLS models. We (\citealt{engstrom2023analyzing}) trained a PLS model directly on the bulk mean spectra and got results matching those in \myfigref{fig:corrected_bulk_preds} (see Figure 7 (c) in \myappref{app:bulk_ref}).

\section{Grain Density and Model Performance}\label{sec:grain_density}
\begin{figure*}[ht]
    \centering
    \begin{subfigure}[t]{0.48\textwidth}
        \centering
        \includegraphics[width=\textwidth]{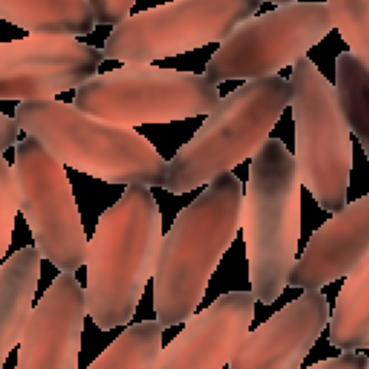}
        \caption{}
        \label{fig:dense_crop}
    \end{subfigure}
    \hfill
    \begin{subfigure}[t]{0.48\textwidth}
        \centering
        \includegraphics[width=\textwidth]{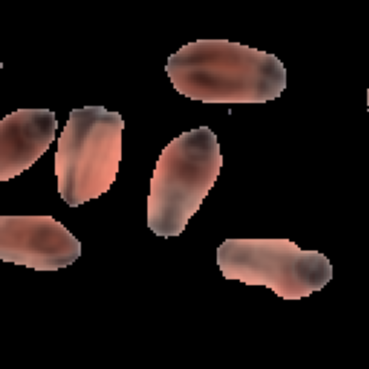}
        \caption{}
        \label{fig:sparse_crop}
    \end{subfigure}
    \caption{Image crops with (a) densely packed grain with a grain density of $0.67$ (b) sparsely packed grain with a grain density of $0.13$. This figure is copied from \citet{engstrom2023analyzing}.}
    \label{fig:crop_density}
\end{figure*}

Another relevant insight due to \citet{engstrom2023analyzing} is an analysis of the relationship between grain density and model performance for protein content regression and grain variety classification. Recall that both datasets consist of sparsely and densely packed image crops with varying grain densities of at least $0.1$ as illustrated in \myfigref{fig:crop_density}.

\begin{figure*}[ht]
    \centering
    \begin{subfigure}[t]{0.48\textwidth}
        \centering
        \includegraphics[width=\textwidth]{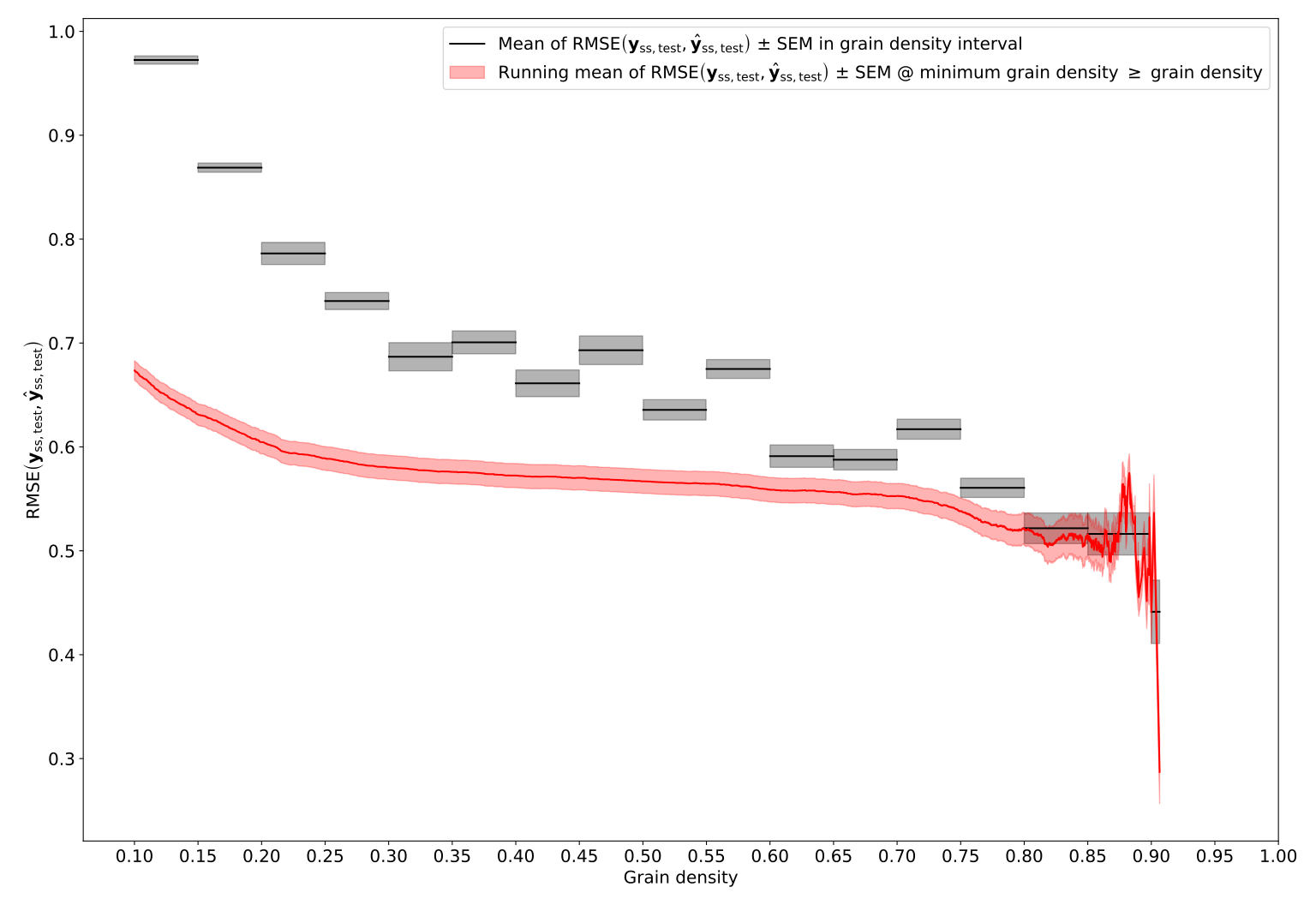}
        \caption{}
        \label{fig:pls_rmse_at_density}
    \end{subfigure}
    \hfill
    \begin{subfigure}[t]{0.48\textwidth}
        \centering
        \includegraphics[width=\textwidth]{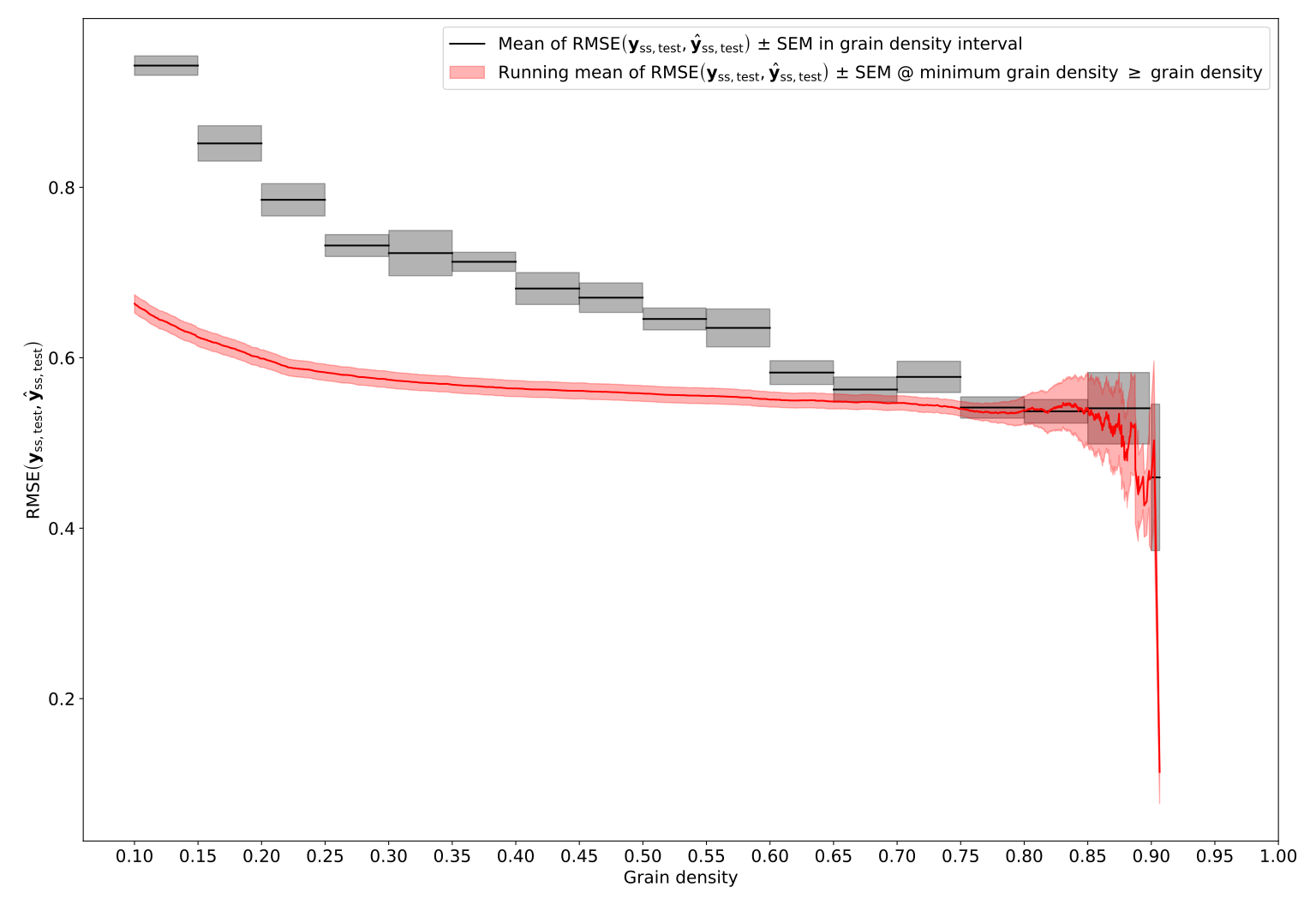}
        \caption{}
        \label{fig:resnet_rmse_at_density}
    \end{subfigure}
    \caption{Predicted protein RMSE as a function of grain density for (a) PLS and (b) ResNet-18. The gray lines show RMSE for predictions on image crops inside that grain density interval. At any point on the grain density axis, the red curve shows RMSE for predictions on image crops with at least that high grain density. This figure is copied from \citet{engstrom2023analyzing}.}
    \label{fig:rmse_at_density}
\end{figure*}

For protein content regression on Dataset \#1, we expect an image crop to have a true protein content closer to the bulk reference as its grain density increases. We expect this to be reflected in our models by making image crop-wise predictions closer to the bulk reference as the grain density increases. We use the ensembles of best-performing ResNet-18 and PLS models mentioned in \mysecref{sec:bulk_and_subsample} to analyze whether this hypothesis holds. \myfigref{fig:rmse_at_density} shows grain density versus mean RMSE and the standard error on the mean (SEM) for both the ResNet-18 and PLS ensembles. The mean RMSEs are computed across the five constituents of each ensemble to see intra-ensemble agreement, which is high for both PLS and ResNet-18. For both model types, it seems clear that RMSE decreases as grain density increases.

\begin{figure*}[ht]
    \centering
    \begin{subfigure}[t]{0.48\textwidth}
        \centering
        \includegraphics[width=\textwidth]{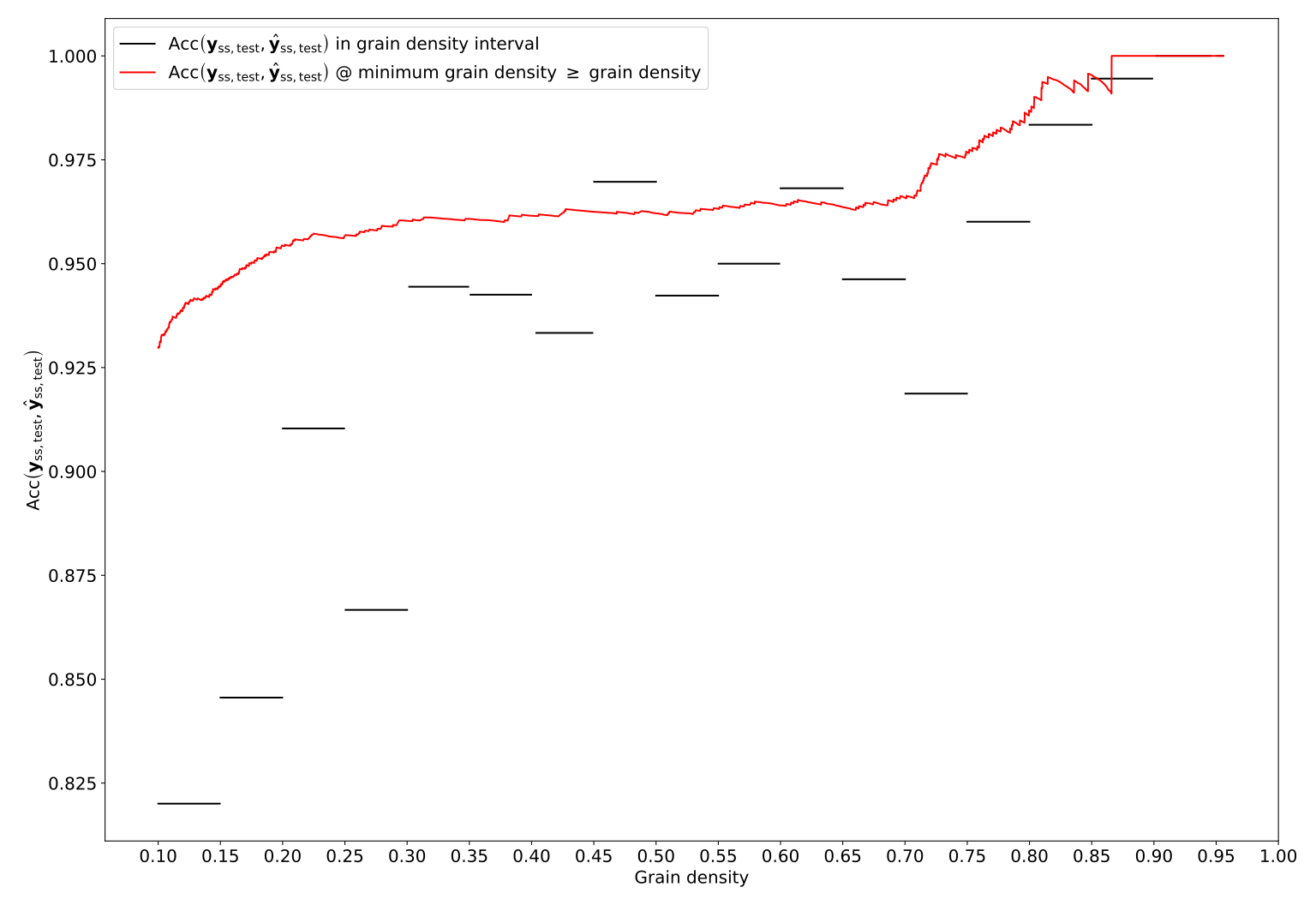}
        \caption{}
        \label{fig:pls_acc_at_density}
    \end{subfigure}
    \hfill
    \begin{subfigure}[t]{0.48\textwidth}
        \centering
        \includegraphics[width=\textwidth]{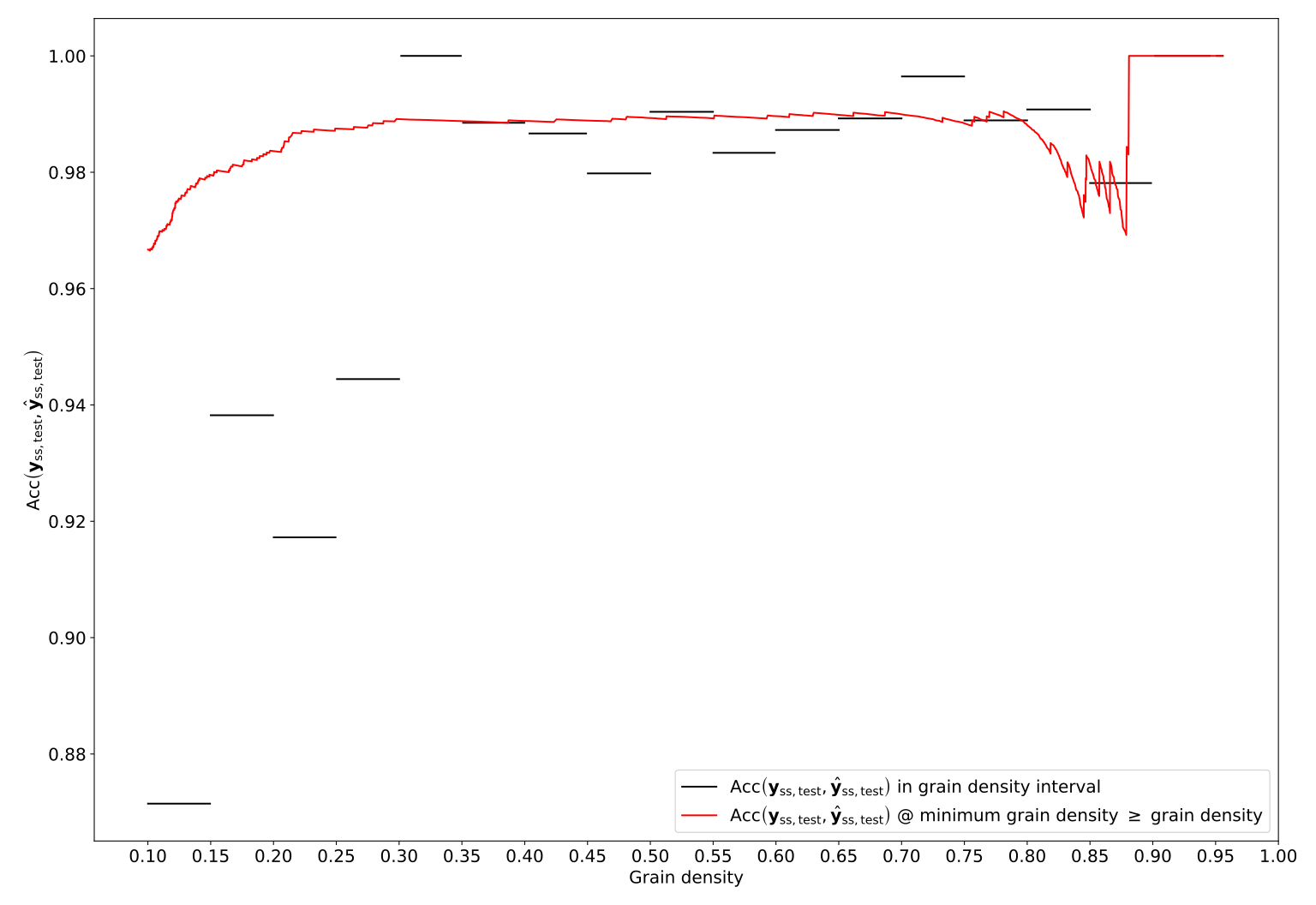}
        \caption{}
        \label{fig:resnet_acc_at_density}
    \end{subfigure}
    \caption{Grain variety classification accuracy as a function of grain density for (a) PLS and (b) ResNet-18. The gray lines show accuracy for predictions on image crops inside that grain density interval. At any point on the grain density axis, the red curve shows accuracy for predictions on image crops with at least that high grain density. This figure is copied from \citet{engstrom2023analyzing}.}
    \label{fig:acc_at_density}
\end{figure*}

As mentioned previously, for grain variety classification between the eight varieties in Dataset \#2, the bulk references are also entirely valid for the image crops due to a bulk consisting of exactly one grain variety. Still, higher-density image crops contain more grain and, thus, potentially, a stronger signal for the models to respond to. \myfigref{fig:acc_at_density} shows accuracy as a function of grain kernel density for the best CNN (a ResNet-18 modification) and a PLS model with optimal preprocessing as determined by \citet{engstrom2023improving}. While the ResNet-18 outperforms the PLS model, both see a clear drop in accuracy when grain density falls below $0.3$ and are otherwise relatively stable above $0.3$. These results contrast those by \citet{dreier2022hyperspectral}, where accuracy falls drastically when grain density decreases below $0.5$. For grain density between $0.1$ and $0.2$, their accuracy falls below $0.4$ for a ResNet-18. The difference, however, is that \citet{engstrom2023improving} train on image crops with a grain density of at least $0.1$, whereas \citet{dreier2022hyperspectral} use a grain density of at least $0.5$ for training and so anything below is out of domain.

Thus, models can benefit from training on image crops with low grain density for grain variety classification, especially if these are encountered during evaluation. Higher grain density also correlates somewhat with higher classification accuracy, especially around the $0.3$ grain density threshold.

\chapter{Chemical Maps}\label{chap:chem_maps}
\lettrine{T}{he} previous chapters have shown that while hyperspectral images can be used to model chemical parameters, the spectral signals alone are sufficient for accurate analysis. Training a deep learning model such as ResNet-18 on hyperspectral images is extremely computationally expensive compared to training PLS on spectra. Coupled with the high cost of an NIR-HSI platform compared to the price of a spectrometer, it may initially seem difficult to justify using NIR-HSI to model chemical parameters. However, a use case exists that justifies and necessitates an NIR-HSI platform for the modeling of chemical parameters. That is the modeling of the spatial distribution of a chemical parameter throughout a sample. The spatial distribution can be visualized in so-called chemical maps, which are images of pixel-wise predictions generated from an input hyperspectral image.

Chemical maps are commonly generated by training a PLS model on mean spectra and subsequently using it to predict every pixel (spectrum) in the input hyperspectral images. This approach has multiple issues, including non-smooth chemical maps, pixel-wise predictions outside the physically possible range, and a discrepancy between the mean of the predicted distribution and a prediction on the mean spectrum. In the following work (\citealt{engstrom2025chem}), we compare PLS-generated chemical maps with those generated by U-Net. U-Net (\citealt{ronneberger2015u}) is a CNN originally designed for semantic segmentation. In semantic segmentation, each pixel of an input image must be classified as belonging to some class. Chemical map generation is the equivalent regression task. Therefore, based on our previous work with CNNs for chemical parameter regression on hyperspectral images (\citealt{engstrom2021predicting, engstrom2023improving}), we augment the U-Net with an initial spectral convolution layer and use it to generate chemical maps. Indeed, with a customized loss function, directly addressing the issues previously mentioned with PLS, we can generate smooth chemical maps with pixel-wise predictions strictly inside the physically possible range with mean values matching the bulk references.

In this work, we use cheap-to-obtain bulk mean references similar to those used in the previous chapters. However, instead of forcing U-Net's pixel-wise predictions to match the bulk mean reference, we allow for pixel-wise deviations from the mean reference by only comparing the mean of the predicted spatial distribution with the bulk mean reference. While a bulk mean reference is computed across a physical sample corresponding to five images, it seems likely that the discrepancy between the mean bulk reference value and the actual true value for each image is smaller and subject to less variance than in our previous work with protein content regression on grain where each bulk sample corresponded to thousands of image crops. Indeed, the line of best fit for the mean predictions generated by \citet{engstrom2025chem} exhibits only a slight deviation from the identity.

\includepdf[pages=-]{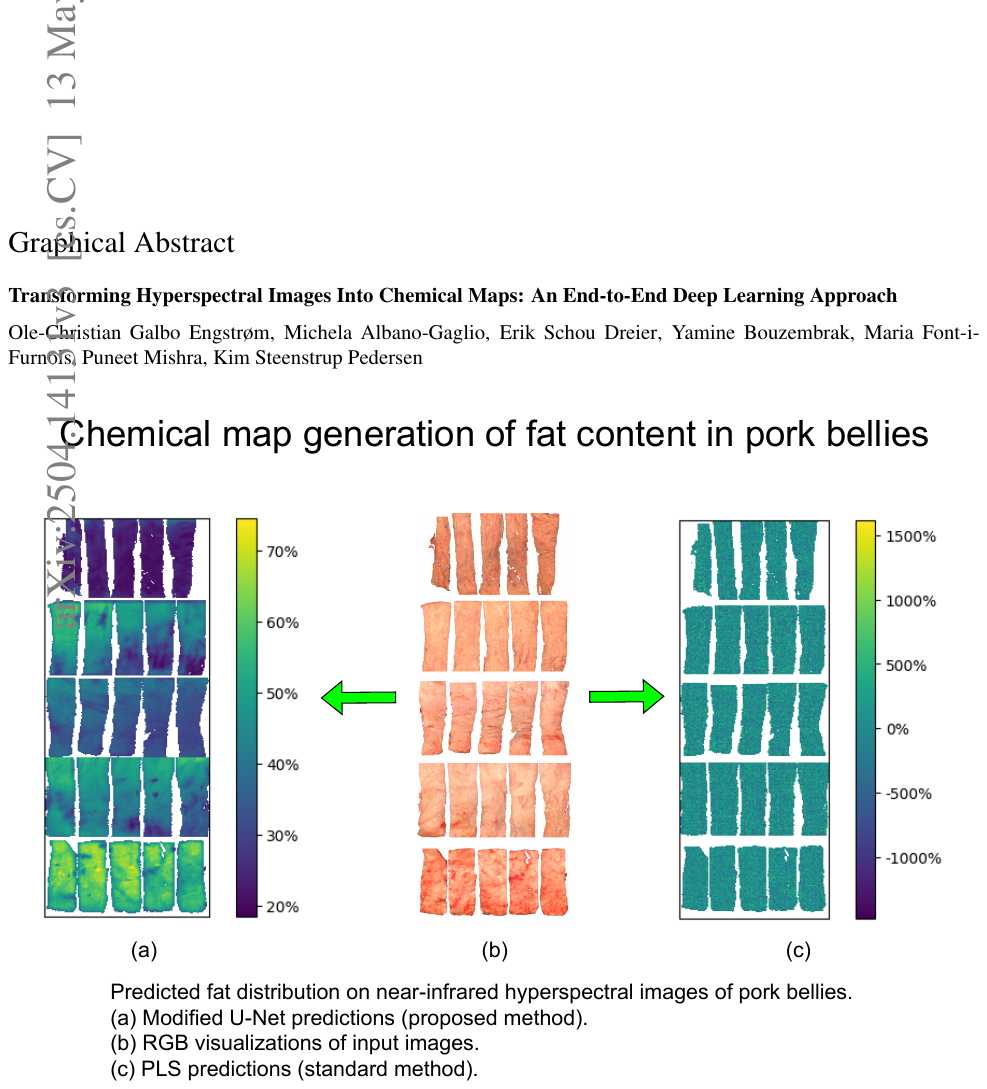}
\chapter{Barley Germination}\label{chap:barley_germination}
\lettrine{T}{he} previous chapters have shown how joint spatio-spectral analysis of NIR-HSI images can be used for various tasks. Still, we have not encountered a parameter that could not be modeled to a somewhat accurate degree using only spatial or spectral information alone. Aside from modeling a chemical parameter's spatial distribution, we have yet to see a parameter that requires joint spatio-spectral analysis of NIR-HSI images. Such a parameter may be barley's germinative capacity, which is the measure of barley's ability to germinate when it is exposed to moisture. A recent review (\citealt{orth2025current}) mentions that both spatial and spectral features are relevant to determining whether a barley kernel will germinate. In particular, during germination, a chemical process occurs inside the kernel but may be inhibited if the kernel is sufficiently damaged by, for example, being cleaved. This chapter is written as a technical report of a recent and previously unpublished study into predicting barley's germinative capacity using the open dataset of RGB and NIR-HSI dataset collected by \citet{engstrom2025dataset}. The technical report describing the dataset, from which this chapter will paraphrase and reuse figures freely, is reprinted in \myappref{app:germination_dataset} for convenience. Although this chapter is written by Ole-Christian Galbo Engstr{\o}m, the experiments were carried out in close collaboration with Erik Schou Dreier, Birthe M{\o}ller Jespersen, and Kim Steenstrup Pedersen.

\section{Introduction}
Barley is commonly used for malting to produce various beverages. During the malting process, barley must germinate rapidly and uniformly (\citealt{GUBLER2005183}). Therefore, to maximize yield, barley's germinative capacity must be evaluated prior to malting. Current official methods for this evaluation are based on the destructive and manual assessment of a few hundred kernels (\citealt{GerminationEBC}). This implies that bulks with heterogeneous germinative capacity can not necessarily be divided into subsamples with uniform germinative capacity.

Furthermore, the relatively small number of kernels used for current methods raises the question of whether the analyzed subsample is representative of the entire bulk. Thus, a rapid and non-destructive assessment method for quantifying barley's germinative capacity can potentially revolutionize the malting industry. \citet{orth2025current} mention that NIR-HSI may be a viable method. Indeed, \citet{arngren2011analysis} used NIR-HSI to model the amount of time before single barley kernels germinate after exposure to moisture. They used a dataset with uniformly distributed reference values and divided 60 hours into eight intervals for the classification of germination time. They achieved 41\% accuracy on single kernels. While this may seem low and likely not satisfactory for a standard method, it is much better than random guessing.

In the following study, like \citet{arngren2011analysis}, we analyze the germination time of single barley kernels using NIR-HSI images. However, we also include RGB images to compare the predictive performance that can be achieved with a relatively cheap color camera that sees spatial, but no chemical information, a full spatio-spectral NIR-HSI image, and NIR mean spectra. Thus, we attempt to assess whether it is possible to predict barley's germinative capacity and to what degree of accuracy with all three modalities.

\section{Dataset}
This section briefly describes the dataset and gives sufficient information to understand the subsequent modeling process. For a complete dataset description, including its acquisition, see \myappref{app:germination_dataset}.

\begin{figure*}[h]
    \begin{subfigure}[t]{0.48\textwidth}
        \centering
        \includegraphics[width=\textwidth]{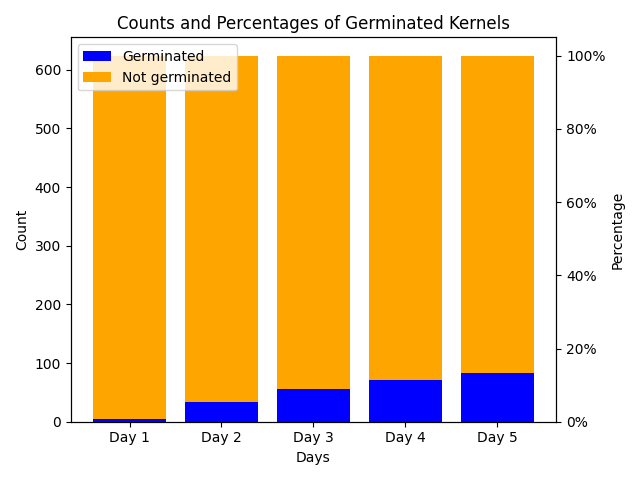}
        \caption{Prospect\_0.}
        \label{fig:Prospect_0_hist}
    \end{subfigure}
    \hfill
    \begin{subfigure}[t]{0.48\textwidth}
        \centering
        \includegraphics[width=\textwidth]{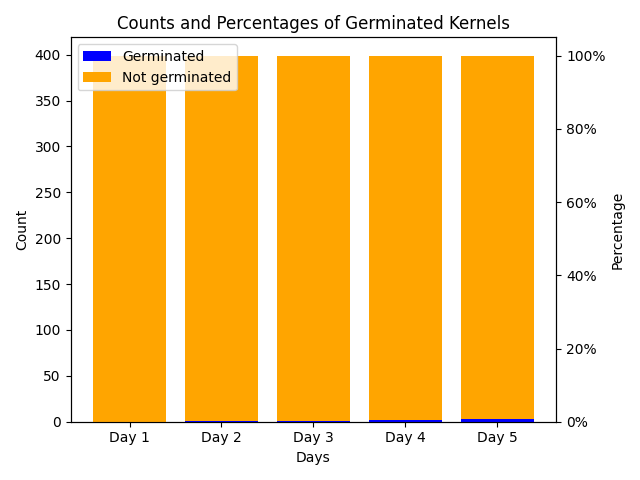}
        \caption{Prospect\_1.}
        \label{fig:Prospect_1_hist}
    \end{subfigure}
    \\
    \begin{subfigure}[t]{0.48\textwidth}
        \centering
        \includegraphics[width=\textwidth]{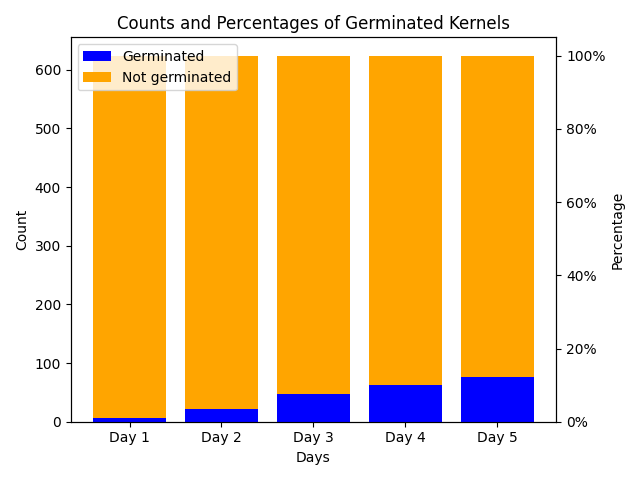}
        \caption{Laureate\_0}
        \label{fig:Laureate_0_hist}
    \end{subfigure}
    \hfill
    \begin{subfigure}[t]{0.48\textwidth}
        \centering
        \includegraphics[width=\textwidth]{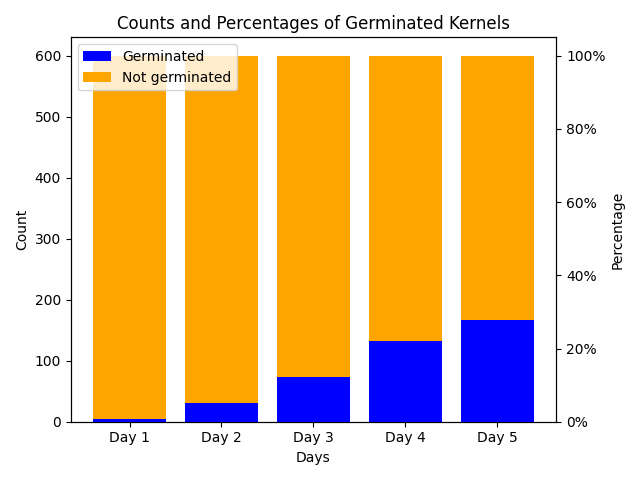}
        \caption{Laureate\_1}
        \label{fig:Laureate_1_hist}
    \end{subfigure}
    \hfill
    \begin{subfigure}[t]{0.48\textwidth}
        \centering
        \includegraphics[width=\textwidth]{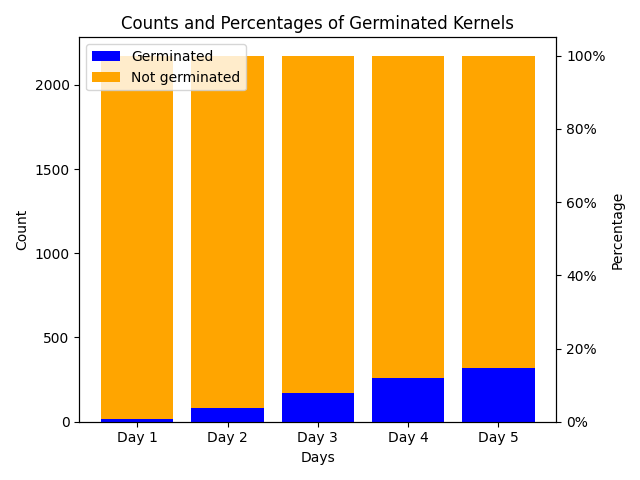}
        \caption{All varieties.}
        \label{fig:all_hist}
    \end{subfigure}
    \caption{Histograms showing how the number of germinated kernels evolves from day to day. This figure is copied from \citet{engstrom2025dataset}.}
    \label{fig:varieties_hists}
\end{figure*}

We use a time series dataset of RGB and NIR-HSI images of 2242 individual barley kernels. The NIR-HSI images contain 204 uniformly distributed wavelength channels between 938 nm and 1662 nm. The barley kernels are taken from four different malting barley samples: two of the Prospect variety and two of the Laureate variety. \mytabref{tab:num_samples} shows the distribution of kernels per sample. We imaged each barley kernel before exposure to moisture and every 24 hours after exposure to moisture for five consecutive days. \mytabref{tab:grain_evolution} shows the twelve images associated with a single grain kernel. Consequently, the barley kernels are labeled by the day (1-5) at which a visible sprout emerged or labeled as not germinated if no visible sprout appeared before image acquisition on day 5. The distribution of labels for germinated kernels is shown in \mytabref{tab:germination} and visualized together with the distribution of non-germinated kernels in \myfigref{fig:varieties_hists}. Associated with each image is also a segmentation mask based on a combination of Otsu's method for thresholding (\citealt{Otsu79a}) and connected components on the thresholded image. The mask is used to extract a mean NIR spectrum from the NIR-HSI image. When extracting the mean NIR spectrum, we first transform every pixel to pseudo-absorbance by taking the negative of its logarithm.

The dataset is split into six evenly sized partitions (folds) by stratified random sampling to ensure identical distributions of germination days throughout the six partitions to the maximum possible degree. For example, 28 kernels germinated on day one, so four partitions will each contain five of these kernels, and the remaining two partitions will contain four each. Five partitions are used for five-fold cross-validation, and the sixth partition is used for testing. This stratification does not consider the distribution of germination days per barley variety, \mysecref{sec:germination_discussion} discusses this.

Malting barley kernels are typically expected to germinate to a degree close to 100\%. Thus, it is clear that something has gone wrong in the germination process of this dataset where less than 20\% of the kernels germinate and typically show just a tiny sprout where a long sprout is typically observed (see, for example, \mytabref{tab:germination}). \citet{engstrom2025dataset} determined the cause for this low germinative capacity to be glue residue. Specifically, the barley kernels were placed inside a 3D-printed grid to facilitate segmentation and to avoid mixing individual kernels during storage. Glue residue was stuck to the underside of these 3D-printed grids. The glue must have mixed with the water used to moist the kernels, and the kernels have subsequently absorbed the contaminated water, which seems to have severely inhibited their germinative capacity.

\begin{table}[h]
\centering
\begin{tabular}{@{}ccccc|c|@{}}
\toprule
             & \multicolumn{4}{c}{Barley Variety} \\ \cmidrule{2-6}
             & Prospect\_0  & Prospect\_1 & Laureate\_0 & Laureate\_1 & Total\\ \midrule
No. kernels  & 624 & 394 & 624 & 600 & 2242 \\ \bottomrule
\end{tabular}
\caption{Number of barley kernels for each of the four barley varieties. This table is copied from \citet{engstrom2025dataset}.}
\label{tab:num_samples}
\end{table}

\begin{table}[h]
\centering
\begin{tabular}{c|c|c|c|c|c|c|}
\cline{2-7}
                              & Pre-moisture & Day 1 & Day 2 & Day 3 & Day 4 & Day 5 \\ \hline
\multicolumn{1}{|c|}{RGB}     & \includegraphics[width=1.25cm]{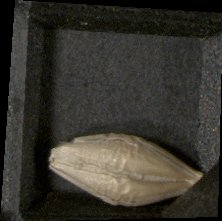} & \includegraphics[width=1.25cm]{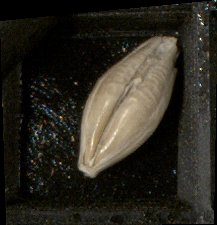} & \includegraphics[width=1.25cm]{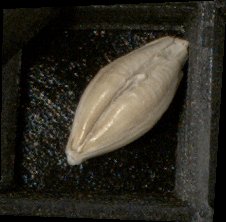} & \includegraphics[width=1.25cm]{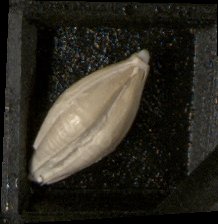} & \includegraphics[width=1.25cm]{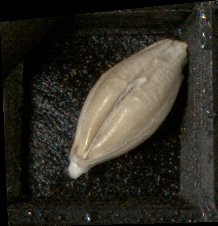} & \includegraphics[width=1.25cm]{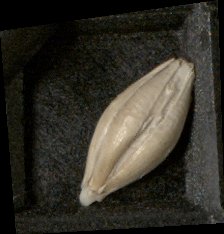} \\ \hline
\multicolumn{1}{|c|}{NIR-HSI} & \includegraphics[width=1.25cm]{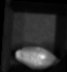} & \includegraphics[width=1.25cm]{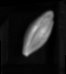} & \includegraphics[width=1.25cm]{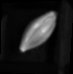} & \includegraphics[width=1.25cm]{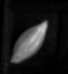} & \includegraphics[width=1.25cm]{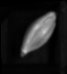} & \includegraphics[width=1.25cm]{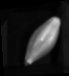} \\ \hline
\end{tabular}
\caption{A single barley kernel imaged before exposure to moisture and every 24 hours after exposure to moisture. A sprout appears on Day 2, meaning that the reference value for germination time for this specific kernel is two days. This table is copied from \citet{engstrom2025dataset}.}
\label{tab:grain_evolution}
\end{table}

\begin{table}[h]
\centering
\begin{tabular}{@{}cccccc|c|@{}}
\toprule
             & \multicolumn{6}{c}{Germination Day} \\ \cmidrule{2-7}
barley variety & 1 & 2 & 3 & 4 & 5 & Any \\ \midrule
Prospect\_0 & 4 & 29 & 23 & 15 & 12 & 83 \\
Prospect\_1 & 1 & 0 & 1 & 0 & 3 & 5 \\
Laureate\_0 & 13 & 21 & 25 & 15 & 17 & 91 \\
Laureate\_1 & 9 & 43 & 57 & 58 & 35 & 202 \\ \midrule
All  & 28 & 93 & 106 & 88 & 67 & 382 \\ \bottomrule
\end{tabular}
\caption{Number of barley kernels that germinated each day out of those shown in \mytabref{tab:num_samples}. No kernels were germinated prior to exposure to moisture. This table is copied from \citet{engstrom2025dataset}.}
\label{tab:germination}
\end{table}

\section{Modeling}
Given the granularity of our labels (\mytabref{tab:germination}), we have identified three tasks in order of increasing complexity.

\begin{enumerate}
    \item Modeling of whether a given kernel is germinated. For this task, we use the images from day five to maximize the number of germinated kernels.
    \item Modeling of whether a given kernel will germinate in the future. For this task, we use the images acquired pre-moisture but label the image as germinated if the kernel germinated at any time during the experiment.
    \item Modeling of not whether a kernel will germinate but when it will germinate. This is the task of taking the images at pre-moisture and predicting their exact germination day.
\end{enumerate}

Due to the results presented in the next section, we decided to undertake only the first two tasks and not attempt the third task.

The dataset is heavily unbalanced, so we apply class weights to mitigate this. We compute the class weights for germinated and not germinated independently for each barley variety, the training and validation partitions, and the test partition. We use \citet{scikit-learn} to compute the class weights based on a heuristic from \citet{king2001logistic}. The following is a description of the class weights. Generally, we can compute class weights for a class $c \in \{1\ldots C\}$ where $C$ is the number of unique classes. In our case, $C=2$, but the following holds in general. Let $S$ be the set of samples with a size of $N=|S|$. Then $S$ can be indexed by $i \in \{1\ldots N\}$ to get sample $S_{i}$. Let $\mathbf{1}_c$ be the indicator function for $c$, evaluating to $1$ if its argument belongs to $c$ and otherwise to 0. Then, $N_{c}=\sum_{i=1}^{N}\mathbf{1}_{c}\left(S_{i}\right)$ is the number of samples belonging to $c$, and the weight for $c$ is defined by

\begin{equation}\label{eq:class_weight}
    w_{c} = \frac{N}{CN_{c}}.
\end{equation}

This implies that the total weight for any given class is constant. That is, $w_{c}N_{c}=\frac{N}{C}$ for any $c$. This is proven by

\begin{equation}\label{eq:sum_of_weights_for_class}
    w_{c}N_{c} = \frac{NN_{c}}{CN_{c}} = \frac{N}{C}.
\end{equation}

This also implies that the total weight across all samples is $N$, proven by

\begin{equation}\label{eq:sum_of_all_weights}
    \sum_{c=1}^{C}w_{c}N_{c} = \frac{NC}{C} = N.
\end{equation}

Thus, the average weight across all samples is $1$. When used in a deep learning context, this also means that applying these weights does not effectively change the learning rate.

\subsection{PLS}

For PLS-based modeling, we use the weighted variation (\citealt{becker2016accounting}) implemented by \citet{engstrom2024ikpls}. We model PLS with the mean absorbance spectra and experiment with two combinations of preprocessing. The first combination consists of applying the Standard Normal Variate (SNV, \citealt{barnes1989standard}) transformation followed by convolution with a Savitzky-Golay (SG, \citealt{savitzky1964smoothing, steinier1972smoothing}) filter with a window size of seven, a polynomial order of two, and a derivative order of two. The second combination does not have row-wise preprocessing but a column-wise weighted centering of spectra and target values. These two combinations correspond to the optimal settings for protein content regression and grain variety classification in the work by \citet{engstrom2023improving}.

We train PLS under a five-fold cross-validation scheme and determine the best number of components as the one maximizing weighted accuracy (WAcc) on the validation partitions. Subsequently, we average the five best numbers of components, refit the PLS model on the entire cross-validation dataset with that many components, and use this model for evaluation on the test set. This approach is similar to the one by \citet{engstrom2025chem}.

\subsection{Image-based Modeling}

For image-based modeling, we employ variants of ResNet-18 (\citealt{he2016deep}) based on the work by \citet{engstrom2023improving}. For RGB images, we use their vanilla variant of ResNet-18. For NIR-HSI images, we employ the ResNet-18 variant with an initial 3D convolution layer with a single filter with a kernel size of (height, width, depth) = (3, 3, 7) and a stride of 1. Subsequently, we employ a downsampler (\citealt{engstrom2023improving}) with three filters, effectively reducing the channel size from 204 to 3. The vanilla ResNet-18 then consumes the output of the downsampler. We also experiment with a variant using the 3D convolution layer without the subsequent downsampler for NIR-HSI images. We implement all models and deep learning pipelines using PyTorch (\citealt{ansel2024pytorch}).

We apply channel-wise centering and scaling to the images based on training partition-wise mean and standard deviation spectra, respectively, RGB-values. We compute the means and standard deviations across all pixels in the training partition for this task.\footnote{In practice, we compute the means and standard deviations only across pixels inside the polygons that define the cutouts from the Petri dishes from which the kernel images are extracted. See Figure 4 in \myappref{app:germination_dataset} for an example. We also only apply centering and scaling inside the polygons, leaving the pixels outside of the polygons as 0.} Due to the extreme number of pixels across an entire training partition, the computation of channel-wise means and standard deviations risks suffering from numerical errors. For this reason, we employ a two-pass variant of Welford's algorithm (\citealt{welford1962note}) and enhance it by replacing its internal summations with Neumaier summation (\citealt{neumaier1974rundlingsfehleranalyse}), which is a compensated summation algorithm improving upon the Kahan-Babu\v{s}ka compensated summation algorithm (\citealt{kahan1965pracniques, babuska1968numerical}). After centering and scaling, we zero-pad the images to be square and have constant side lengths of 336 pixels for RGB images and 112 pixels for NIR-HSI images. These are larger than any side lengths in the unpadded datasets, so no additional resizing or cropping is necessary. During training, we shuffle the training partition uniformly at random and independently for each epoch. We also employ random horizontal and vertical flipping, each with a probability of $0.5$.

In the ResNet's, we initialize all weights by sampling from Kaiming He Normal distributions (\citealt{he2015delving}) and all biases to zero. The loss function is the summation of weighted binary cross-entropy (WBCE) and L2-regularization (L2) of the weights using a $10^{-6}$ multiplier for L2.

We train the models under a five-fold cross-validation. We use batches of 32 images and optimize the models with Adam (\citealt{kingma2017adam}) using hyperparameters $\beta_{1}=0.9$ and $\beta_{2}=0.999$ and an initial learning rate of $10^{-2}$. During training, we keep track of the current best weights and associated optimizer state. The current best weights are those achieving the lowest validation partition WBCE. For the first 300 epochs, there is no change in the learning rate. Starting from the 301st epoch, we initialize counters for learning rate reduction and early stopping. After 100 epochs with no improvement in WBCE on the validation set, we restore the current best weights and associated optimizer state. Then, we multiply the learning rate $10^{-1}$ but cap it at $10^{-7}$. The training process halts if there is no improvement in the validation partition WBCE after 400 epochs or if it reaches 10,000 epochs. This implies that a model is trained for at least 700 epochs. In practice, all models were halted by the early stopping criterion.

Similar to our previous works (\citealt{engstrom2023improving, engstrom2025chem}), we construct an ensemble of the five models yielded by cross-validation. A uniform mean of its constituents' individual predictions gives the ensemble's predictions. We evaluate the ensemble on the test set. This gives us a model that is simultaneously trained and validated on all cross-validation partitions, like our final PLS model.

\section{Results}
For PLS, the model using only column-wise weighted centering had a few percentage points better WAcc across the validation partitions on both tasks. The best number of components was 11 for task 1 and 8 for task 2. For ResNet-18 on NIR-HSI images, the variant using a Downsampler achieved superior WBCE for task 1, and for task 2, it was tied with the variant without a downsampler. We keep the best PLS model and best ResNet-18 model for evaluation and disregard the others.

\begin{figure*}[h]
    \begin{subfigure}[t]{0.48\textwidth}
        \centering
        \includegraphics[width=\textwidth]{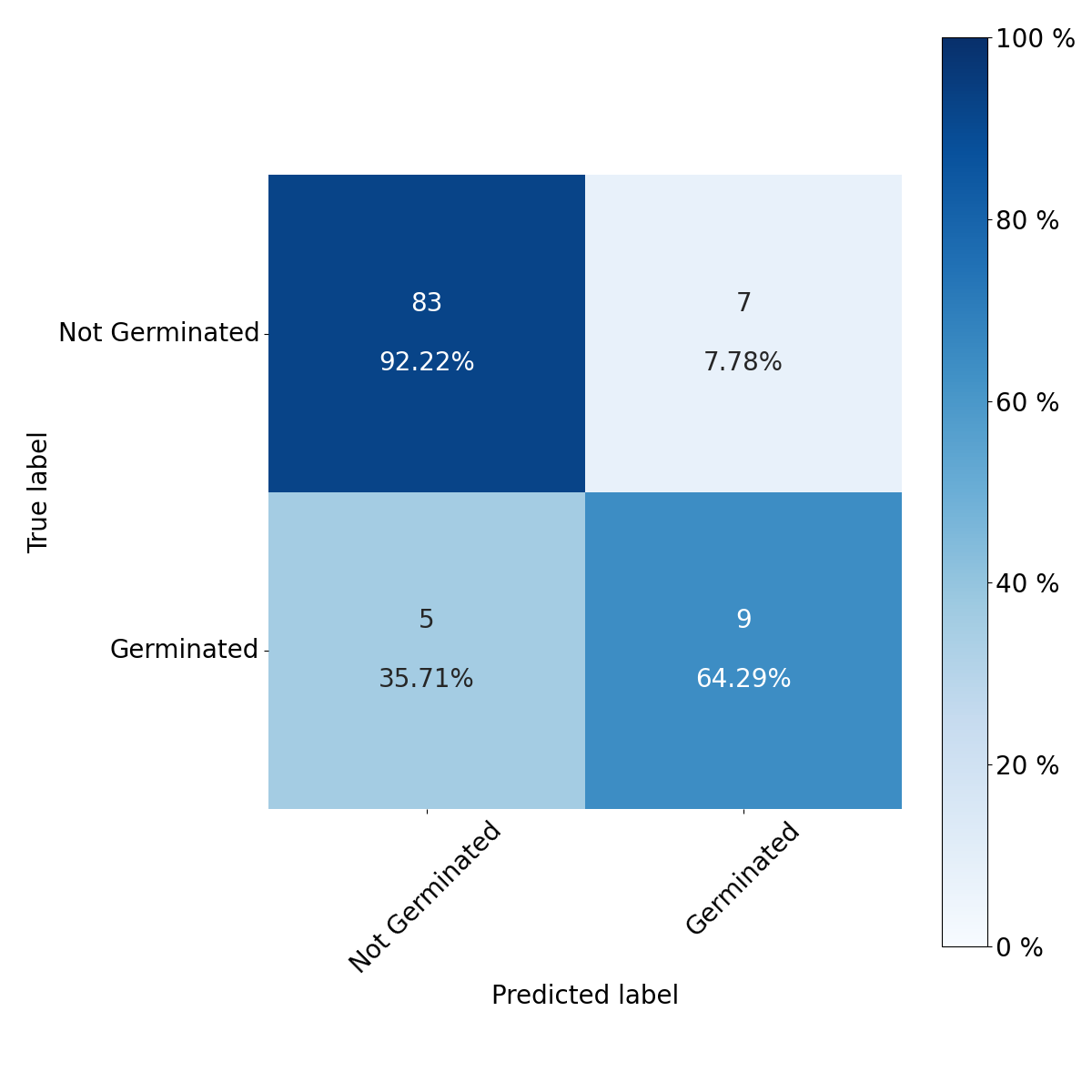}
        \caption{Prospect\_0.}
    \end{subfigure}
    \hfill
    \begin{subfigure}[t]{0.48\textwidth}
        \centering
        \includegraphics[width=\textwidth]{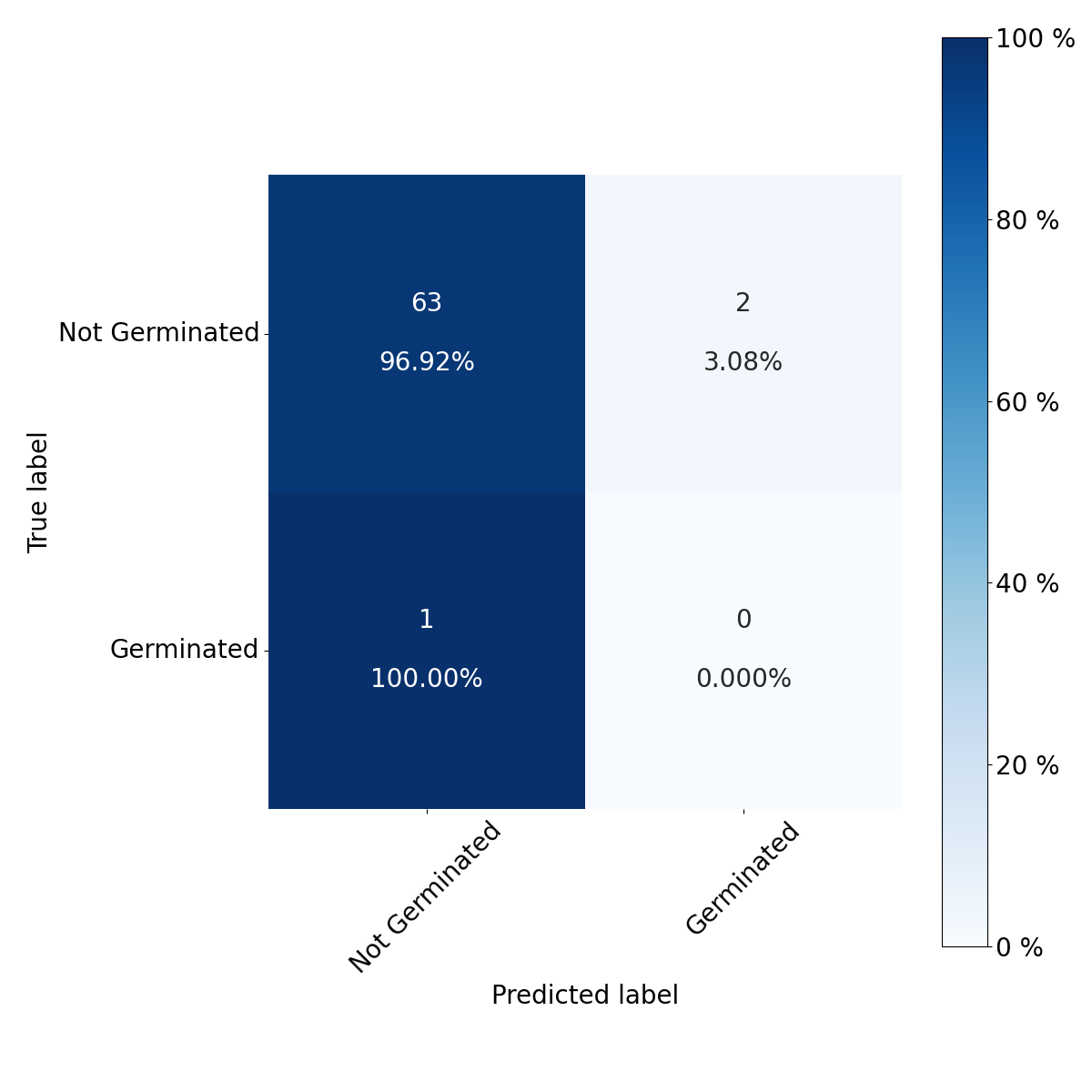}
        \caption{Prospect\_1.}
    \end{subfigure}
    \\
    \begin{subfigure}[t]{0.48\textwidth}
        \centering
        \includegraphics[width=\textwidth]{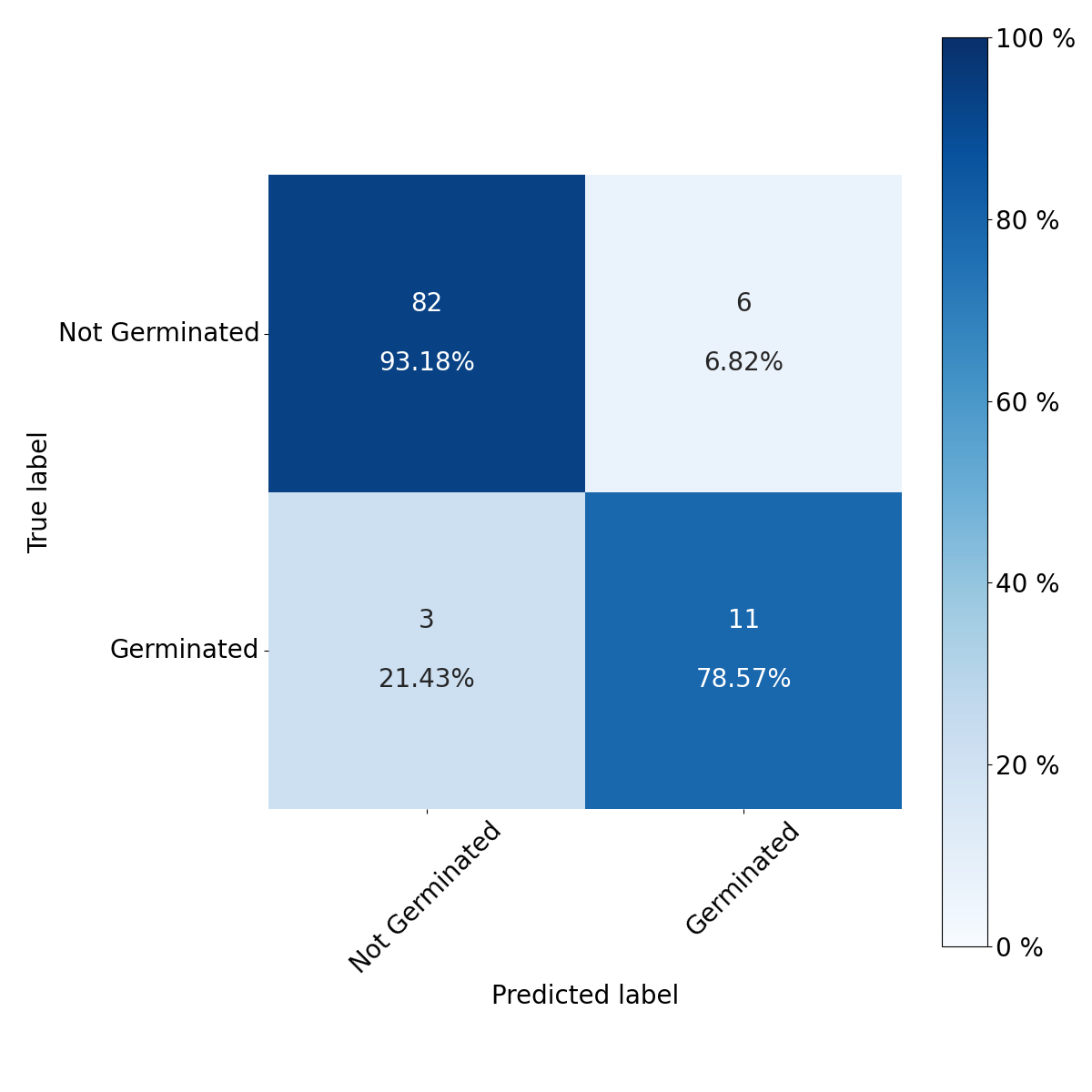}
        \caption{Laureate\_0.}
    \end{subfigure}
    \hfill
    \begin{subfigure}[t]{0.48\textwidth}
        \centering
        \includegraphics[width=\textwidth]{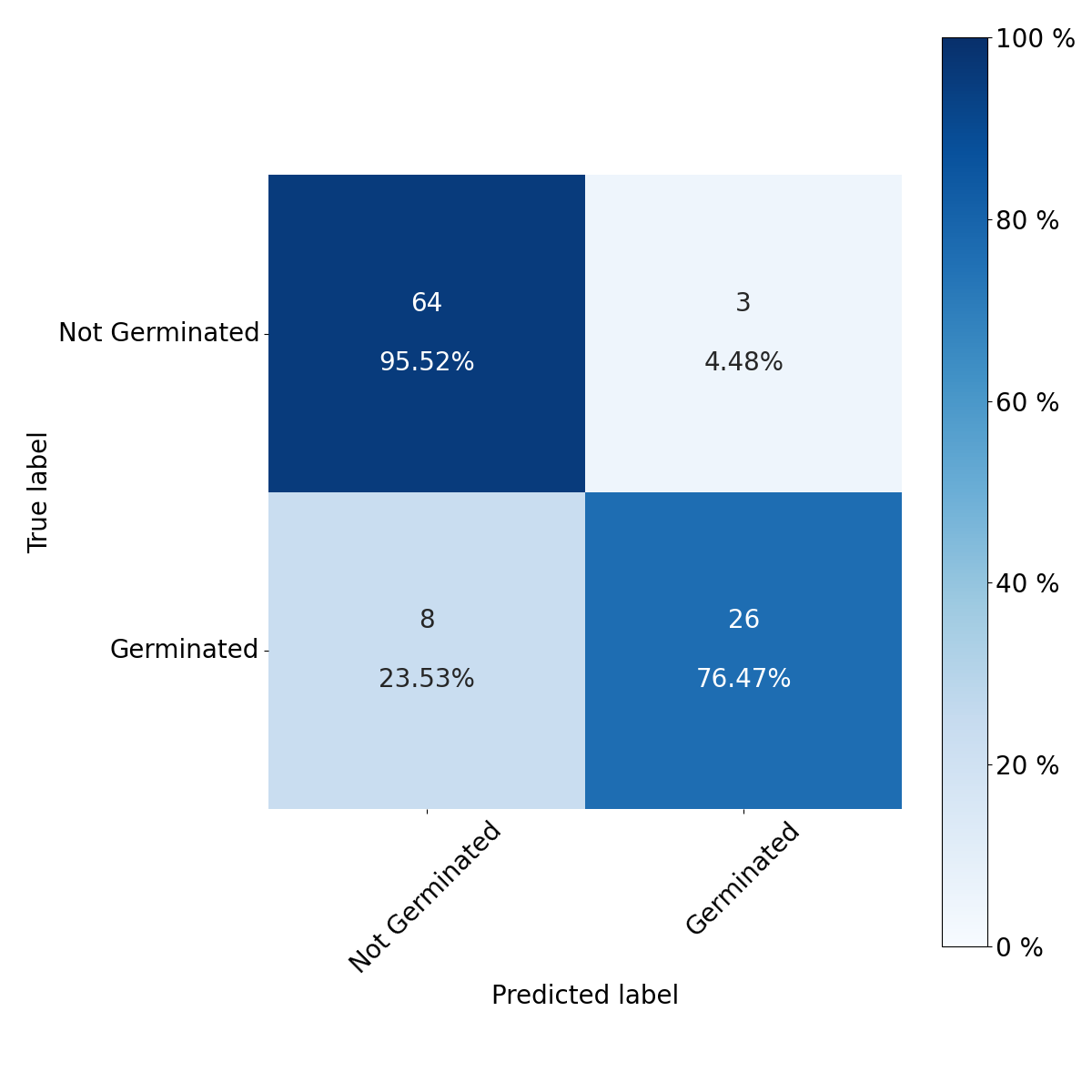}
        \caption{Laureate\_1.}
    \end{subfigure}
    \caption{Confusion matrices for test set predictions by the ResNet-18 NIR-HSI model on task 1.}
    \label{fig:conf_mat_task_1_resnet_nir_hsi}
\end{figure*}

\begin{figure*}[h]
    \begin{subfigure}[t]{0.48\textwidth}
        \centering
        \includegraphics[width=\textwidth]{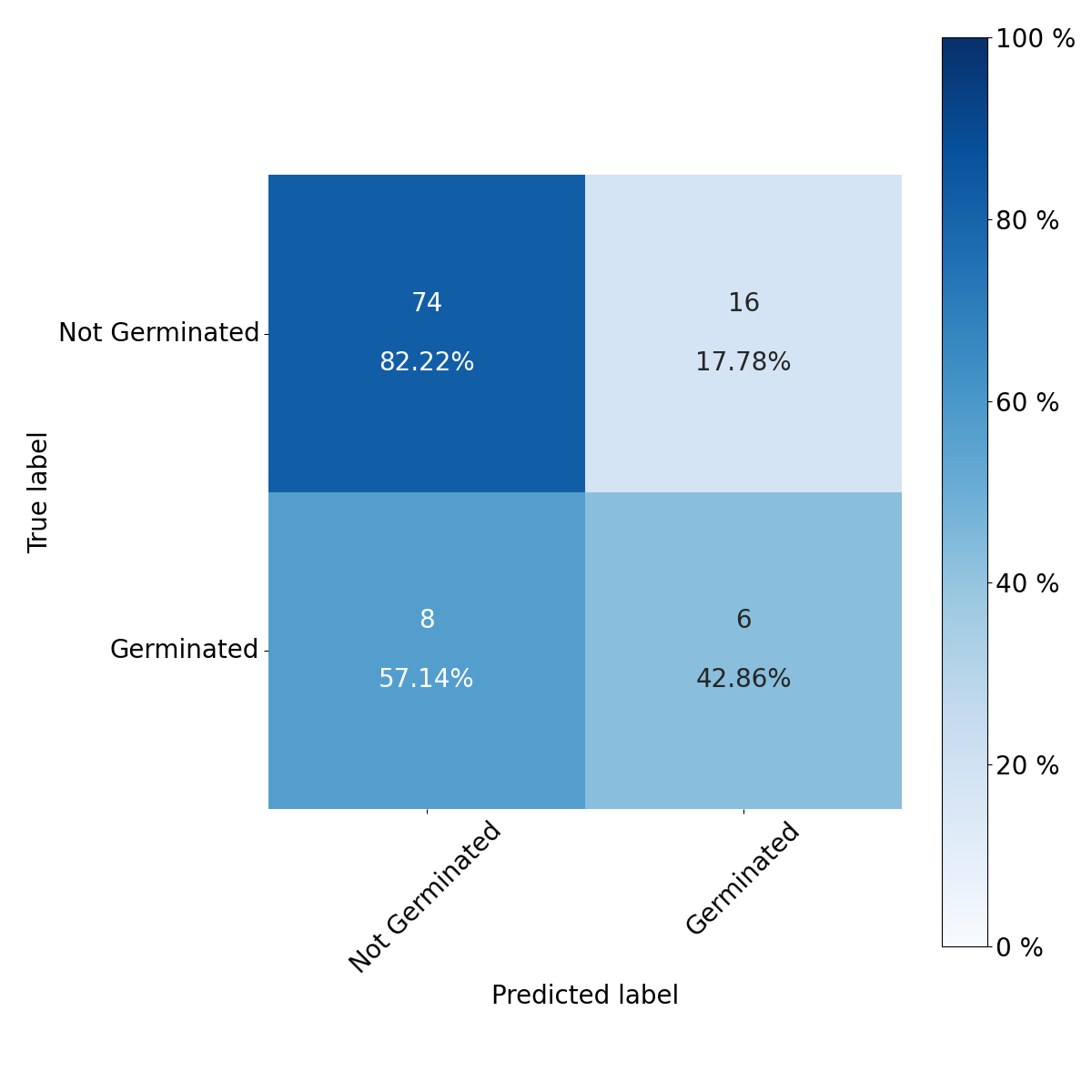}
        \caption{Prospect\_0.}
    \end{subfigure}
    \hfill
    \begin{subfigure}[t]{0.48\textwidth}
        \centering
        \includegraphics[width=\textwidth]{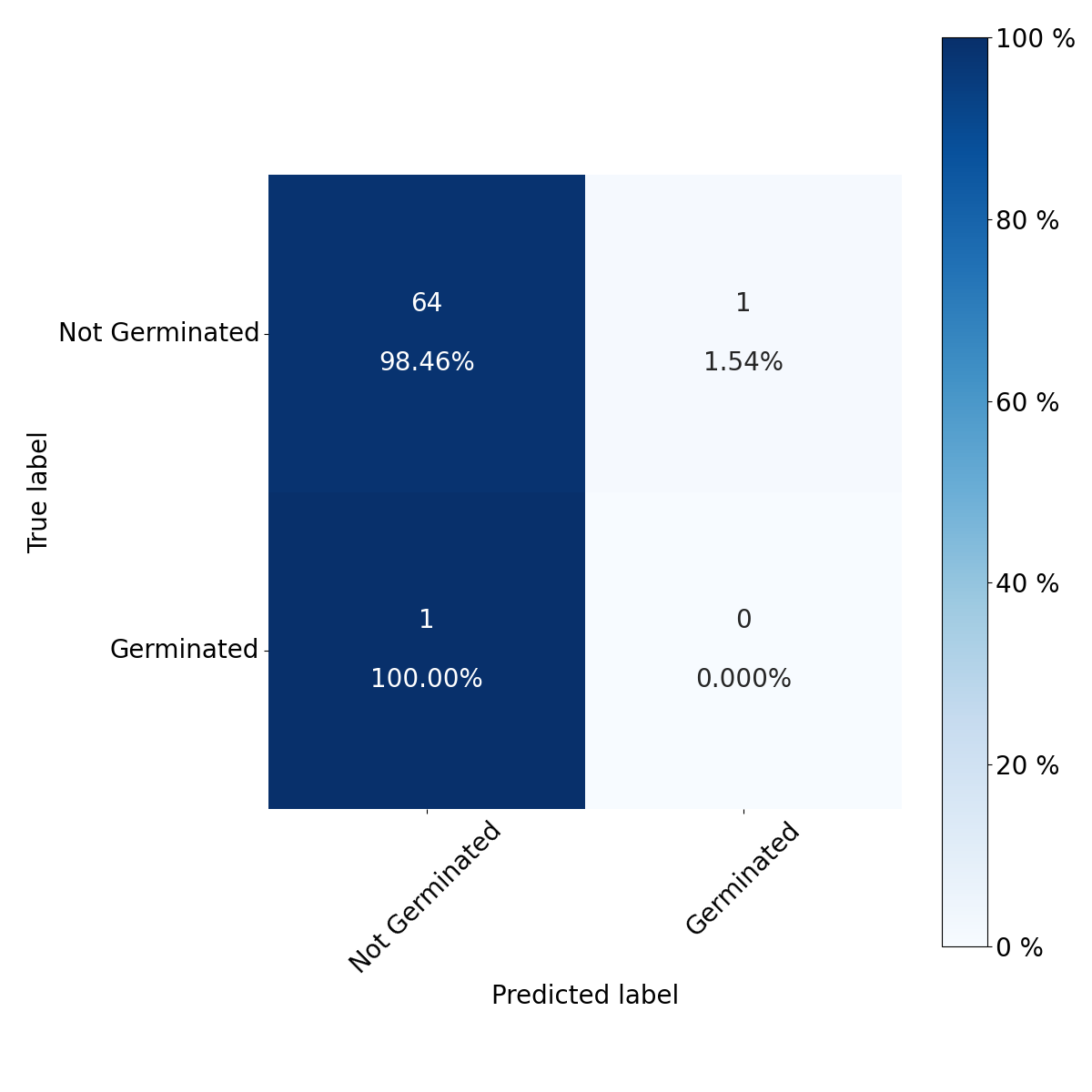}
        \caption{Prospect\_1.}
    \end{subfigure}
    \\
    \begin{subfigure}[t]{0.48\textwidth}
        \centering
        \includegraphics[width=\textwidth]{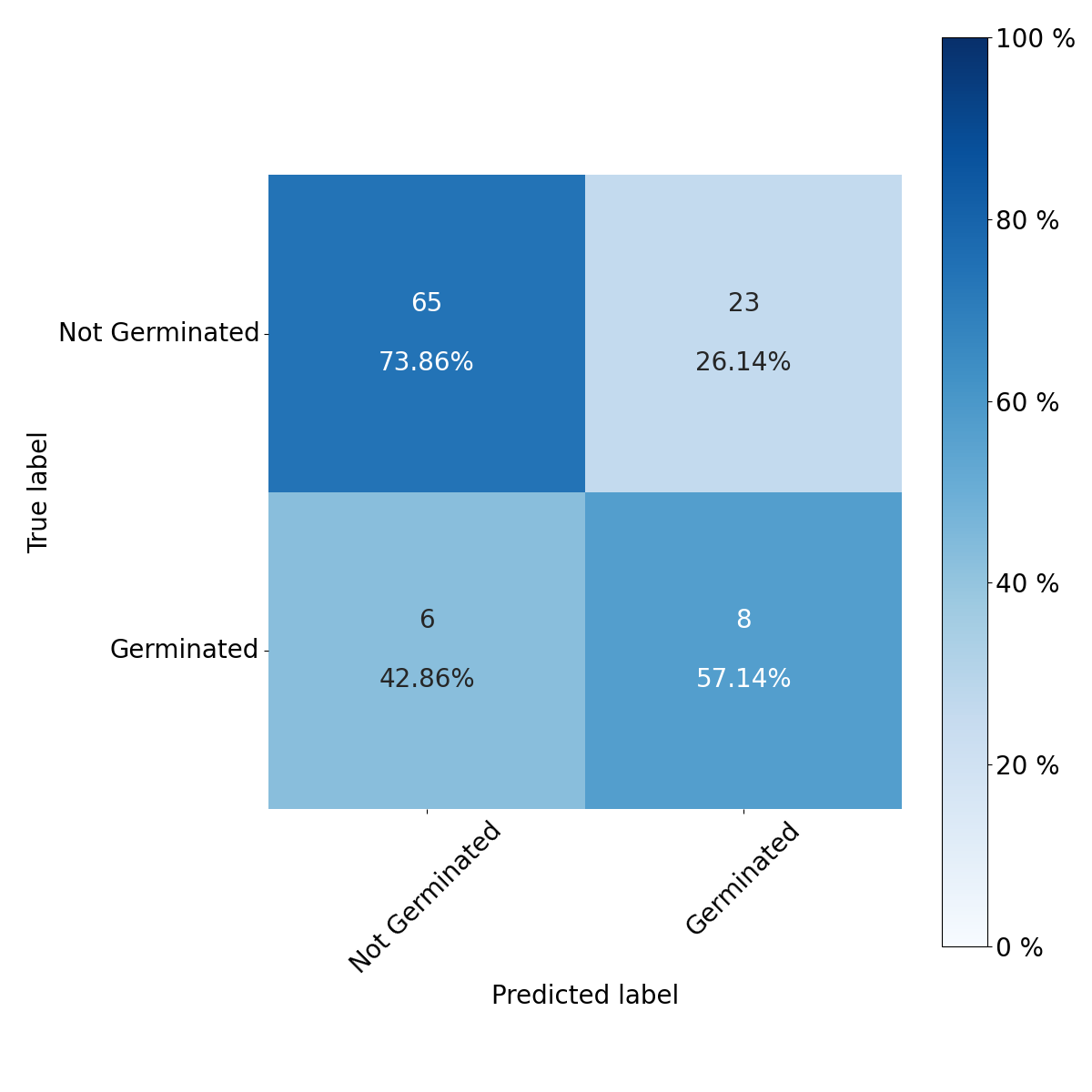}
        \caption{Laureate\_0.}
    \end{subfigure}
    \hfill
    \begin{subfigure}[t]{0.48\textwidth}
        \centering
        \includegraphics[width=\textwidth]{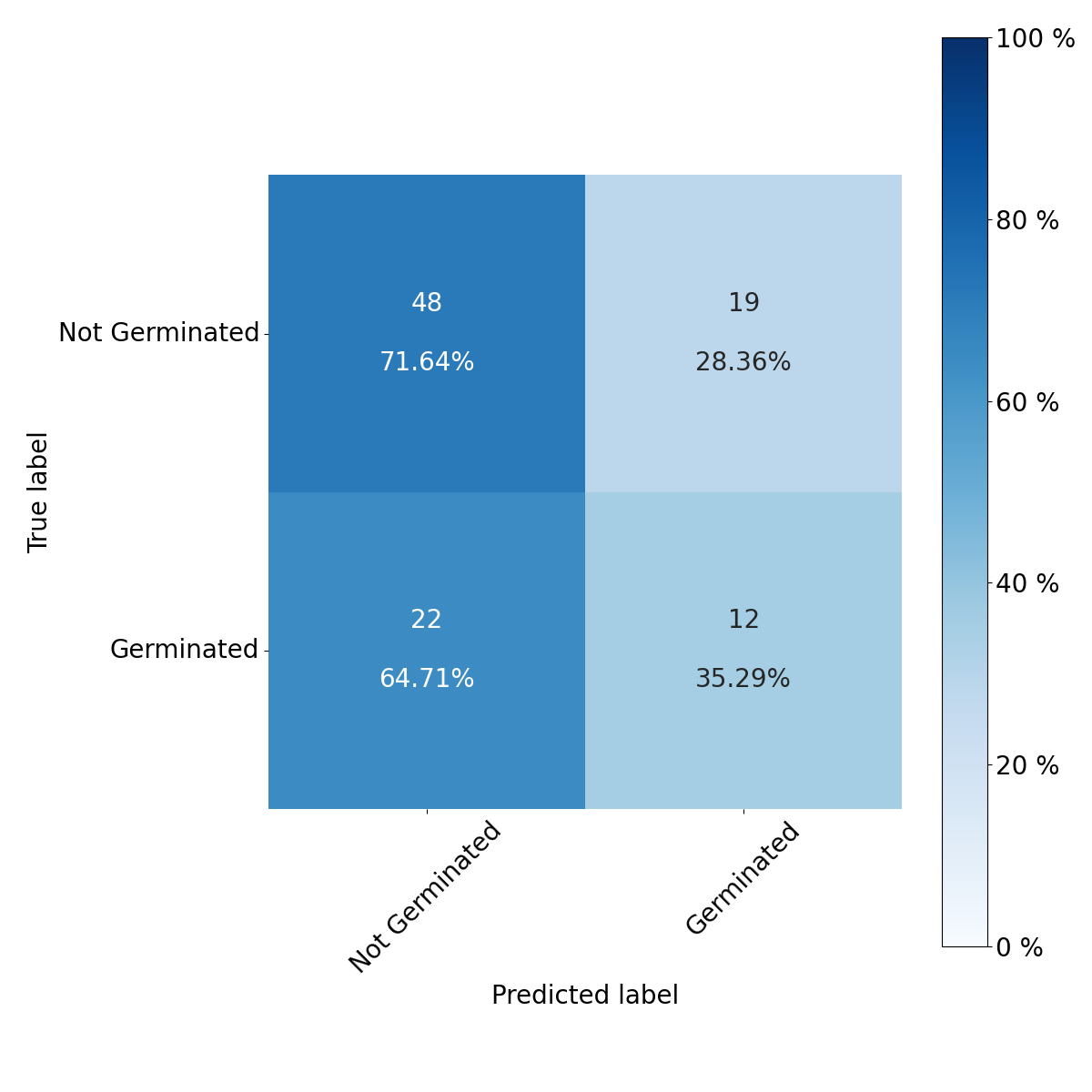}
        \caption{Laureate\_1.}
    \end{subfigure}
    \caption{Confusion matrices for test set predictions by the ResNet-18 NIR-HSI model on task 2.}
    \label{fig:conf_mat_task_2_resnet_nir_hsi}
\end{figure*}

\mytabref{tab:germination_test_results} shows balanced accuracy on the test set for each model for each barley variety. Balanced accuracy is the mean of recall for the two classes (germinated and not germinated). ResNet-18 on NIR-HSI is the best model overall, winning on 6/8 metrics. To avoid cluttering this section, we show confusion matrices only for ResNet-18 on NIR-HSI images. \myfigref{fig:conf_mat_task_1_resnet_nir_hsi} and \myfigref{fig:conf_mat_task_2_resnet_nir_hsi} show confusion matrices for tasks 1 and 2, respectively.

\begin{table}[h]
\centering
\begin{tabular}{@{}cccccc@{}}
\toprule
             & & \multicolumn{4}{c}{barley variety} \\ \cmidrule{3-6}
Task & Model and Modality & P\_0 & P\_1$^{*}$ & L\_0 & L\_1\\ \midrule
1 & PLS NIR & 56 & 42 & 76 & 56 \\
1 & ResNet-18 RGB & 71 & 46 & 78 & 76 \\
1 & ResNet-18 NIR-HSI & \textbf{78} & \textbf{48} & \textbf{86} & \textbf{86} \\ \midrule
2 & PLS NIR & 56 & 45 & \textbf{71} & 54 \\
2 & ResNet-18 RGB & 48 & 40 & 64 & \textbf{57} \\
2 & ResNet-18 NIR-HSI & \textbf{63} & \textbf{49} & 66 & 53 \\ \bottomrule
\end{tabular}
\caption{Test set balanced accuracy (in \%) for each barley variety. P and L are short for Prospect and Laureate, respectively. The best score for each barley variety and task is in \textbf{bold}. $^{*}$Prospect\_1 only has a single germinated kernel in the test set. Wrongfully predicting this one kernel will give a balanced accuracy of at most 50\%.}
\label{tab:germination_test_results}
\end{table}

\section{Discussion}\label{sec:germination_discussion}
From the analyses conducted in this study, our dataset seems insufficient for thorough modeling of the germination process. There is a clear problem with too few germinated kernels due to the issues with glue residue mentioned by \citet{engstrom2025dataset}. In particular, Prospect\_1 having only five kernels germinate in total renders meaningful modeling and evaluation virtually impossible. The other barley varieties have a higher germinative capacity but have also been severely inhibited. Furthermore, for task 2, we assume equal conditions for the kernels. In particular, for task 2, we cannot model if the glue residue will inhibit some kernels more than others. This is because the images used for task 2 are acquired before the glue is mixed with the kernels. In practice, we do not know how much glue was absorbed by each kernel.

The labeling was done manually by inspecting the RGB images, which have the highest spatial resolution, thus making labeling easier for a human. During image acquisition, germinated kernels were placed to make the sprouts visible to the cameras. Therefore, we do not expect systematic errors due to germinated kernels facing away from the camera.

Inspecting \mytabref{tab:germination_test_results}, applying PLS to NIR spectra seems to yield about the same performance on tasks 1 and 2. It seems better than random, indicating that spectral information may be relevant for both tasks. By analyzing RGB images with ResNet-18, the detection of germinated kernels is decently accurate. However, considering task 2, the performance of RGB imaging deteriorates significantly and appears more or less random, indicating that spatial and color information alone is insufficient to predict germinative capacity. Still, RGB imaging can undoubtedly be used to detect physical defects such as cleaved kernels. As the dataset consisted of kernels without visible damage, this potential capacity has had no influence on the metrics. Finally, applying ResNet-18 to NIR-HSI images is the most promising strategy. It can detect germinated kernels to a higher degree than both RGB imaging and PLS on NIR spectra, and it also outperforms them on task 2 on average. Disregarding Prospect\_1, it seems that both PLS on NIR spectra and ResNet-18 on NIR-HSI images can predict whether a barley kernel will germinate or not to a degree better than random. Still, further studies on a larger dataset are required to reach firm conclusions.

In reality, barley germination is a continuous process, and so it is likely best modeled as such. However, due to the manual inspection of kernels at discrete intervals, like \citet{arngren2011analysis}, we assign labels based on snapshots of the process at these intervals. Still, a prediction closer to the actual germination day is better than a prediction further away. For example, if a kernel germinates on day 1, predicting day 2 is better than predicting day 3 and should be treated as such. This realization is relevant for task 3, which we decided not to endeavor. For NIR-HSI analysis to be truly useful as a standard analytical method, task 3 must be solved (\citealt{orth2025current}). If we were to undertake task 3, a continuous metric such as (weighted) mean squared error would have been appropriate for the regression. However, some kernels do not germinate, so assigning a continuous label to them seems challenging. A possible solution to this would be joint classification, based on task 2, and subsequent regression on kernels predicted to germinate by the classifier. By modifying ResNet-18 to have both a classification and a regression head, it can simultaneously make predictions for task 2 and regression predictions for task 3. Such a strategy requires joint optimization, which may be helpful for the overall performance due to cross-learning.

Finally, the stratified sampling in this study was not done per barley variety. So, although the distributions of germinated and non-germinated kernels are identical across partitions, the distributions of germinated and non-germinated kernels for each variety are not necessarily identically distributed. The class weights, however, are computed for each variety and will mitigate this issue. Still, due to the very few germinated kernels, there may simply not be enough variation in the images used for training and validation. Indeed, despite the class weights, the confusion matrices shown in \myfigref{fig:conf_mat_task_1_resnet_nir_hsi} and \myfigref{fig:conf_mat_task_2_resnet_nir_hsi} seem to be biased towards predicting not germinated for all varieties. Although not shown here, the confusion matrices for PLS and ResNet-18 on RGB images exhibit a similar pattern throughout both tasks on all barley varieties. The exception is PLS for task 1 on both Laureate samples. However, as both Laureate samples germinate more than both Prospect samples, perhaps the PLS model has been biased accordingly despite the weighting due to the low variation.

Further investigations based on the same dataset could resolve some of the issues mentioned in this discussion. However, a larger dataset is needed for conclusive results.

\section{Conclusion}
Initially, we set out to solve three tasks. Namely, predicting whether a barley kernel has germinated, whether it will germinate, and, finally, on which day it will germinate if it will germinate. However, we decided that further modeling would not yield conclusive results due to the performance on the first two tasks and a limited amount of germinated kernels for each day. While the dataset used for this analysis has its clear limitations, we see indications that NIR-HSI is a superior modality to both NIR spectroscopy and RGB imaging for detecting already germinated kernels. Further studies on a larger dataset are required for stronger conclusions, especially regarding the prediction of barley's germinative capacity. Still, we maintain a careful optimism about using joint spatio-spectral analysis of barley's germinative capacity through NIR-HSI images.

\part{ Final Remarks}\label{part:final_remarks}
\chapter{Discussion}\label{chap:discussion}
\lettrine{T}{his} chapter discusses the hypotheses, studies, and results presented in the previous parts. The discussion is structured in three sections. \mysecref{sec:fast_pls} is concerned with the PLS and fast cross-validation algorithms presented in \mypartref{part:fast_pls}. Likewise, \mysecref{sec:nir-hsi} offers an analysis of the results achieved in \mypartref{part:hsi} and relates them to the five research hypotheses enumerated in \mychapref{chap:intro}. Finally, \mysecref{sec:future_work} offers perspectives for future research directions based on this thesis.

\section{Fast Partial Least Squares}\label{sec:fast_pls}
Recall that PLS models a linear relationship between $\X \in \mathbb{R}^{N \times K}$ and $\Y \in \mathbb{R}^{N \times M}$ yielding a regression matrix $\B \in \mathbb{R}^{K \times M}$. \mychapref{chap:ikpls} presented IKPLS Algorithms 1 and 2. IKPLS Algorithm 1 is fast if $N \approx K$ and has reasonable runtime if $N < K$. Conversely, if $N > K$, then IKPLS Algorithm 2 is the faster choice, as it computes $\XT\X$ and uses this instead of the larger $\X$ to derive $\B$. These algorithms are well-suited for NIR-HSI as $K$ is typically in the hundreds, while $N$ may be much larger as in the datasets analyzed in \mychapref{chap:cnn_design}. Conversely, for datasets of NIR spectra measured with spectrometers, it may well be the case that $N \ll K$. In this case, IKPLS Algorithm 2 will be devastatingly slow, and while IKPLS Algorithm 1 will be reasonably fast, an alternative that derives $\B$ through $\X\XT$ will be much faster. This could, for example, be the PLS algorithms devised by \citet{rannar1994pls}, computing both $\X\XT$ and $\Y\YT$, or \citet{liland2020much}, computing only $\X\XT$.

When computing $\X\XT$ or $\Y\YT$, the fast cross-validation algorithms introduced in \mychapref{chap:fast_cv} can not be used to speed up cross-validation, as it computes $\XT\X$ and $\XT\Y$. Instead, in this case, another fast cross-validation algorithm by \citet{rannar1995pls} with improvements by \citet{liland2020much} can be used to speed up the computation of $\X\XT$ and $\Y\YT$. Four unique combinations exist of centering and scaling for $\X\XT$, namely, centering, scaling, neither, and both. The fast cross-validation algorithm by \citet{liland2020much} correctly handles centering and no preprocessing but relies on the dataset-wise standard deviation for scaling, thus introducing data leakage from validation partitions to training partitions. This is unlike the algorithms in \mychapref{chap:fast_cv}, which correctly handle all 16 (12 unique) combinations of centering and scaling for $\XT\X$ and $\XT\Y$.

Another point about the fast cross-validation algorithms presented in \mychapref{chap:fast_cv} is regarding asymptotic optimality. If an algorithm is asymptotically optimal, then it is, at most, a constant factor worse than the best possible algorithm that can theoretically exist. The following is an argument that, in a specific setting, our algorithms (\citealt{engstrom2025fast}) have asymptotically optimal time and space complexities. Recall that the time complexity is $\Theta(NK(K+M))$ and the space complexity is $\Theta((N+K)(K+M))$. Also, recall that $P$ is the number of partitions (folds) in the cross-validation scheme. To recap, the space complexity boils down to storing the input $\X$ and $\Y$ and a single pair of matrix products $\XT\X$ and $\XT\Y$. The space complexity is then based on the assumption that matrix product pairs for all training partitions will use the same space. That is, the space used under one partition will be overwritten under the next partition. Consider the case where the training partition-wise matrix products must be stored, perhaps for later analysis with multiple methods. Then the space complexity increases to $\Theta((N+PK)(K+M))$ while the time complexity is unaffected. Consider the case of leave-one-out cross-validation (LOOCV), the case where $P=N$. Then the space complexity becomes $\Theta((N+NK)(K+M))=\Theta(NK(K+M))$, matching the time complexity. Time complexity can never decrease below space complexity as it requires at least one unit of time to fill one unit of space. Under the conditions that we must store $\X$, $\Y$, and all $P=N$ pairs of $\XT\X$ and $\XT\Y$, the space complexity matches exactly this. Thus, neither space nor time complexity can be reduced further, and the algorithm is asymptotically optimal under these conditions.

\section{Near-Infrared Hyperspectral Image Analysis}\label{sec:nir-hsi}

\subsection{Joint Spatio-Spectral Modeling}
Based on the studies in Chapters \ref{chap:cnn_design} and \ref{chap:barley_germination}, it seems that joint spatio-spectral analysis of the entire NIR-HSI image may outperform spatial or spectral analysis alone for some parameters. In particular, when both spatial and spectral information is relevant for the parameter, as is the case with grain variety classification (\citet{dreier2022hyperspectral, engstrom2023improving}) and classification of germinated and non-germinated barley kernels (\citealt{orth2025current}). Conversely, spectral analysis seems to perform comparably with joint spatio-spectral analysis for regression of chemical parameters (\citealt{engstrom2021predicting, engstrom2023improving, engstrom2025chem}). This is shown for regression of protein content in wheat in \mychapref{chap:cnn_design} and for fat content regression in pork bellies in \mychapref{chap:chem_maps}.

It should be noted, however, that these studies compare ResNets for NIR-HSI, RGB, and grayscale images with PLS for NIR spectra. Thus, differences in performance may be due to both the data modality and the models themselves. For example, a more thorough analysis could include a 1D ResNet for NIR spectral analysis. Still, as PLS is the primary tool for analysis of NIR spectra of food (\citealt{brereton2018chemometrics, sorensen2021nir}), and CNNs are the primary tool for image analysis (\citealt{lecun2015deep}), the comparisons can likely be deemed valid.

\subsection{Learnable Preprocessing}
Spectral preprocessing, learnable or manually applied, is highly beneficial when using 2D CNNs for joint spatio-spectral modeling of chemical parameters (\citealt{engstrom2021predicting, engstrom2023improving}). \mychapref{chap:cnn_design} showed how augmenting a 2D CNN with an initial spectral (1D) or spatio-spectral (3D) convolution layer with only a few filters can be used to increase performance on protein content regression in wheat kernels, and we hypothesize that this generalizes to other chemical parameters as well. In particular, the initial layer learns weights much similar to those of a Savitzky-Golay (SG) filter (\citealt{savitzky1964smoothing, steinier1972smoothing}). Thus, CNNs can learn suitable preprocessing directly from the data, contrasting the manual selection of preprocessing for PLS. However, the number of convolution filters and the kernel size must still be chosen manually. Additionally, each filter will output a transformed representation of the input NIR-HSI image. This implies that if $n$ filters are used, then the output of this initial layer will be $n$ times as large as the input NIR-HSI image, which may already be quite large. Therefore, using multiple filters is associated with a significant computational overhead, both in terms of memory and runtime.

Alternatively, as shown in \mychapref{chap:cnn_design}, a spectral preprocessing may be chosen by grid search cross-validation of a PLS model and subsequently applied before feeding the input data to a 2D CNN to yield results comparable to using the initial 1D or 3D convolution layer. This approach decreases the computational overhead of training the CNN, especially since the PLS cross-validation can be carried out with the algorithm presented in \mychapref{chap:fast_cv}. Still, this assumes a correlation between the optimal preprocessing for PLS and CNNs and requires that the optimal preprocessing is included in the grid search. Therefore, it may be more reasonable to have the CNN learn directly from the input data instead of relying on PLS to find a suitable preprocessing. When learning from the data, the studies by \citet{engstrom2021predicting} and \citet{engstrom2023improving} showed that applying a single 1D spectral filter may lead to unstable training. Therefore, a general suggestion is to use a single initial spatio-spectral filter, which offers better stability and does not multiply the size of the input NIR-HSI image in its output. Indeed, this is the approach taken in the subsequent studies in \mychapref{chap:chem_maps} and \mychapref{chap:barley_germination}.

In addition to the initial 1D and 3D convolutions, we experimented with so-called downsamplers in \mychapref{chap:cnn_design}. Downsamplers reduce the spectral dimension to one by computing a dot product between a learned vector and every spectrum. By stacking the outputs from multiple downsamplers, we can control the size of the spectral dimension in the output of this layer, not unlike choosing the number of components for a PLS model. Such downsampling has a regularizing effect and may be beneficial in some cases, as shown in \ref{chap:cnn_design} and \ref{chap:barley_germination}. Still, this layer also requires manually choosing the number of filters. Another possible approach is to use many downsamplers and then apply L1 regularization to the output like in sparse autoencoders (\citealt{bank2023autoencoders}). This way, the spectral regularization can be coupled with the loss function for joint optimization of spectral regularization and objective. Either approach, however, requires manual tuning of hyperparameters.

\subsection{Bulk References}
Chapters \ref{chap:cnn_design} and \ref{chap:bulk_references} studied datasets of NIR-HSI images of bulk grain samples with reference values for grain variety and mean protein content, respectively. To generate enough data to train CNNs, we cropped the images of bulks and the bulks' reference values to their respective image crops. This data inflation strategy is entirely valid for the grain variety dataset as each NIR-HSI image contained exactly one grain type. On the other hand, wheat's protein content is heterogeneously distributed, so the bulk reference values may not reflect the true protein content in the image crops. On average, however, by definition, the bulk reference value will match the true protein content in the image crops. Therefore, we hypothesized that training an accurate regression model with these pairs of image crops and bulk mean reference values would be possible. However, as shown in \mychapref{chap:bulk_references}, the average protein prediction for all image crops belonging to a bulk is not a good estimate of the bulk's mean protein reference value.\footnote{In the analysis by \citet{engstrom2023analyzing}, all image crops are weighted equally when taking their average predicted protein value. However, they do have different grain density (grain-to-background ratio). Thus, using a grain-density-weighted average would have been more accurate when considering the mean prediction. Still, as the grain was randomly distributed on the images and each bulk yielded more than 1,000 image crops, any errors due to the non-weighted average are assumed to cancel out.} This is due to a discrepancy between the least squares solution for regression from image crop-wise predictions, respectively, \textit{average} image crop-wise predictions to bulk references. Therefore, bias and scale corrections are needed to accurately estimate bulk mean protein content when pairing subsamples with bulk references. In fact, deep learning models should always have their bias, and sometimes scale, corrected for any task (\citealt{igel2023remember}).

When correcting for the bias and scale, the mean predictions align well with the bulk mean reference values. However, a similar performance can be achieved much faster by applying PLS to the mean spectra of the bulks and their reference values (\citealt{engstrom2023analyzing}). Thus, applying PLS to the mean spectrum is the most sensible choice if the mean of a chemical parameter is of interest and only bulk mean references are available. Additionally, a mean spectrum can be acquired with a spectrometer, even accurately for heterogeneous samples (\citealt{tanzilli2024does}), avoiding the need for an expensive NIR-HSI platform.

\subsection{Chemical Maps}
However, a NIR-HSI image is required if a chemical parameter's spatial distribution is to be modeled, that is, chemical map generation. Based on the paragraphs above, it may seem that training a PLS model on a mean spectrum with an associated bulk mean reference can readily be used to generate chemical maps through pixel-wise predictions. As mentioned in \mysecref{sec:related_work}, this approach is commonly taken. \mychapref{chap:chem_maps} studied chemical map generation of fat in pork bellies and showed that pixel-wise predictions may lead to unsatisfactory chemical maps with predictions of physically impossible fat values vastly outside the $0-100\%$ range (\citealt{engstrom2025chem}). Additionally, there is a discrepancy between the fat predictions on mean spectra and the fat predictions obtained by averaging the chemical maps (this is also noticed by, for example, \citealt{BARBIN20131162}). Instead, by augmenting a 2D U-Net with a spectral convolution filter as studied in \mychapref{chap:cnn_design}, chemical maps could be generated directly from the input NIR-HSI image and optimized against a bulk mean reference. This approach handled all the issues mentioned for pixel-wise PLS and showed that joint spatio-spectral analysis can outperform spectral analysis of the spatial distribution of a chemical parameter.

\subsection{Barley Germination}
NIR-HSI images are beneficial for analyzing parameters where both spatial and spectral information are relevant. However, we had yet to find a food quality parameter that could not be somewhat accurately modeled with spectral or spatial information alone, apart from modeling the spatial distribution of a chemical parameter. Modeling barley's germinative capacity seemed a promising candidate as barley germination relies on both chemical and physical, visible parameters (\citealt{orth2025current}). \mychapref{chap:barley_germination} studied this. Unfortunately, the dataset suffered from a low degree of germination (\citealt{engstrom2025dataset}), making modeling and evaluation unreliable. A larger dataset with more germinated kernels is required to determine whether barley's germinative capacity can be accurately modeled with NIR-HSI. On the other hand, classification of germinated and non-germinated kernels was possible with both NIR spectra, RGB images, and NIR-HSI images, and the latter modality gave the highest balanced accuracy. These results corroborate earlier studies where joint spatio-spectral modeling outperformed spectral and spatial modeling alone of parameters where spatial and spectral information are relevant.

\section{Future Work}\label{sec:future_work}
In \mychapref{chap:ikpls}, we introduced the IKPLS implementation that was later augmented to incorporate sample weighting (\citealt{becker2016accounting}). The fast cross-validation algorithm presented in \mychapref{chap:fast_cv} does not work with sample weights. Thus, augmenting the algorithm to work with weights is a natural extension. Indeed, such augmentation seems feasible while maintaining proper centering and scaling and without increasing either space or time complexity. In the same spirit, attempts to augment the fast cross-validation algorithm by \citet{liland2020much} with sample weights and standard deviation scaling could be made, both jointly and separately, to decrease the computational cost of cross-validation on wide datasets.

The results achieved for chemical map generation in \mychapref{chap:chem_maps} yield several interesting directions for future research. As mentioned by \citet{engstrom2025chem}, using U-Net for simultaneous modeling of chemical maps for multiple parameters may enhance results due to cross-learning and the possibility of adding terms in the loss function regarding the interaction between multiple chemical parameters. Coupling this with semantic segmentation to distinguish food samples from the background in NIR-HSI images may yield a complete end-to-end chemical map generation model.

Additionally, the work by \citet{engstrom2025chem} showed that a CNN can be trained with bulk mean references and relatively few images. This indicates a possibility for training on large NIR-HSI images of entire bulk grain samples, thus circumventing the need to generate image crops, apply mean references to these, and subsequently compensate for the biases as otherwise done in Chapters \ref{chap:cnn_design} and \ref{chap:bulk_references}. In particular, it seems possible that a single model can simultaneously perform instance segmentation to identify individual kernels and subsequently regress the protein content in these, optimized such that the model's mean protein prediction across the entire image matches the bulk reference value. As it is feasible to obtain reference protein values for a few grain kernels, these can be used to evaluate the accuracy of the predictions on a kernel level, not just on a bulk level. With this framework, it may also be possible to train the model on large datasets of individual grains with bulk references and later fine-tune it on a small dataset with kernel-wise reference values.

As previously mentioned, the results in \mychapref{chap:barley_germination} warrant further research to yield a conclusive result regarding whether NIR-HSI enables accurate modeling of barley's germinative capacity. The insights from \citet{engstrom2025chem} regarding using bulk mean references to model the distribution of parameters may also apply to this task, albeit with slight modification. Indeed, if the use of bulk mean references allows accurate modeling, the dataset acquisition process used by \citet{engstrom2025dataset} simplified, reducing both manual labor and the complexity of the image acquisition pipeline. In particular, the following setup is imagined.
\begin{enumerate}
    \item Acquire images of several bulks of barley kernels once before exposure to moisture.
    \item Expose each bulk grain sample to moisture to initialize the germination process.
    \item For each bulk, count the number of germinated kernels at discrete intervals, for example, once a day.
\end{enumerate}

Then, for each bulk image, we train a model for simultaneous instance segmentation (localization of individual kernels) and prediction of germinative capacity (how many days until the kernel will sprout) for each kernel. Then, due to step 3, we have the entire discrete distribution of germinative capacities. Therefore, the model can be optimized to minimize the Kullback-Leibler divergence (\citealt{kullback1951information}) between the true distribution and the predicted distribution. Finally, as with the future research idea regarding protein distribution in grain, the model can be fine-tuned and evaluated on a smaller dataset of barley kernels with individual reference values. Unlike the approach taken by \citet{engstrom2025dataset}, this approach requires only a single imaging session per grain kernel and does not require the tracking or labeling of each kernel.

In summary, there seems to be ample opportunity for modeling parameters in the future using both bulk and individual reference values.
\chapter{Conclusion}\label{chap:conclusion}
\lettrine{T}{he} first part of this thesis has presented two practical open-sourced tools for Python that speed up classical chemometrics. The first tool contains several fast PLS implementations that are particularly useful for datasets with many samples. The second tool implements a set of exceptionally fast
algorithms that can speed up cross-validation for PLS and many other traditional machine learning and chemometric models.

The second part of this thesis has investigated food analysis through applications of NIR-HSI images for PLS- and CNN-based modeling. Importantly, studies showed how extending any preexisting 2D CNN with an initial, dedicated spectral convolution layer significantly enhances its ability to predict chemical parameters in NIR-HSI images by learning spectral preprocessing. Furthermore, compared to spatial and spectral analysis, joint spatio-spectral analysis of entire NIR-HSI images improves the modeling of parameters where both chemical and visual information are relevant.

Additionally, while analyzing spectral information alone is sufficient and recommended for accurate modeling of chemical parameters, the joint spatio-spectral analysis yields significantly better results than pixel-wise spectral analysis when modeling the spatial distribution of chemical parameters to generate chemical maps. Indeed, this thesis presented a novel approach to chemical map generation, requiring only a bulk mean reference value to estimate the chemical map while considering spatial dependencies. Thus, this study showed that NIR-HSI images can be used to model the spatial distribution of chemical parameters throughout the images.

Studies also investigated pairing bulk mean reference values with subsamples for dataset inflation. Generally, this approach is feasible and may yield accurate predictions of the bulk mean references. However, a bias and scale correction must be applied post-training when this approach is applied to regression analysis on heterogeneous distributions.

The last study in this thesis investigated barley's germinative capacity as a candidate for a food quality parameter requiring NIR-HSI for accurate modeling. However, it was inconclusive, primarily due to a fault in the dataset acquisition that led to a limited amount of germinated kernels. Therefore, it remains unknown whether a parameter exists that can not be analyzed decently accurately with spatial or spectral information but can be with joint spatio-spectral analysis.

This thesis concludes that joint spatio-spectral analysis of NIR-HSI images is beneficial and should be considered over spectral or visual analysis alone in two cases. First, it improves the modeling of parameters where both chemical and visual information are relevant. Second, it improves the modeling of spatial distributions of chemical parameters compared to pixel-wise spectral analysis.

Finally, based on the results achieved, this thesis offered perspectives on future research of NIR-HSI applications for food analysis. In particular, it may be possible to accurately model both discrete and continuous distributions of quality parameters in heterogeneous bulk samples of grain using cheap-to-obtain bulk mean reference values.

\cleardoublepage
\phantomsection
\addcontentsline{toc}{part}{Bibliography}
\defbibheading{mybib}{\chapter*{#1}}
\printbibliography[heading=mybib]

\part{Appendices}\label{part:appendices}
\appendix
\chapter{Technical Report: Bulk References}\label{app:bulk_ref}
The technical report by \citet{engstrom2023analyzing} follows. Two typos should be addressed: In Figure 7, subfigure e is the validation set, and f is the test set.
\includepdf[pages=-]{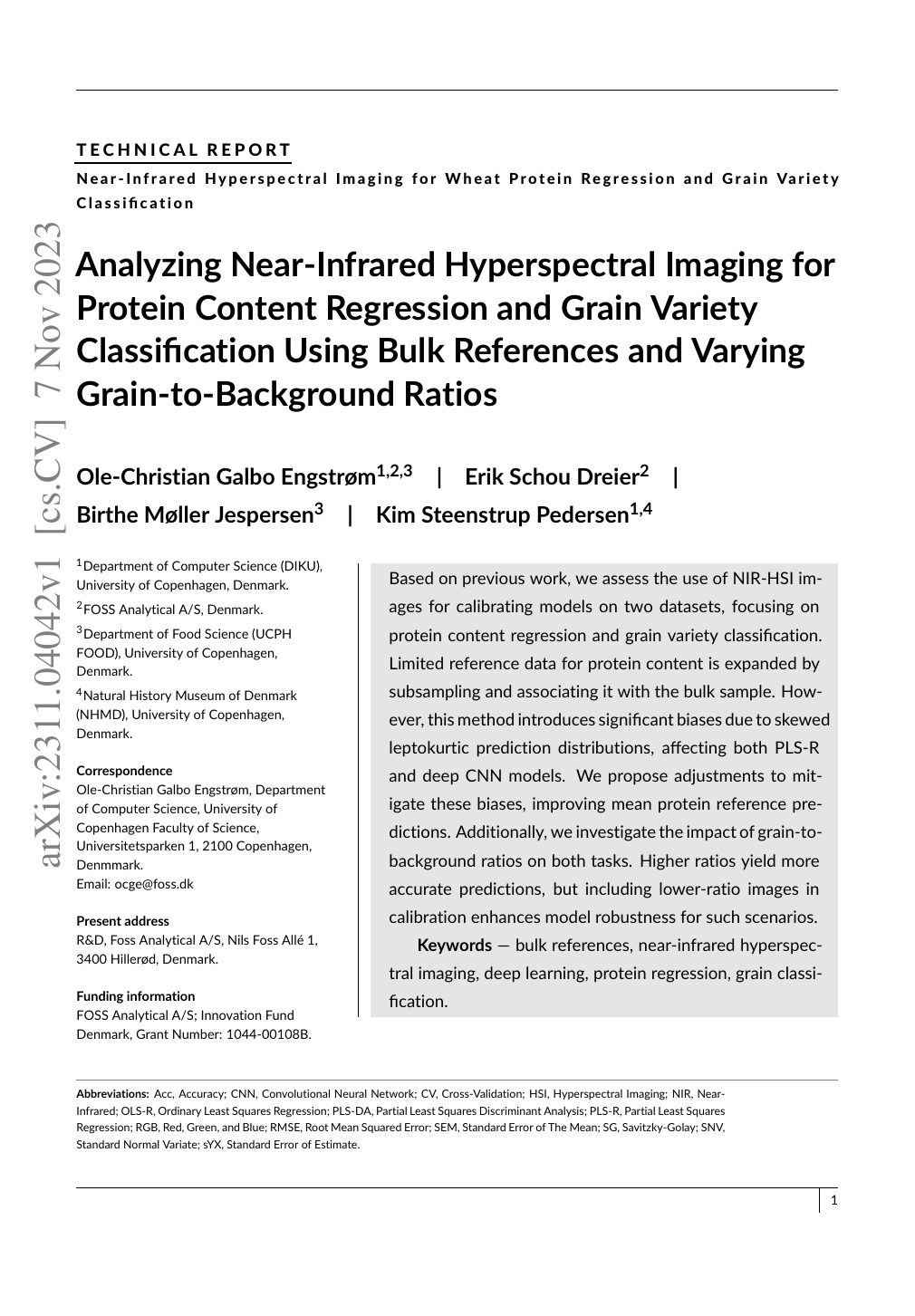}
\chapter{Technical Report: Barley Germination Dataset}\label{app:germination_dataset}
The technical report by \citet{engstrom2025dataset} follows.

\includepdf[pages=-]{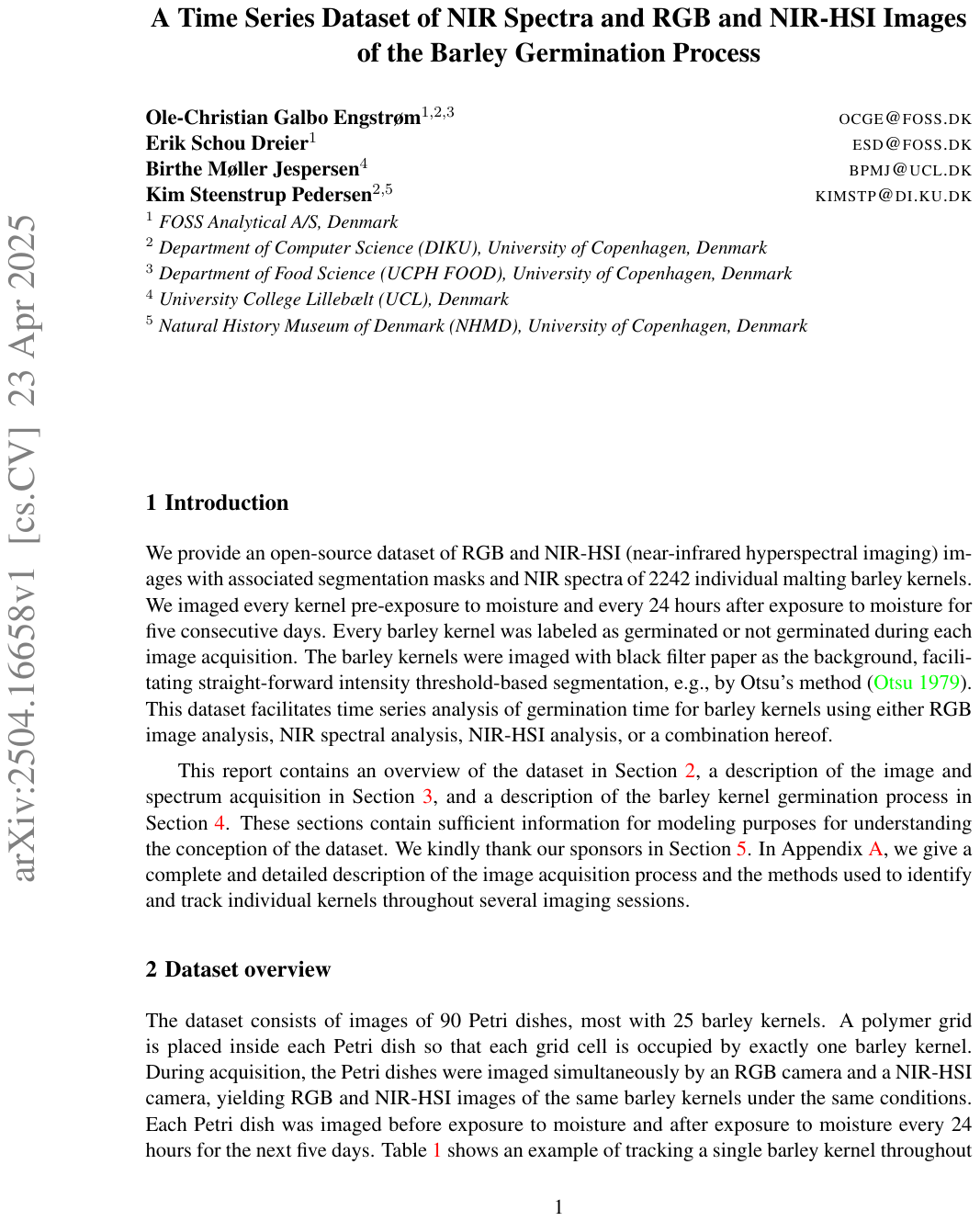}

\end{document}